\documentclass[10pt,journal,compsoc]{IEEEtran}
\ifCLASSOPTIONcompsoc
  \usepackage[nocompress]{cite}
\else
  \usepackage{cite}
\fi
\usepackage{graphicx}
\usepackage{amsmath}
\usepackage{amssymb}
\usepackage{booktabs}
\usepackage{multirow}
\usepackage[inline]{enumitem}
\usepackage{pifont}%
\usepackage{algorithm}
\usepackage{algpseudocode} 
\newcommand{\cmark}{\ding{51}}%
\newcommand{\xmark}{\ding{55}}%
\usepackage[dvipsnames]{xcolor}
\usepackage[colorlinks,linkcolor=red, filecolor=black, urlcolor=purple, citecolor=RoyalBlue]{hyperref}
\usepackage[capitalize]{cleveref}
\crefname{section}{Sec.}{Secs.}
\Crefname{section}{Section}{Sections}
\Crefname{table}{Table}{Tables}
\crefname{table}{Tab.}{Tabs.}

\def\eg{{\textit{e.g.}}}
\def\ie{{\textit{i.e.}}}
\ifCLASSINFOpdf
\else
\fi
\begin{document}
%
\title{An Effective Motion-Centric Paradigm for 3D Single Object Tracking in Point Clouds}
\author{Chaoda Zheng$^{\dagger}$,
Xu Yan$^{\dagger}$, 
Haiming Zhang, \\ 
Baoyuan Wang, 
Shenghui Cheng,  
Shuguang Cui~\IEEEmembership{Fellow,~IEEE},
Zhen Li$^{\star}$
\IEEEcompsocitemizethanks{\IEEEcompsocthanksitem C. Zheng, X. Yan, H.Zhang, S.Cui and Z. Li are currently with the School of Science and Engineering (SSE), the Future Network of Intelligence Institute (FNii), and the Guangdong Provincial Key Laboratory of Future Networks of Intelligence, the Chinese University of Hong Kong, Shenzhen. B. Wang is with Xiaobing.AI and S. Cheng is with Westlake University.\protect\\
E-mail: \{chaodazheng, xuyan1, haimingzhang\}@link.cuhk.edu.cn, \hfil\break \{lizhen,shuguangcui\}@cuhk.edu.cn,
}
\thanks{$\dagger$ denotes equal contribution. $\star$ denotes corresponding author.}}
%
%

\markboth{Journal of \LaTeX\ Class Files,~Vol.~14, No.~8, Octorber~2022}%
{Shell \MakeLowercase{\textit{et al.}}: Bare Demo of IEEEtran.cls for Computer Society Journals}
%



\IEEEtitleabstractindextext{%
\begin{abstract}
  3D single object tracking in LiDAR point clouds (LiDAR SOT) plays a crucial role in autonomous driving.
  Current approaches all follow the Siamese paradigm based on appearance matching.
  However, LiDAR point clouds are usually textureless and incomplete, which hinders effective appearance matching.
  Besides, previous methods greatly overlook the critical motion clues among targets.
  In this work, beyond 3D Siamese tracking, we introduce a {\textbf{motion-centric paradigm}} to handle LiDAR SOT from a new perspective.
  Following this paradigm, we propose a matching-free two-stage tracker {\textbf{M$^2$-Track}}.
  At the $1^{st}$-stage, $M^2$-Track localizes the target within successive frames via \textbf{m}otion transformation.
  Then it refines the target box through \textbf{m}otion-assisted shape completion at the $2^{nd}$-stage.
  Due to the motion-centric nature, our method shows its impressive generalizability with limited training labels and provides good differentiability for end-to-end cycle training.
  This inspires us to explore semi-supervised LiDAR SOT by incorporating a pseudo-label-based motion augmentation and a self-supervised loss term.
  Under the fully-supervised setting, extensive experiments confirm that $M^2$-Track significantly outperforms previous state-of-the-arts on three large-scale datasets while running at \textbf{57FPS} ({\textbf{$\sim$ 3\%}}, {\textbf{$\sim$ 11\%}} and {\textbf{$\sim$ 22\%}} precision gains on KITTI, NuScenes, and Waymo Open Dataset respectively).
  While under the semi-supervised setting, our method performs on par with or even surpasses its fully-supervised counterpart using fewer than half labels from KITTI.
  Further analysis verifies each component's effectiveness and shows the motion-centric paradigm's promising potential for auto-labeling and unsupervised domain adaptation.
  The code is available at \url{https://github.com/Ghostish/Open3DSOT}.
\end{abstract}

}

\maketitle

\IEEEdisplaynontitleabstractindextext

%
\IEEEpeerreviewmaketitle

\IEEEraisesectionheading{\section{Introduction}\label{sec:intro}}

\IEEEPARstart{S}ingle Object Tracking (SOT) is a basic computer vision problem with various applications, such as autonomous driving~\cite{qi2021offboard,yan2020pointasnl,yan2021sparse} and surveillance system~\cite{thys2019fooling}. Its goal is to keep track of a specific target across a video sequence, given only its initial state (appearance and location).

\begin{figure}[t]
  \centering
   \includegraphics[width=1\linewidth]{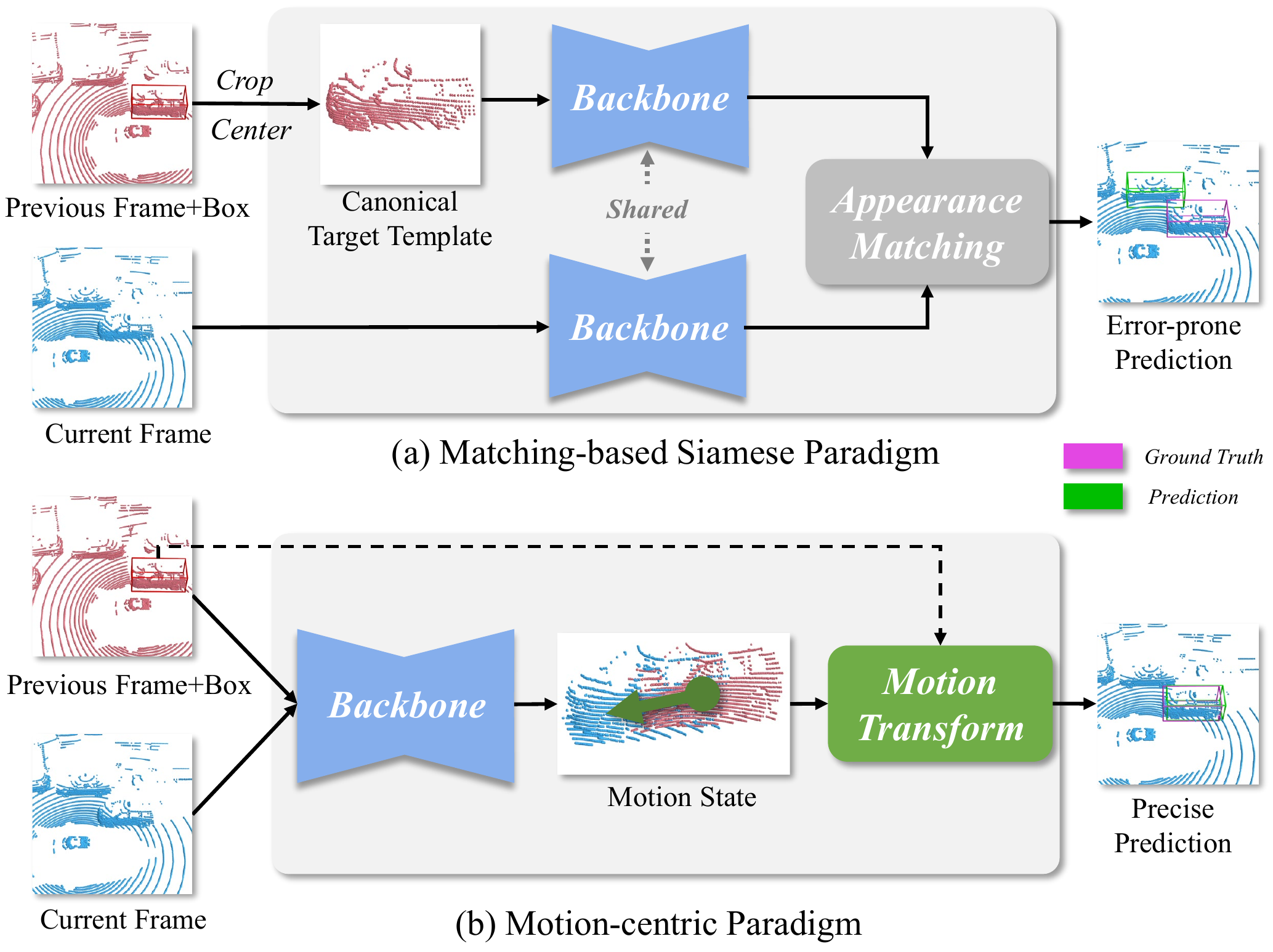}

   \caption{
     \textbf{Top}. Previous Siamese approaches obtain a canonical target template using the previous target box and search for the target in the current frame according to the matching similarity, which is sensitive to distractors. 
     \textbf{Bottom}. Our motion-centric paradigm learns the relative target motion from two consecutive frames and then robustly localizes the target in the current frame via motion transformation.
     Besides fully-supervised learning, the motion-centric nature also shows preferable properties for semi-supervised learning. 
     }
   \label{fig:fig1}
\end{figure}

Existing LiDAR-based SOT methods~\cite{Giancola_2019_CVPR,qi2020p2b,zheng2021box,zarzar2019efficient,shan2021ptt,fang20203d} all follow the Siamese paradigm, which has been widely adopted in 2D SOT since it strikes a balance between performance and speed.
During the tracking, a Siamese model searches for the target in the candidate region with an \textbf{appearance matching} technique, which relies on the features of the target template and the search area extracted by a shared backbone (see Fig.~\ref{fig:fig1}~(a)).

Though the appearance matching shows satisfactory results on KITTI dataset~\cite{Geiger2012CVPR} for the 3D SOT on cars, we observe that the data has the following proprieties: 
\begin{enumerate*}[label=\roman*)]
    \item the target's motion between two consecutive frames is minor, which ensures no drastic appearance change;
    \item there are few/no distractors in the surrounding of the target.
\end{enumerate*}
However, the above characteristics do not hold in natural scenes. 
Due to self-occlusion, significant appearance changes may occur in consecutive LiDAR views when objects move fast, or the hardware only supports a low frame sampling rate.
 Besides, the negative samples grow significantly in dense traffic scenes. In these scenarios, it is not easy to locate a target based on its appearance alone (even for human beings).
These observations suggest that appearance matching is a suboptimal solution for LiDAR SOT.
Besides appearance, the target's movements among successive frames are also critical for effective tracking, because it provides spatial and temporal clues while being insensitive to occlusion and distractors.
Knowing this, researchers have proposed various 2D Trackers to temporally aggregate information from previous frames~\cite{wang2021transformer,bhat2020know}. However, the motion information is rarely explicitly modeled since it is hard to be estimated under the perspective distortion.
Fortunately, 3D scenes keep intact information about the object motion, which can be easily inferred from the relationships among annotated 3D bounding boxes (BBoxes)\footnote{This is greatly held for rigid objects (\eg, cars), and it is approximately true for non-rigid objects (\eg, pedestrian).}.
Although 3D motion matters for tracking, previous approaches have greatly overlooked it.
 Due to the Siamese paradigm, previous methods have to transform the target template (initialized by the object point cloud in the first target 3D BBox and updated with the last prediction) from the world coordinate system to its own object coordinate system. 
 This transformation ensures that the shared backbone extracts a canonical target feature, but it adversely breaks the motion connection between consecutive frames.

\begin{figure*}[t]
    \centering
     \includegraphics[width=\linewidth]{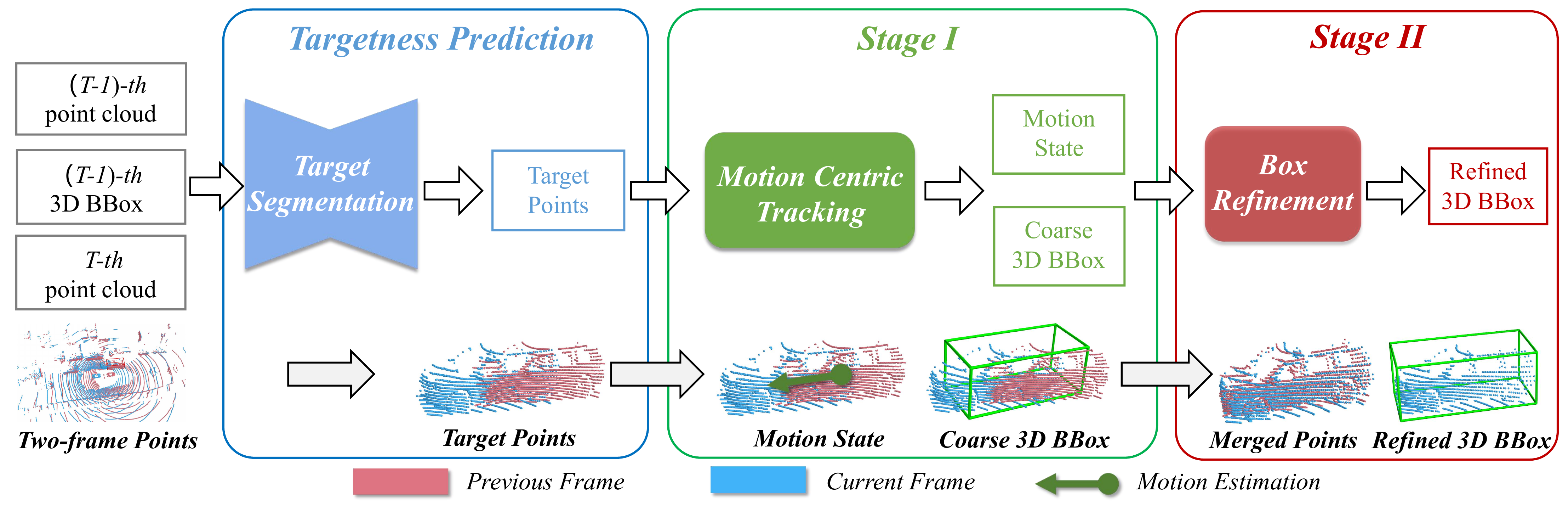}
  
     \caption{
       \textbf{The overall architecture of $M^2$-Track.}
       Given two consecutive point clouds and the possible target BBox at the previous frame, $M^2$-Track first segments the target points from their surroundings via joint spatial-temporal learning. 
       At the $1^{st}$ stage, the model takes in the target points and obtains a coarse BBox at the current frame via motion prediction and transformation.
       The coarse BBox is further refined at the $2^{nd}$ stage using motion-assisted shape completion. A detailed illustration with data flows is presented in the supplementary. 
       }
     \label{fig:pipeline1}
  \end{figure*}

Based on the above observations, we propose to tackle 3D SOT from a different perspective instead of sticking to the Siamese paradigm.
For the first time, we introduce a new \textbf{motion-centric paradigm} that localizes the target in sequential frames without appearance matching by explicitly modeling the target motion between successive frames (Fig.~\ref{fig:fig1}~(b)). 
Following this paradigm, we design a novel two-stage tracker {\textbf{M$^2$-Track}} (Fig.~\ref{fig:pipeline1}).
During the tracking, the $1^{st}$-stage aims at generating the target BBox by predicting the inter-frame relative target motion. 
Utilizing all the information from the $1^{st}$-stage, the $2^{nd}$-stage refines the BBox using a denser target point cloud, which is aggregated from two partial target views using their relative motion.
We evaluate our model on KITTI~\cite{Geiger2012CVPR}, NuScenes\cite{caesar2020nuscenes} and Waymo Open Dataset (WOD)~\cite{sun2020scalability}, where NuScenes and WOD cover a wide variety of real-world environments and are challenging for their dense traffics. 
The experiment results demonstrate that our model outperforms the existing methods by a large margin while running faster than the previous top-performer~\cite{zheng2021box}. 
Besides, the performance gap becomes even more significant when more distractors exist in the scenes. 
Furthermore, we demonstrate that our method can directly benefit from appearance matching when integrated with existing methods.

Apart from $M^2$-Track impressive performance under the fully-supervised setting, its motion-centric nature also helps to ease the difficulties of insufficient annotations: 
\textbf{1)} Instead of appearance, it models the relative target motion, which can be easily synthesized from only a few labeled data;
\textbf{2)} It does not rely on a ``cropped" template to do tracking due to its matching-free nature, enabling natural gradient flow for the cycle training (\ie, tracking targets forward-backward symmetrically in time without the requirements of labels at intermediate frames).
Based on the above observations, we address the challenges of \textbf{semi}-supervised LiDAR SOT with a \textbf{m}otion-centric framework, which we dub \textbf{SEMIM}.
Inspired by previous works~\cite{zhao2020sess,wang20213dioumatch}, SEMIM trains $M^2$-Track under SSL setting using pseudo labels.
Considering the motion-centric nature, we first design a pseudo-label-based motion augmentation to automatically ``annotate" the unlabeled sequences by putting the annotated objects into unseen backgrounds.
Although pseudo labels are noisy, they provide a reasonable approximation of the ground truth target locations.
Treating the pseudo labels as anchors, we transform the annotated objects into unlabeled frames using a \textit{delete-cut-paste} operation (see Fig.~\ref{fig:aug}).
Compared to random pasting~\cite{yan2018second}, our strategy remains appropriate target motion, and reduces undesired object collision across frames.
Thanks to the matching-free nature, we incorporate a \textit{cycle-consistent loss} in SSL training, which complements the supervision signal of pseudo labels.
On the one hand, the cycle-consistent loss ensures the tracking is consistent when input frame pairs are flipped along time, reducing the negative influence of noisy pseudo labels.
On the other hand, using pseudo labels prevent the cycle-consistent loss from falling into the degenerate solution (\ie, the ``zero-motion").
The experiments on the KITTI~\cite{Geiger2012CVPR} dataset show that SEMIM achieves exciting results when even trained with fewer than 400 frames, outperforming the baseline by $>$ 20\% in precision.
For car objects, SEMIM only needs 26\% labels to perform on par with the 100\%-supervised baseline and even surpasses it when more labels are available.
Only using labeled data from KITTI~\cite{Geiger2012CVPR}, we further test SEMIM's tracking performance on Waymo Open Dataset~\cite{sun2020scalability}, which shows SEMIM's great potential for offline auto-labeling and unsupervised domain adaptation.

In summary, our main contributions are as follows:
\textbf{1)} A novel motion-centric paradigm for real-time LiDAR SOT, which is free of appearance matching. %
\textbf{2)} A specific second-stage pipeline named {{M$^2$-Track}} that leverages the motion-modeling and motion-assisted shape completion. 
\textbf{3)} State-of-the-art online tracking performance with significant improvement on three widely adopted datasets (\ie, KITTI, NuScenes and Waymo Open Dataset).
\textbf{4)} Formulation of the semi-supervised LiDAR SOT problem for the first time and impressive SSL performance.

This paper is an extension of our conference paper published in CVPR 2022~\cite{zheng2022beyond}.
In this version, 
\textbf{1)} we extend the motion-centric tracker to handle SSL in LiDAR SOT, which is left unexplored in previous literature (Sec.~\ref{sec:semi}).
\textbf{2)}~Without any modification to the main network architecture, we significantly boost the performance of the original $M^2$-Track~\cite{zheng2022beyond} by designing an improved motion augmentation (Sec.~\ref{sec:aug}) and multi-frame ensembling (Sec.~\ref{sec:multi}).
\textbf{3)} We present a more comprehensive analysis of the motion-centric paradigm by implementing a vanilla motion-centric tracker (Sec.~\ref{sec:vanilla}) and including  extra ablation study (Sec.~\ref{sec:fully_exp}).


%

\section{Related Work}
\label{sec:related_work}

\noindent\textbf{Single Object Tracking.}
A majority of approaches are built for camera systems and take 2D RGB images as input~\cite{xu2020siamfc++,li2019siamrpn++,wang2021transformer,zhu2018distractor,bhat2019learning,li2018high}. Although achieving promising results, they face great challenges when dealing with low light conditions or textureless objects.
In contrast, LiDARs are insensitive to texture and robust to light variations, making them a suitable complement to cameras. 
This inspires a new trend of SOT approaches~\cite{Giancola_2019_CVPR,qi2020p2b,zheng2021box,zarzar2019efficient,shan2021ptt,fang20203d} which operate on 3D LiDAR point clouds.
These 3D methods all inherit the Siamese paradigm based on appearance matching.
As a pioneer, \cite{Giancola_2019_CVPR} uses the Kalman filter to heuristically sample a bunch of target proposals, which are then compared with the target template based on their feature similarities. The proposal which has the highest similarity with the target template is selected as the tracking result.
Since heuristic sampling is time-consuming and inhibits end-to-end training, \cite{zarzar2019efficient,qi2020p2b} propose to use a Region Proposal Network (RPN) to generate high-quality target proposals efficiently.
Unlike \cite{zarzar2019efficient} which uses an off-the-shelf 2D RPN operating on bird’s eye view (BEV), \cite{qi2020p2b} adapts SiamRPN~\cite{li2018high} to 3D point clouds by integrating a point-wise correlation operator with a point-based RPN~\cite{qi2019deep}.
The promising improvement brought by \cite{qi2020p2b} inspires a series of follow-up works~\cite{zheng2021box,shan2021ptt,fang20203d,hui20213d}. 
They focus on either improving the point-wise correlation operator~\cite{zheng2021box} by feature enhancement,
or refining the point-based RPN~\cite{fang20203d,shan2021ptt,hui20213d} with more sophisticated structures.

\noindent The appearance matching achieves excellent success in 2D SOT because images provide rich texture, which helps the model distinguish the target from its surrounding.
However, LiDAR point clouds only contain geometric appearances that lack texture information. Besides, objects in LiDAR sweeps are usually sparse and incomplete. These bring considerable ambiguities which hinder effective appearance matching.
Recently, some publications~\cite{lan2022temporal,gao2023spatio} have made advancements in Siamese tracking by leveraging temporal information from multiple frames to enhance the target template.
However, they still suffer from matching ambiguities since they cannot explicitly model the target motion.
Unlike existing 3D approaches, our work no more uses any appearance matching. Instead, we examine a new motion-centric paradigm and show its great potential for LiDAR SOT.

\noindent\textbf{3D Multi-object Tracking / Detection.}
In parallel with 3D SOT, 3D multi-object tracking (MOT) focuses on tracking multiple objects simultaneously. Unlike SOT where the user can specify a target of interest, MOT relies on an independent detector~\cite{shi2019pointrcnn,qi2019deep,yan2018second} to extract potential targets, which obstructs its application for unfamiliar objects (categories unknown by the detector).
Current 3D MOT methods predominantly follow the ``tracking-by-detection" paradigm, which first detects objects at each frame and then heuristically associates detected BBoxes based on objects' motion or appearance~\cite{yin2021center,Weng2020_AB3DMOT,chiu2020probabilistic,liang2020pnpnet,wu20213d}. Recently, \cite{luo2021exploring} proposes to jointly perform detection and tracking by combining object detection and motion association into a unified pipeline.
In addition to using uni-modal approaches, there has been a recent emergence of methods that leverage both LiDAR and camera data to perform 3D MOT~\cite{liu2023bevfusion,wang2023camo,kim2021eagermot}. These multi-sensor fusion techniques offer increased robustness to occlusion and object misdetection, thanks to the redundancy provided by combining data from multiple sensors.
Our motion-centric tracker draws inspiration from the motion-based association in MOT. But unlike MOT, which applies motion estimation on detection results, our approach does not depend on any detector and can leverage the motion prediction to refine the target BBox further.

\noindent\textbf{Spatial-temporal Learning on Point Clouds.}
Our method utilizes spatial-temporal learning to infer relative motion from multiple frames.
Inspired by recent advances in natural language processing~\cite{sundermeyer2012lstm,chung2014empirical,vaswani2017attention}, there emerges methods that adapt LSTM~\cite{huang2020lstm}, GRU~\cite{yin2020lidar}, or Transformer~\cite{fan2021point} to model point cloud videos.
However, their heavy structures make them impractical to be integrated with other downstream tasks, especially for real-time applications.
Another trend forms a spatial-temporal (ST) point cloud by merging multiple point clouds with a temporal channel added to each point~\cite{rempe2020caspr,qi2021offboard,hu2020you}. Treating the temporal channel as an additional feature (like RGB or reflectance), one can process such an ST point cloud using any 3D backbones~\cite{qi2017pointnet++,qi2017pointnet} without structural modifications. We adopt this strategy to process successive frames for simplicity and efficiency.

\noindent\textbf{Semi-/Un-Supervised Learning in Point Clouds.}
The semi-/un- supervised learning has been studied in various point cloud tasks, such as object detection~\cite{zhao2020sess,wang20213dioumatch}, semantic segmentation~\cite{cheng2021sspc,Jiang_2021_ICCV} and scene flow estimation~\cite{mittal2020just,jin2022deformation}.
Most of these methods first use the supervision from labeled data (either given or manually synthesized) to generate pseudo labels for unlabeled data.
Guided by the pseudo labels, they then train their models on unlabeled data by applying consistent regularization with the help of the teacher-student framework~\cite{zhao2020sess,wang20213dioumatch} or contrastive learning~\cite{Jiang_2021_ICCV}.
However, these techniques cannot be directly adopted to handle LiDAR SOT, which is dominated by matching-based approaches.
Current LiDAR SOT approaches heavily rely on annotated target bounding boxes (bboxes) to generate search regions around the targets as the training samples. For the tracking task, it is not practical to search for the target in a whole LiDAR point cloud, because most of the regions in the scene are redundant and may distract correct tracking. 
For this reason, poor pseudo labels greatly amplify the training noise for the tracker because they may lead to meaningless search regions.
In our work, we relieve this issue by correcting the pseduo labels with a \textit{delete-cut-paste} operation, which not only reduces the noise of pseudo labels but also augments target motion in favor of the motion-centric tracker.

\noindent\textbf{Unsupervised Learning in 2D Tracking.}
To train a 2D tracker without labels, some methods construct template-search pairs from still frames to explore the self-supervised signals of the videos in the spatial dimension~\cite{li2021crop,sio2020s2siamfc}.
However, these approaches are sensitive to appearance change, which is common in LiDAR data.
And the augmentation strategies of these methods are only applicable to 2D images, which have a very limited range and regular structures.
Another trend relies on the cycle consistency in the temporal dimension~\cite{yuan2020self,shen2022unsupervised}.
Since matching-based methods select target templates with an indifferentiable cropping operation, the cycle tracking errors are not correctly penalized.
For 2D trackers, this could be alleviated with a more sophisticated region-masking operation~\cite{shen2022unsupervised}.
By contrast, the matching-free nature of the motion-centric paradigm naturally helps us avoid this problem.
\section{Motion-Centric LiDAR SOT}
\label{sec:method}
\subsection{Problem Statement}\label{sec:problem}
Given the initial state of a target, our goal is to localize the target in each frame of an input sequence of a dynamic 3D scene. 
The frame at timestamp $t$ is a LiDAR point clouds ${\mathcal{P}_t \in \mathbb{R}^{N_t \times 3}}$ with $N_t$ points and $3$ channels, where the point channels encode the $xyz$ global coordinate.
The initial state of the target is given as its 3D BBox at the first frame $\mathcal{P}_1$. A 3D BBox $\mathcal{B}_t \in \mathbb{R}^7$ is parameterized by its center ($xyz$ coordinate), orientation (heading angle $\theta$ around the $up$-axis), and size (width, length, and height).  For the tracking task, we further assume that the size of a target remains unchanged across frames even for non-rigid objects (for a non-rigid object, its BBox size is defined by its maximum extent in the scene).
For each frame $\mathcal{P}_t$, the tracker outputs the amodal 3D BBox of the target with access to only history frames $\{\mathcal{P}_i\}_{i=1}^t$. 

\subsection{Motion-centric Paradigm}\label{sec:paradigm}
Given an input LiDAR sequence and the 3D BBox of the target in the first frame, the motion-centric tracker aims to localize the target frame by frame using explicit motion modeling. 
Unlike previous LiDAR SOT methods, which track targets via appearance matching, a motion-centric tracker predicts the "\textbf{relative target motion (RTM)}" and localizes the target by motion transformation.
RTM is a rigid-body transformation that is defined between two target bounding boxes.
Since objects of interest are always aligned with the ground, we only consider 4DOF RTM, which comprises a translation offset $(\Delta x,\Delta y,\Delta z)$ and a yaw offset $\Delta \theta$.
Given a known target state at time $t$ (the target bbox $\mathcal{B}_{t}$ in frame $\mathcal{P}_{t}$) and a new incoming frame $\mathcal{P}_{t'}$ at time $t'$ ($t' > t$), 
a motion-centric tracker first predicts the RTM of the given target between $\mathcal{P}_{t}$ and $\mathcal{P}_{t'}$, and then obtains the target bbox $\mathcal{B}_{t'}$ at $\mathcal{P}_{t'}$ via rigid-body transformation. 
The whole process can be formulated as a function $\mathcal{F}$:\
\begin{equation}\label{eq:tracker}
  \begin{aligned}
  \mathcal{F}&: \mathbb{R}^{N_{t'} \times C} \times \mathbb{R}^{N_{t} \times C} \times \mathbb{R}^7\mapsto \mathbb{R}^7, \\
  \mathcal{G}&: \mathbb{R}^{N_{t'} \times C} \times \mathbb{R}^{N_{t} \times C} \times \mathbb{R}^7\mapsto \mathbb{R}^4, \\
  \mathcal{B}_{t'} = 
  \mathcal{F}&(\mathcal{P}_{t'},\mathcal{P}_{t},\mathcal{B}_{t}) = 
  \text{Transform}(\mathcal{B}_{t}, \mathcal{G}(\mathcal{P}_{t'},\mathcal{P}_{t},\mathcal{B}_{t})),
  \end{aligned}
  \end{equation}
where $\mathcal{G}$ predicts the 4DOF RTM and  ``Transform'' applies rigid-body transformation on $\mathcal{B}_{t}$ using the predicted RTM.

\subsection{$M^2$-Track: Motion-Centric Tracking Pipeline}\label{sec:pipeline}
Following the motion-centric paradigm, we design a two-stage motion-centric tracking pipeline $M^2$-Track (illustrated in Fig.~\ref{fig:pipeline1}).
$M^2$-Track first coarsely localizes the target through target segmentation and \textbf{m}otion transformation at the $1^{st}$ stage, and then refines the BBox at the $2^{nd}$ stage using \textbf{m}otion-assisted shape completion.
More details of each module are given as follows.

 \begin{figure}[t]
   \centering
    \includegraphics[width=0.95\linewidth]{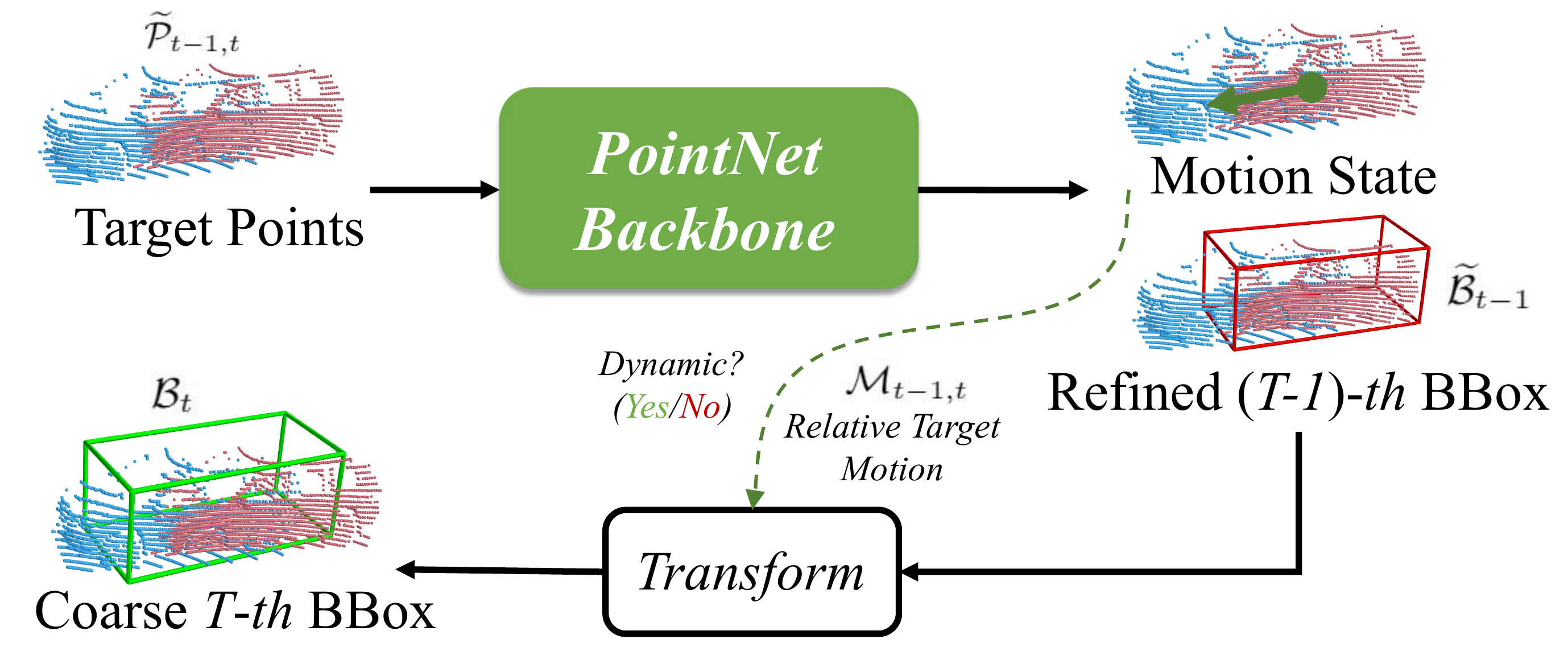}
 
    \caption{
      \textbf{Stage I.}
      Taking in the segmented target points $\mathcal{\widetilde{P}}_{t-1,t}$ and the target BBox $\mathcal{{B}}_{t-1}$ at the previous frame, the model outputs a relative target motion state (including an RTM $\mathcal{M}_{t-1,t}$ and 2D binary motion state logits), a refined target BBox $\mathcal{\widetilde{B}}_{t-1}$ at the previous frame, and a coarse target BBox $\mathcal{B}_{t}$ at the current frame.
      }
    \label{fig:stage1}
 \end{figure}

\noindent\textbf{Target segmentation with spatial-temporal learning} 

\noindent
To learn the relative target motion, we first need to segment the target points from their surrounding. 
Taking as inputs two consecutive frames $\mathcal{P}_{t}$ and $\mathcal{P}_{t-1}$ together with the target BBox $\mathcal{B}_{t-1}$,  we do this by exploiting the spatial-temporal relation between the two frames (illustrated in the first part of Fig.~\ref{fig:pipeline1}).
Similar to \cite{rempe2020caspr,piergiovanni20214d}, we construct a spatial-temporal point cloud $\mathcal{P}_{t-1,t} \in \mathbb{R}^{(N_{t-1} + N_{t}) \times 4} = \{p_i = (x_i,y_i,z_i,t_i)\}_{i=1}^{N_{t-1}+N_t}$ from $\mathcal{P}_{t-1}$ and $\mathcal{P}_{t}$ by adding a temporal channel to each point and then merging them together.
Since there are multiple objects in a scene, we have to specify the target of interest according to $\mathcal{B}_{t-1}$.
To this end, we create a prior-targetness map $\mathcal{S}_{t-1,t} \in \mathbb{R}^{N_{t-1} + N_t}$ to indicate target location in $\mathcal{P}_{t-1,t}$, where $s_i \in \mathcal{S}_{t-1,t}$ is defined as:
\begin{equation}
  \begin{aligned}
  s_i = 
  \left\{
    \begin{matrix}
      0& \text{ if } p_i \text{ is in } \mathcal{P}_{t-1} \text{ and } p_i \text{ is not in }\mathcal{B}_{t-1}\\
      1& \text{ if } p_i \text{ is in } \mathcal{P}_{t-1} \text{ and } p_i \text{ is in }\mathcal{B}_{t-1}\\
      0.5& \text{ if } p_i \text{ is in } \mathcal{P}_{t}
    \end{matrix}\right.
  \end{aligned}
\end{equation}
Intuitively, one can regard $s_i$ as the prior-confidence of $p_i$ being a target point. For a point in $\mathcal{P}_{t-1}$, we set its confidence according to its location with respect to $\mathcal{B}_{t-1}$. Since the target state in $\mathcal{P}_{t}$ is unknown, we set a median score 0.5 for each point in $\mathcal{P}_{t}$.
Note that $\mathcal{S}_{t-1,t}$ is not 100\% correct for points in $\mathcal{P}_{t-1}$ since $\mathcal{B}_{t-1}$ could be the previous output by the tracker.
After that, we form a 5D point cloud by concatenating $\mathcal{P}_{t-1,t}$ and $\mathcal{S}_{t-1,t}$ along the channel axis, and use a PointNet-based segmentation network~\cite{qi2017pointnet} to obtain the target mask, which is finally used to extract a spatial-temporal target point cloud $\mathcal{\widetilde{P}}_{t-1,t} \in \mathbb{R}^{({M_{t-1} + M_{t})} \times 4}$, where $M_{t-1}$ and $M_{t}$ are the numbers of target points in frame ($t-1$) and $t$ respectively.

 \begin{figure}[t]
  \centering
   \includegraphics[width=0.95\linewidth]{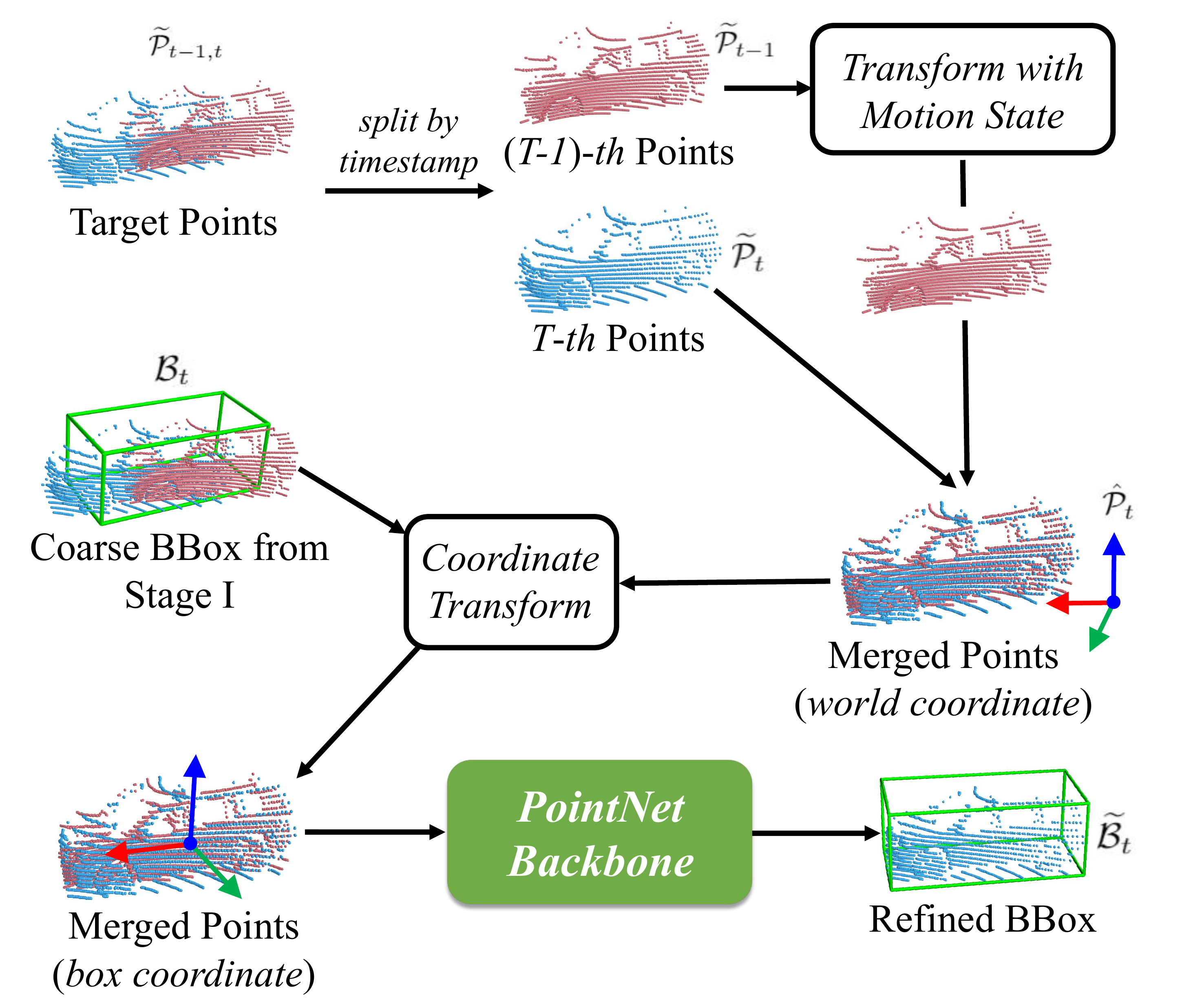}

   \caption{
    \textbf{Stage II.}
    Taking the segmented target points $\mathcal{\widetilde{P}}_{t-1,t}$ and the coarse target BBox $\mathcal{B}_{t}$ as inputs, the model regresses a refined target BBox $\mathcal{\widetilde{B}}_{t}$ on a denser point cloud, which is merged from two partial target point clouds according to their relative motion state.
     }
   \label{fig:stage2}
\end{figure}

\noindent\textbf{Stage I: Motion-Centric BBox prediction}\\
\noindent As shown in Fig.~\ref{fig:stage1}, we encode the spatial-temporal target point clouds $\mathcal{\widetilde{P}}_{t-1,t}$ into an embedding using a PointNet encoder. A multi-layer perceptron (MLP) is applied on top of the embedding to obtain the motion state of the target, which includes a 4D RTM $\mathcal{M}_{t-1,t}$ and 2D binary classification logits indicating whether the target is dynamic. To reduce accumulation errors while performing frame-by-frame tracking, we generate a refined previous target BBox $\mathcal{\widetilde{B}}_{t-1}$ by predicting its RTM with respect to $\mathcal{B}_{t-1}$ through another MLP (More details are presented in the supplementary). Finally, we can get the current target BBox $\mathcal{B}_{t}$ by transforming $\mathcal{\widetilde{B}}_{t-1}$ using $\mathcal{M}_{t-1,t}$ if the target is classified as dynamic. Otherwise, we simply set $\mathcal{B}_{t}$ as $\mathcal{\widetilde{B}}_{t-1}$.

\noindent\textbf{Stage II: BBox refinement with shape completion}\\
\noindent Inspired by two-stage detection networks~\cite{shi2019pointrcnn,shi2020points}, we improve the quality of the $1^{st}$-stage BBox $\mathcal{B}_{t}$ by additionally regressing a relative offset, which can be regarded as an RTM between $\mathcal{B}_{t}$ and the refined BBox $\mathcal{\widetilde{B}}_{t}$ (illustrated in Fig.~\ref{fig:stage2}).
Unlike detection networks, we refine the BBox via a novel \textbf{motion-assisted shape completion} strategy.
Due to self-occlusion and sensor movements, LiDAR point clouds suffer from great incompleteness, which hinders precise BBox regression. To mitigate this, we form a denser target point cloud by using the predicted motion state to aggregate the targets from two successive frames.
According to the temporal channel, two target point clouds $\mathcal{\widetilde{P}}_{t-1} \in \mathbb{R}^{M_{t-1} \times 3}$ and $\mathcal{\widetilde{P}}_{t} \in \mathbb{R}^{M_t \times 3}$ from different timestamps are extracted from $\mathcal{\widetilde{P}}_{t-1,t}$.
Based on the motion state, we transform $\mathcal{\widetilde{P}}_{t-1}$ to the current timestamp using $\mathcal{M}_{t-1,t}$ if the target is dynamic. The transformed point cloud (identical as $\mathcal{\widetilde{P}}_{t-1}$ if the target is static) is merged with  $\mathcal{\widetilde{P}}_{t}$ to form a denser point cloud $\mathcal{\hat{P}}_{t} \in \mathbb{R}^{(M_{t-1} + M_t) \times 3}$. Similar to \cite{shi2019pointrcnn,qi2021offboard}, we transform $\mathcal{\hat{P}}_{t}$ from the world coordinate system to the canonical coordinate system defined by $\mathcal{B}_{t}$.
We apply a PointNet on the canonical $\mathcal{\hat{P}}_{t}$ to regress another RTM with respect to $\mathcal{B}_{t}$. Finally, the refined target BBox $\mathcal{\widetilde{B}}_{t}$ is obtained by applying a rigid-body transformation on $\mathcal{B}_{t}$ using the regressed RTM.

\begin{figure}[t]
	\begin{centering}
		\includegraphics[width=0.9\linewidth]{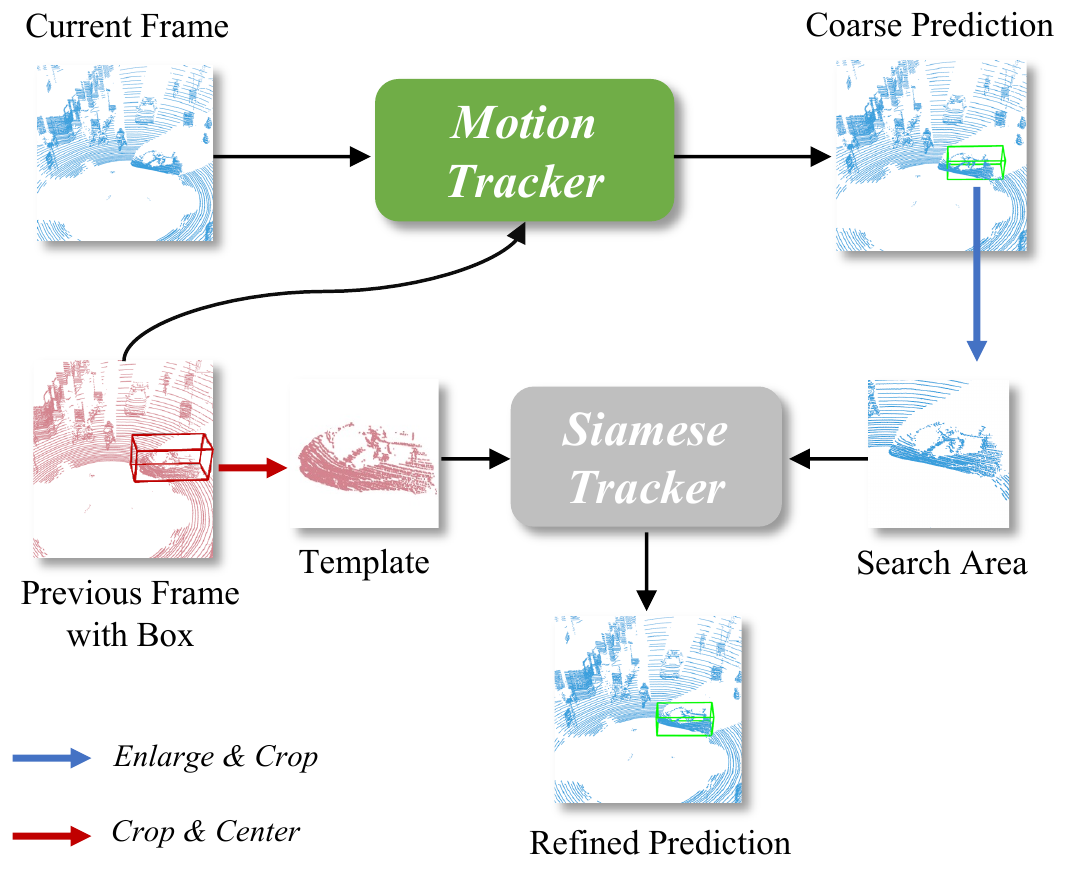}
		\caption{\textbf{Combining Motion- and Matching-based Models.} We utilize a Siamese tracker to search for the target in a small neighborhood around the output of the motion-based tracker.}
		\label{fig:motion_matching}
	\end{centering}	
\end{figure}

\subsection{Box-aware Feature Enhancement}\label{sec:box_aware}
As shown in \cite{zheng2021box}, LiDAR SOT directly benefits from the part-aware and size-aware information, which can be depicted by point-to-box relation.
Inspired by \cite{zheng2021box}, we construct a distance map $\mathcal{C}_{t-1} \in \mathbb{R}^{N_{t-1} \times 9}$ by computing the pairwise Euclidean distance between $\mathcal{P}_{t-1}$ and 9 key points of $\mathcal{B}_{t-1}$ (eight corners and one center arranged in a predefined order with respect to the canonical box coordinate system). 
After that, we extend $\mathcal{C}_{t-1}$ to size $(N_{t-1} + N_t) \times 9$ with zero-padding (for points in $\mathcal{P}_{t}$) and
additionally concatenate it with $\mathcal{P}_{t-1,t}$ and $\mathcal{S}_{t-1,t}$.
The overall box-aware features are then sent to the PointNet segmentation network to obtain better target segmentation.
Besides outputting the segmentation results, we further predict the box-aware feature for each point, which is supervised by its corresponding ground truth. 
During Stage I and Stage II, we also concatenate the predicted box-aware features along with the point coordinates to construct the inputs to PointNet.

\subsection{Multi-frame Ensembling}\label{sec:multi}
Although $M^2$-Track always takes two consecutive frames (${t-1}~\text{and} ~t$) as inputs in the formulation, it can also learn RTMs from frame pairs with time intervals larger than 1, which inspired us to improve the tracking results using multiple frames.
Assuming that we already know the target tracklet $\{\mathcal{B}_{t-n}\}_{n=1}^{N}$ from time $(t-N)$ to $(t-1)$, we can obtain $N$ target proposals in time $t$ by applying $M^2$-Track on different time intervals:

\begin{equation}\label{eq:multi}
  \begin{aligned}
    \text{Proposals} = \{\mathcal{F}(\mathcal{P}_t,\mathcal{P}_{t-n},\mathcal{B}_{t-n})\}_{n=1}^N
\end{aligned}
\end{equation}
where $\mathcal{F}$ is the tracker function defined in Eqn.~\eqref{eq:tracker}, which is implemented as an $M^2$-Track.

After obtaining $N$ proposals, we check the number of points in each proposal and select the proposal with the maximum number of points as the final prediction.

\subsection{M-Vanilla: Proof-of-concept Implementation}\label{sec:vanilla}
To showcase the effectiveness of the motion-centric paradigm, we also introduce a proof-of-concept tracker named M-Vanilla, which strictly adopts an architecture depicted in the lower part of Fig.~\ref{fig:fig1}.
M-Vanilla closely resembles Stage I in $M^2$-Track, with the distinction that it takes both foreground and background points as inputs and omits the motion classification branch.
Given two consecutive frames $\mathcal{P}_{t}$ and $\mathcal{P}_{t-1}$ together with the previous target BBox $\mathcal{B}_{t-1}$, we first construct a spatial-temporal point cloud as described in Sec.~\ref{sec:pipeline}, and further enhance this spatial-temporal point cloud with the box-aware features as described in Sec.~\ref{sec:box_aware}.
After that, we directly pass the spatial-temporal point cloud through a mini-PointNet~\cite{qi2017pointnet} to obtain the 4D RTM without any target segmentation.
Finally, M-Vanilla transforms the previous target BBox $\mathcal{B}_{t-1}$ using the predicted RTM to obtain the current target BBox $\mathcal{B}_{t}$.

\subsection{Combining Motion- and Matching-based Models}\label{sec:motion_matching}
While motion modeling aids in reducing appearance ambiguities, appearance matching plays a crucial role in capturing fine-grained patterns necessary for achieving highly-precise target localization.
To achieve both robust and accurate tracking, we further investigate a simple combination of motion- and matching-based models as illustrated in Fig.~\ref{fig:motion_matching}.
Given two consecutive frames $\mathcal{P}_{t}$ and $\mathcal{P}_{t-1}$ together with the previous target BBox $\mathcal{B}_{t-1}$, we first employ a motion-centric tracker to obtain a target BBox $\mathcal{B}_{motion}$ at time $t$.
Subsequently, we enlarge $\mathcal{B}_{motion}$ by a small margin and collect points inside it to generate a search area $\mathcal{P}_{search}$, which is then normalized to the object coordinate system defined by $\mathcal{B}_{motion}$.
Simultaneously, we use the previous target BBox $\mathcal{B}_{t-1}$ to crop the previous frame $\mathcal{P}_{t-1}$ and center the cropped point cloud to create a target template $\mathcal{P}_{temp}$.
The search area $\mathcal{P}_{search}$ and the target template $\mathcal{P}_{temp}$ are then sent to a Siamese tracker to refine $\mathcal{B}_{motion}$ by appearance matching, yielding the final target BBox $\mathcal{B}_{final}$.

In this setting, the motion tracker provides a reliable initialization of the target location, reducing distractors and thus ensuring more accurate appearance matching in the Siamese tracker.
Please refer to Sec.~\ref{sec:motion_matching_exp} for related experiments.

\subsection{Improved Motion Augmentation}\label{sec:aug}
To encourage the model to learn various motions during the training, we randomly flip both targets' points and BBoxes in their horizontal axes and rotate them around their $up$-axes by \textit{Uniform}[$-10^{\circ}$,$10^{\circ}$]. We also randomly translate the targets by offsets drawn from \textit{Uniform} [-0.3, 0.3] meter.
In our conference paper~\cite{zheng2022beyond}, the above augmentation is applied to \textit{all} training samples.
While synthesized motion reduces the motion bias in the data and thus improves the model's generalizability, realistic motion in the original data is also valuable since it depicts real object movement in natural scenes. 
However, realistic motion, especially the static one, is submerged by such an augmentation.
We alleviate this problem with a coin-flip test, where we use augmented data with probability $p \le 1$ while keeping the original one with chance $1-p$.
In our experiments, $p$ is empirically set to 0.5.
Though simple, such a coin-flip test significantly boosts the performance of $M^2$-Track (Tab. ~\ref{tab:augmentation}).
Inspired by~\cite{mittal2020just}, we further augment the training data by flipping point cloud sequences along the temporal dimension, which avoids biasing the model in favor of only the forward motion.
Because it doubles the motion space, such a temporal-flipping strategy is very helpful when the labeled data is very limited (\eg, under the semi-supervised setting). 

\subsection{Implementation Details}\label{sec:impl_detail}
\noindent\textbf{Loss functions.}
The loss function contains classification losses and regression losses and is defined as:
\begin{equation}\label{eq:loss_full}
  \begin{aligned}
    \mathcal{L}_m = &\lambda_1 \mathcal{L}_{\text{cls\_target}} + \lambda_2 \mathcal{L}_{\text{cls\_motion}} + \lambda_3 \mathcal{L}_{\text{reg\_box\_aware}} + \\
    &\lambda_4 (\mathcal{L}_{\text{reg\_motion}} + \mathcal{L}_{\text{reg\_refine\_prev}} + \mathcal{L}_{\text{reg\_1st}} + \mathcal{L}_{\text{reg\_2nd}})
\end{aligned}
\end{equation}
$\mathcal{L}_{\text{cls\_target}}$ and $\mathcal{L}_{\text{cls\_motion}}$ are standard cross-entropy losses for target segmentation and motion state classification at the $1^{st}$-stage (Points are considered as the target if they are inside the target BBoxes; A target is regarded as dynamic if its center moves more than 0.15 meter between two frames).
All regression losses are defined as the Huber loss~\cite{ren2015faster} between the predicted and ground-truth RTMs (inferred from ground-truth target BBoxes),
where $\mathcal{L}_{\text{reg\_motion}}$ is for the RTM between targets in the two frames; $\mathcal{L}_{\text{reg\_refine\_prev}}$ is for the RTM between the predicted and the ground-truth BBoxes at timestamp $(t-1)$; $\mathcal{L}_{\text{reg\_1st}}$ / $\mathcal{L}_{\text{reg\_2nd}}$ is for the RTM between the $1^{st}$ / $2^{nd}$-stage and ground-truth BBoxes;
$\mathcal{L}_{\text{reg\_box\_aware}}$ is the regression loss for the predicted 9D box-aware features.
We empirically set $\lambda_1 = \lambda_2 = 0.1$ and $\lambda_3 = \lambda_4 = 1$.

\noindent\textbf{Network input.}
Since SOT only takes care of one target in a scene, we only need to consider a subregion where the target may appear. 
Basically, the input to $M^2$-Track is a frame pairs.
For two consecutive frames at $(t-1)$ and $t$ timestamp, we choose the subregion by enlarging the target BBox at $(t-1)$ timestamp by 2 meters. We then sample 1024 points from the subregion respectively at $(t-1)$ and $t$ timestamp to form $\mathcal{P}_{t-1}$ and $\mathcal{P}_{t}$. To simulate testing errors during the training, we feed the model a perturbed BBox by adding a slight random shift to the ground-truth target BBox at $(t-1)$ timestamp.
To achieve multi-frame ensembling, we use a frame and its previous $N \ge 2$ frames to build each input, making the network aware of RTMs with various time intervals.

\noindent\textbf{Training \& inference.} We train our models using the Adam optimizer with batch size 256 and an initial learning rate 0.001, which is decayed by 10 times every 20 epochs. The training takes $\sim 4$ hours to converge on a V100 GPU for the KITTI Cars. During the inference, the model tracks a target frame-by-frame in a point cloud sequence given the target BBox at the first frame.

\begin{figure*}[t]
	\begin{centering}
		\includegraphics[width=\textwidth]{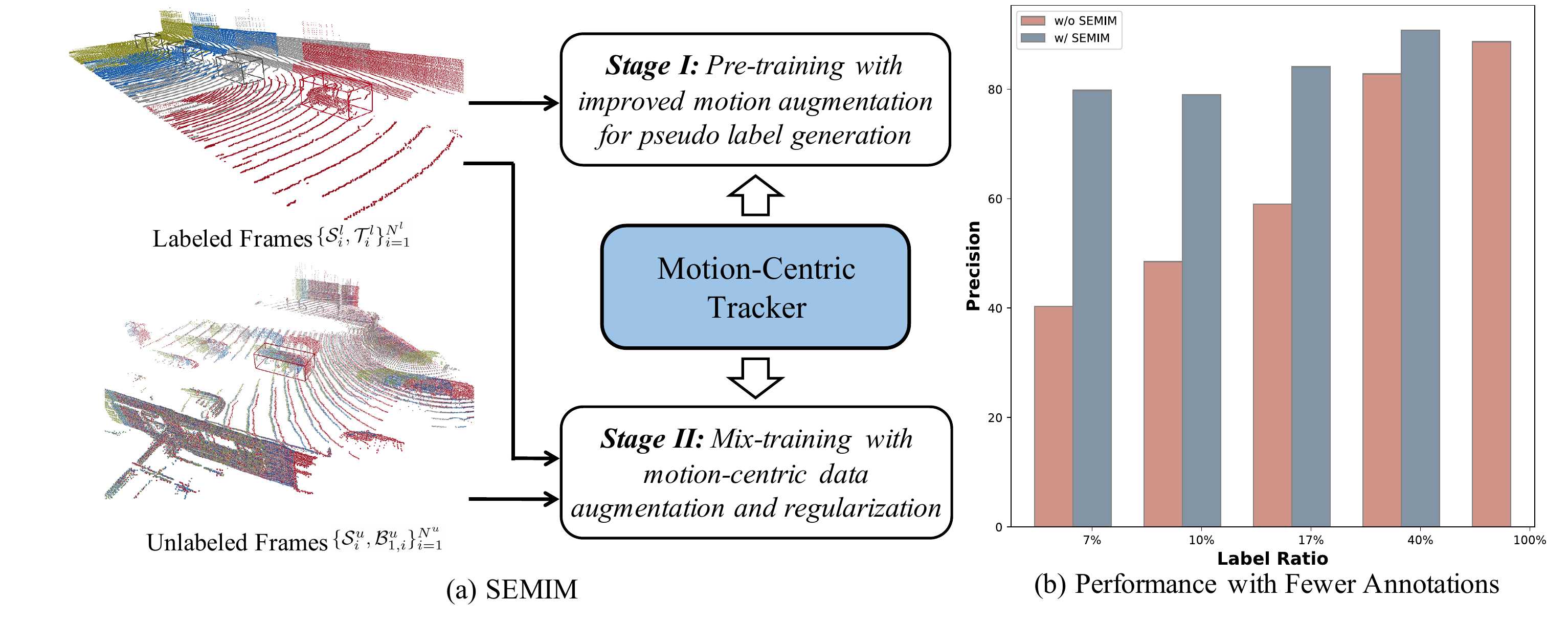}
		\caption{(a) The workflow of SEMIM, which performs semi-supervised learning based on pseudo-labeling. (b) On KITTI pedestrians, $M^2$-Track's performance drops drastically under low label ratios. SEMIM significantly boosts the performance when labeled data is insufficient.
		}
		\label{fig:semi_fig1}
	\end{centering}	
	
\end{figure*}

\begin{figure}[t]
	\centering
	\includegraphics[width=1\linewidth]{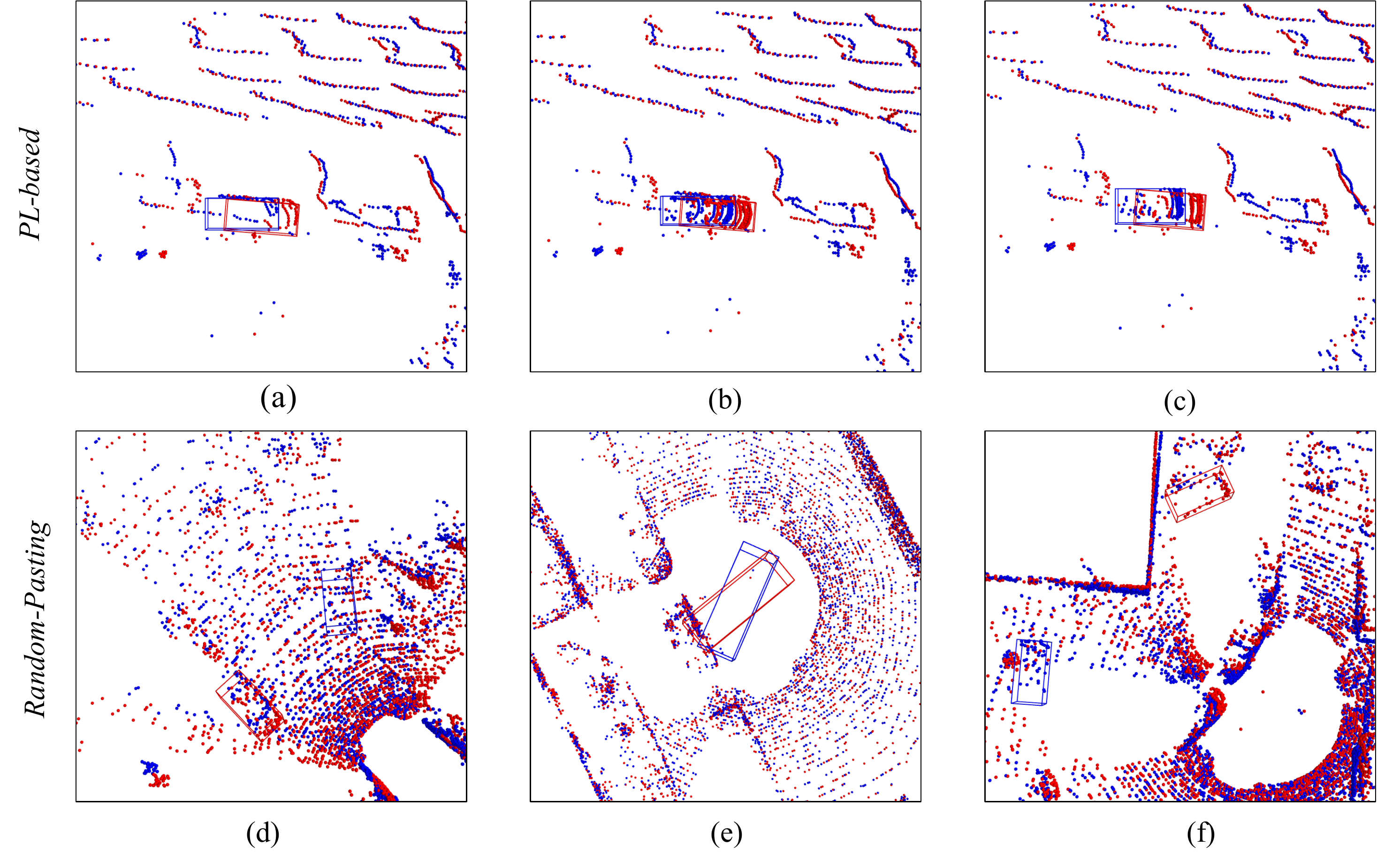}
	\caption{\textbf{(a):} An unlabeled frame pair with pseudo labels (PL); \textbf{(b)(c):} Results of \textit{delete-cut-paste} on (a) with $\gamma =1$ and $\gamma=1.25$; \textbf{(d)(e)(f):} Random pasting causes unrealistic motion and object collision. \textcolor{red}{Red} and \textcolor{blue}{Blue} denote frames at time $t$ and $t\texttt{-}1$, respectively.}
	\label{fig:aug}
\end{figure}

\section{Semi-supervised LiDAR SOT}\label{sec:semi}
\subsection{Semi-supervised Setting}
In the semi-supervised learning (SSL) setting, we have access to a set of labeled sequences $\{\mathcal{S}_i^l, \mathcal{T}_i^l\}_{i=1}^{N^l}$, where $\mathcal{T}_i^l = \{\mathcal{B}^l_{t,i} \}_{t=1}^{T}$ is the ground truth tracklet for the $i$-th labeled target and $\mathcal{S}_i^l = \{\mathcal{P}^l_{t,i} \}_{t=1}^{T}$ is the corresponding point cloud sequence.
Along with labeled data, we have a set of unlabeled sequences $\{\mathcal{S}_i^u, \mathcal{B}^u_{1,i}\}_{i=1}^{N^u}$, where only the first target bbox $\mathcal{B}^u_{1,i}$ is available for the $i$-th unlabeled sequence.
${N^l}$ and ${N^u}$ are the number of labeled and unlabeled sequences, respectively.
We aim at dealing with LiDAR SOT under a challenging condition, where ${N^l} \ll {N^u}$.

\begin{algorithm}[!t]
	\footnotesize
	\caption{Workflow of our pseudo-label-based motion augmentation}
	\label{alg:delete-cut-paste}
	\begin{algorithmic}[1]
		\renewcommand{\algorithmicrequire}{\textbf{Input:}}
		\renewcommand{\algorithmicensure}{\textbf{Output:}}
		\Require Labeled data $\{\mathcal{S}_i^l, \mathcal{T}_i^l\}_{i=1}^{N^l}$, an unlabeled frame pair $(\mathcal{P}_{t_1}^u, \mathcal{P}_{t_2}^u)$ and the pseudo target bboxes $\mathcal{B}_{t_1}^u, \mathcal{B}_{t_2}^u$, the scale factor $\gamma$ for deletion.
		%
		\State $i \gets RandInt(min=1,max=N^l)$
		\State $t_1' \gets RandInt(min=1,max=length(\mathcal{S}_i^l) - (t_2 - t_1))$
		\State $t_2' \gets t_1' + (t_2 - t_1)$
		\State $\mathcal{P}_{t_1'}^l,\mathcal{P}_{t_2'}^l \gets \mathcal{S}_i^l[t_1'],\mathcal{S}_i^l[t_2']$
		\State $\mathcal{B}_{t_1'}^l,\mathcal{B}_{t_2'}^l \gets \mathcal{T}_i^l[t_1'],\mathcal{T}_i^l[t_2']$
		\For{$j=1, 2$}
		\State $\mathcal{P}_{t_j}^u \gets Delete(\mathcal{P}_{t_j}^u, \mathcal{B}_{t_j}^u, \gamma)$, 
		\EndFor
		\For{$j=1, 2$}
		\State $\mathcal{O}_{t_j'} \gets CutAndCenter(\mathcal{P}_{t_j'}^l, \mathcal{B}_{t_j'}^l)$
		\EndFor
		\For{$j=1, 2$}
		\State $\mathcal{P}_{t_j}^u \gets TransformAndPaste(\mathcal{O}_{t_j'}, \mathcal{P}_{t_j}^u, \mathcal{B}_{t_j}^u)$
		\State $\mathcal{B}_{t_j}^u \gets Resize(\mathcal{B}_{t_j}^u, \mathcal{B}_{t_j'}^l)$
		\EndFor
		\Ensure $\mathcal{P}_{t_1}^u, \mathcal{P}_{t_2}^u$ ,$\mathcal{B}_{t_1}^u, \mathcal{B}_{t_2}^u$
	\end{algorithmic}
\end{algorithm}

\subsection{SEMIM}\label{sec:semim}
Leveraging the motion-centric tracker, SEMIM is a training pipeline that accomplishes semi-supervised LiDAR SOT.
As shown in Fig.~\ref{fig:semi_fig1}~(a), following the standard pseudo-labeling setup, SEMIM consists of two training stages:
\begin{enumerate*}[label=\textbf{(\arabic*)}]
  \item the pre-training stage,
  \item and the mixed-training stage.
\end{enumerate*}
For the pre-training stage, SEMIM pre-trains an $M^2$-Track on the labeled data $\{\mathcal{S}_i^l, \mathcal{T}_i^l\}_{i=1}^{N^l}$ using the fully-supervised loss.
After that, we run the pre-trained model on the unlabeled data $\{\mathcal{S}_i^u, \mathcal{B}^u_{1,i}\}_{i=1}^{N^u}$ to obtain pseudo labels $\{\mathcal{T}_i^u\}_{i=1}^{N^u}$.
In the mixed-training stage, SEMIM trains another $M^2$-Track from scratch on the mixed dataset $\{\mathcal{S}_i^l, \mathcal{T}_i^l\}_{i=1}^{N^l} \cup \{\mathcal{S}_i^u, \mathcal{T}_i^u\}_{i=1}^{N^u}$ using both ground-truth and pseudo labels.

In general, due to insufficient annotations, the pseudo labels generated by the pre-trained model are too noisy to provide reliable supervision signals.
Since LiDAR SOT approaches rely on the previous prediction to decide the search region of the incoming frame, the noise in pseudo labels is accumulated frame-by-frame, making the pseudo labels even more unreliable.
Thanks to the motion-centric nature, SEMIM addresses this with a pseudo-label-based motion augmentation and a cycle-consistent loss term.
\subsubsection{Pseudo-Label-Based Motion Augmentation}
%
As shown in Fig.~\ref{fig:aug} (a), although pseudo labels (in the form of bboxes) do not perfectly fit the underlying targets due to the inevitable noise, they provide a reasonable approximation of the RTMs, which are useful for training a motion-centric tracker.
Based on this observation, we propose a pseudo-label-based motion augmentation, which aims at
\begin{enumerate*}[label=\textbf{(\arabic*)}]
  \item reducing the misalignment between the pseudo labels and the targets of interest,
  \item and keeping the RTMs among pseudo labels.
\end{enumerate*}

\noindent\textbf{The workflow.}
We first introduce a \textit{delete-cut-paste} operation that puts an object from one frame to another while avoiding collision.
The \textit{delete-cut-paste} operation takes as input two points-box pairs: a destination pair $(\mathcal{P}_{dest},\mathcal{B}_{dest})$ and a source pair $(\mathcal{P}_{src},\mathcal{B}_{src})$.
It first \textit{delete} the points inside the destination bbox $\mathcal{B}_{dest}$ from the point cloud $\mathcal{P}_{dest}$, leaving only the background outside $\mathcal{B}_{dest}$.
During the implementation, we scale up the size of $\mathcal{B}_{dest}$ by $\gamma$ to include more potential target points for better deletion.
After that, its uses the source $\mathcal{B}_{src}$ bbox to \textit{cut} out the points in the source point cloud $\mathcal{P}_{src}$, obtaining an object point cloud $\mathcal{O}_{src}$ inside $\mathcal{B}_{src}$.
Finally, its \textit{pastes} $\mathcal{O}_{src}$ into $\mathcal{P}_{dest}$ in the position of $\mathcal{B}_{dest}$, which is then resized accordingly to fit the pasted object.
The whole workflow of our pseudo-label-based motion augmentation is given in Alg.~\ref{alg:delete-cut-paste}.
For any two frames in the same unlabeled sequence, we sample another two frames with the same interval from any labeled sequence and use the \textit{delete-cut-paste} operation to replace the points inside the pseudo boxes.
Note that the same interval is needed to ensure consistent object appearance after pasting.
During the training, we apply this augmentation to each unlabeled frame pair with probability $p \le 1$, enabling the network to see both original and augmented pseudo-labeled data.
\begin{figure}
	\centering
	\includegraphics[width=0.95\linewidth]{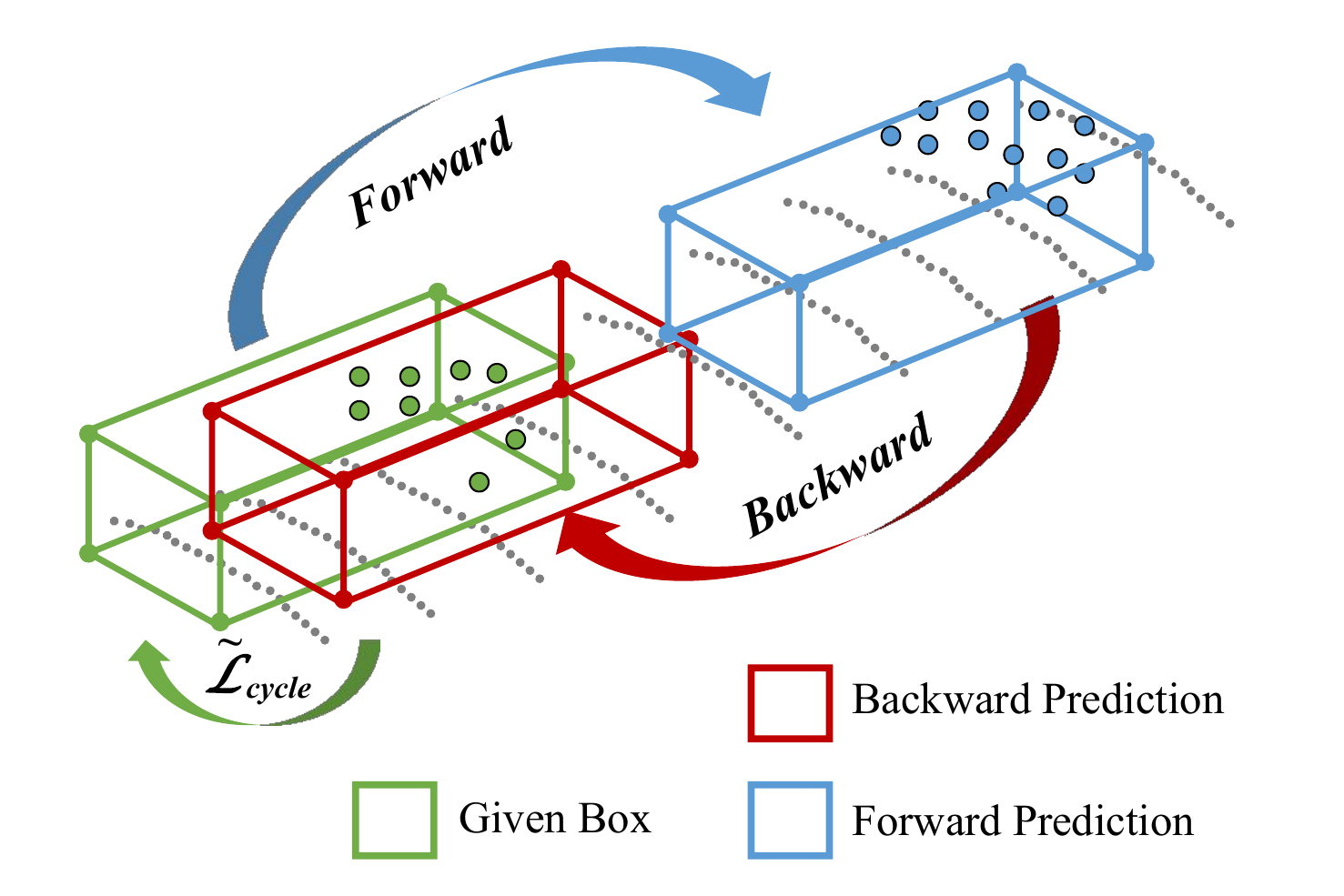}
	\caption{Self-supervised cycle-consistent loss, which is naturally differentiable due to the motion-centric nature.}
	\label{fig:cycle}
\end{figure}

\noindent\textbf{Comparison with similar techniques.}
Our pseudo-label-based motion augmentation relies on the \textit{delete-cut-paste} operation, which is similar to the random pasting that has been widely adopted in the 3D detection and 2D tracking tasks~\cite{yan2018second,li2021crop}.
Random pasting directly puts the object points and the corresponding bboxes into another frame.
Without using anchors, random pasting cannot handle object collision and is likely to put objects in meaningless areas since LiDAR frames have an extensive range (see Fig.~\ref{fig:aug}), which may harm the training.
In contrast to random pasting, \cite{fang2021lidar} proposed a rendering-based augmentation framework to avoid the collision.
Nevertheless, \cite{fang2021lidar} deals with each frame independently and thus the same object pasted in different frames may have unreal motion.
By contrast, our augmentation generates RTMs that are unseen in the labeled dataset by exploiting the pseudo boxes as the anchors to paste.
With the help of \textit{delete-cut-paste}, the object points and the pseudo boxes become perfectly aligned without collision.

\subsubsection{Semi-supervised Loss Design}
After generating pseudo labels by applying the pre-trained model on unlabeled sequences, we start training another model using combined data (labeled and unlabeled).
For labeled data, we directly apply the fully-supervised loss defined by Eqn.~\eqref{eq:loss_full} as the regularization.
Treating the pseudo labels as the ground truths, we can also apply the same loss term for the predictions of unlabeled data, which yields $\mathcal{L}_{forward}$ that penalizes the forward tracking errors:
\begin{equation}\label{eq:loss_forward}
  \begin{aligned}
\mathcal{L}_{forward} = 
\frac{1}{N^l} \sum\limits_{i=1}^{N^l} \mathcal{L}_{m}(\mathcal{\hat T}_i^l, \mathcal{T}_i^l) + 
\lambda\frac{1}{N^u} \sum\limits_{i=1}^{N^u} \mathcal{L}_{m}(\mathcal{\hat T}_i^u, \mathcal{T}_i^u),
\end{aligned}
\end{equation}
where $\mathcal{L}_{m}$ is the same loss term as Eqn.~\eqref{eq:loss_full}. 
$\mathcal{\hat T}_i^l$ / $\mathcal{\hat T}_i^u$ denotes the predicted tracklet of labeled / unlabeled data.
$\mathcal{T}_i^l$ and $\mathcal{T}_i^u$ stand for the ground truth and the pseudo tracklets, respectively.
$\lambda$ is the coefficient to control the weight of the unlabeled loss term.

As the number of unlabeled data is much larger than that of the labeled data, $\mathcal{L}_{forward}$ causes significant instability when individually applied in training because the pseudo labels are usually noisy.
%
%
To address this issue, we incorporate a self-supervised cycle-consistent loss $\mathcal{L}_{cycle}$, which is illustrated in Fig.~\ref{fig:cycle}.
Given any input triplet $(\mathcal{P}_{t'},\mathcal{P}_{t},\mathcal{B}_{t})$, the tracker estimates the target bbox $\mathcal{B}_{t'}$ at $\mathcal{P}_{t'}$.
We then build another triplet $(\mathcal{P}_{t},\mathcal{P}_{t'},\mathcal{B}_{t'})$ to do the tracking in a reverse direction.
The cycle-consistent loss penalizes the errors between $\mathcal{B}_{t}$ and the backward estimated bbox $\mathcal{\hat B}_{t}$:
\begin{equation}\label{eq:loss_cycle}
\begin{aligned}
\mathcal{L}_{cycle} &= \mathcal{L}_{box}(\mathcal{B}_{t},\mathcal{\hat B}_{t}), \\
\mathcal{\hat B}_{t} = \mathcal{F}&(\mathcal{P}_{t},\mathcal{P}_{t'},\mathcal{F}(\mathcal{P}_{t'},\mathcal{P}_{t},\mathcal{B}_{t})),
\end{aligned}
\end{equation}
where $\mathcal{L}_{box}$ is defined as the huber loss~\cite{ren2015faster} between the centers and the heading angles of $\mathcal{B}_{t}$ and $\mathcal{\hat B}_{t}$:
\begin{equation}
  \begin{aligned}
\mathcal{L}_{box} = \text{huber}([x,y,z,sin(\theta)],[\hat x,\hat y,\hat z,sin(\hat \theta)]).
\end{aligned}
\end{equation}
And $\mathcal{F}$ is the tracker function defined in Eqn.~\eqref{eq:tracker}.
We do not penalize the bbox size because it is unchanged for the same target.
For simplicity, we formulate $\mathcal{L}_{cycle}$ on only one input sample.

By imposing an extra regularization on the cycle consistency, $\mathcal{L}_{cycle}$ effectively helps to reduce the noisy influence of the pseudo labels.
But $\mathcal{L}_{cycle}$ also causes instability when applied to the unlabeled data alone.
When the forward tracking predicts off-course target bboxes, using these inaccurate bboxes to construct backward triplets introduces many noisy distractions.
With such an ill-posed penalty, $\mathcal{L}_{cycle}$ may eventually degenerate to the "zero motion" solution.
Fortunately, the "zero motion" solution for $\mathcal{L}_{cycle}$ does not produce a zero $\mathcal{L}_{forward}$.
One can regard the pseudo labels in $\mathcal{L}_{forward}$ as reasonable anchors that pull the motion predictions out of zeros.
The final loss in SEMIM can be written as:
\begin{equation}\label{eq:loss_total}
  \begin{aligned}
\mathcal{L}_{semi} = \mathcal{L}_{forward} + \alpha\mathcal{\widetilde{L}}_{cycle},
\end{aligned}
\end{equation}
where $\mathcal{\widetilde{L}}_{cycle}$ is the average of $\mathcal{L}_{cycle}$ on all the unlabeled training triplets.
Thanks to the motion-centric tracker, $\mathcal{\widetilde{L}}_{cycle}$ is naturally differentiable with respect to the tracker parameters because no cropping happens during the cycle training.

\section{Experiments}
\label{sec:exp}

%
\begin{figure*}[t]
  \centering
   \includegraphics[width=\linewidth]{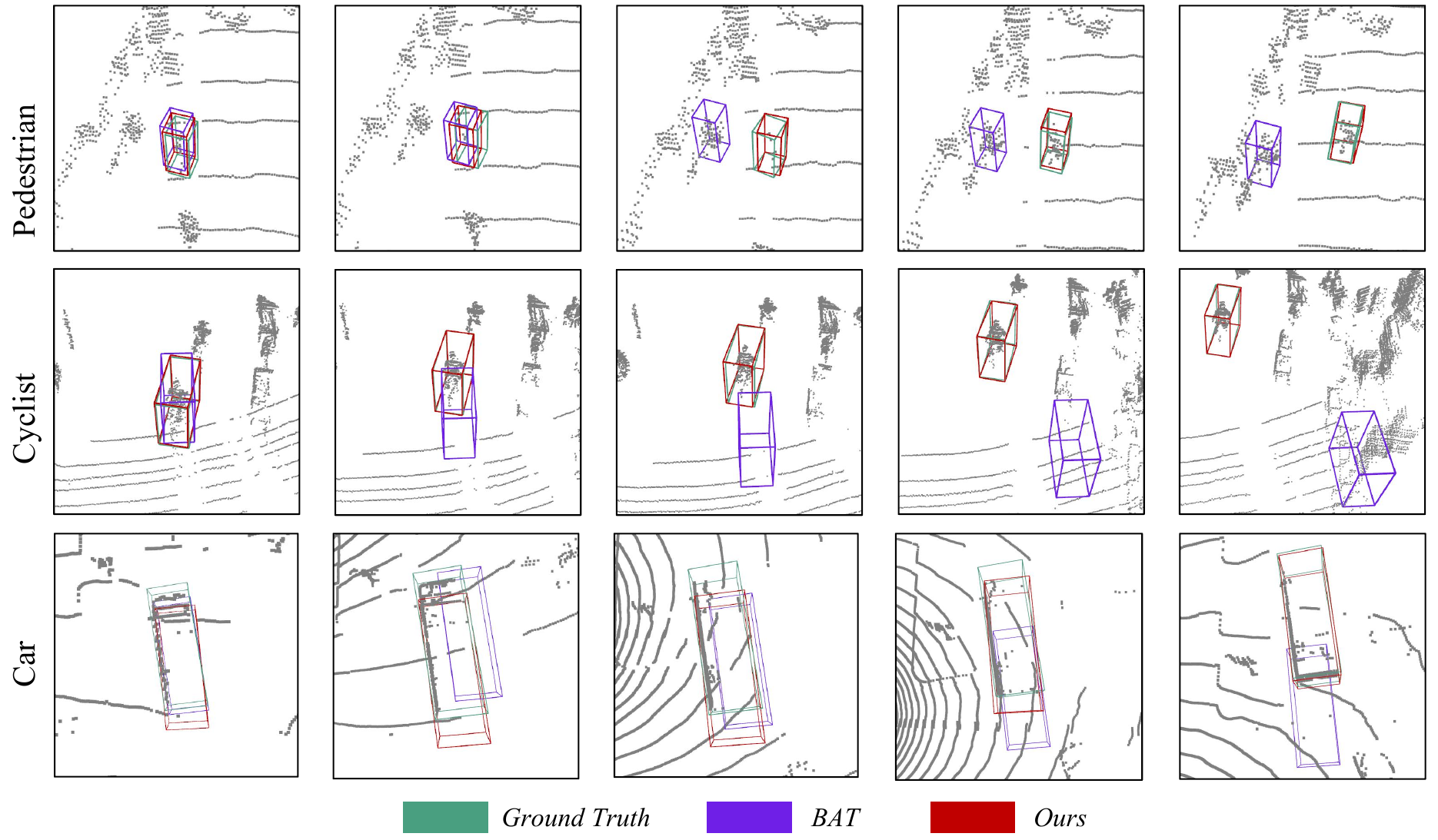}

   \caption{Visualization results. \textbf{Top:} Distractor case in KITTI. \textbf{Middle:} Large motion case in KITTI. \textbf{Bottom:} Case in NuScenes.}
   \label{fig:visualization}
\end{figure*}
\subsection{Experiment Setups}\label{sec:setup}
\noindent\textbf{Datasets.} We extensively evaluate our approach on three large-scale datasets: KITTI~\cite{Geiger2012CVPR}, NuScenes\cite{caesar2020nuscenes} and Waymo Open Dataset (WOD)~\cite{sun2020scalability}. 
We follow~\cite{Giancola_2019_CVPR} to adapt these datasets for 3D SOT by extracting the tracklets of annotated tracked instances from each of the scenes.
\textbf{KITTI} contains 21 training sequences and 29 test sequences. We follow previous works~\cite{Giancola_2019_CVPR,qi2020p2b,zheng2021box} to split the training set into train/val/test splits due to the inaccessibility of the test labels.
\textbf{NuScenes} contains 1000 scenes, which are divided into 700/150/150 scenes for train/val/test. Officially, the train set is further evenly split into ``train\_track" and ``train\_detect" to remedy overfitting. Following~\cite{zheng2021box}, we train our model with ``train\_track" split and test it on the val set. 
\textbf{WOD} includes 1150 scenes with 798 for training, 202 for validation, and 150 for testing. We do training and testing respectively on the training and validation set.
Note that NuScenes and WOD are much more challenging than KITTI due to larger data volumes and complexities.
The LiDAR sequences are sampled at 10Hz for both KITTI and WOD.
Though NuScenes samples at 20Hz, it only provides the annotations at 2Hz.
Since only annotated keyframes are considered, such a lower frequency for keyframes introduces additional difficulties for NuScenes.

\begin{table}
  \renewcommand\tabcolsep{4pt} 
  \footnotesize

  \caption{Comparison among our methods and the state-of-the-art methods on the KITTI datasets. \textit{Mean} shows the average result weighed by frame numbers. $^*$ after our methods denotes that models are trained with improved motion augmentation.}
  
    \begin{center}
    \begin{tabular}{lc|ccccc}
        \toprule[.05cm]
        \multirow{2}*{}
        & Category & Car   & Pedestrian & Van & Cyclist &Mean \\
        & Frame Number &\textit{6424} &\textit{6088} &\textit{1248} &\textit{308} &\textit{14068} \\
        \midrule
        \midrule
        \multirow{15}*{\rotatebox{90}{Success}}
        & SC3D~\cite{Giancola_2019_CVPR} &41.3 & 18.2 &40.4 &41.5 &31.2 \\
        & SC3D-RPN~\cite{zarzar2019efficient} &36.3 & 17.9 & - &{43.2} & -  \\
        & P2B~\cite{qi2020p2b} &56.2 &28.7 &40.8 &32.1 &42.4 \\
        & 3DSiamRPN~\cite{fang20203d} &{58.2} &{35.2} &{45.6} &36.1 &{46.6} \\
        & LTTR~\cite{cui20213d} &{65.0} &{33.2} &{35.8} &{66.2} &{48.7} \\
        & PTT~\cite{shan2021ptt} &{67.8} &{44.9} &{43.6} &37.2 & {55.1} \\
        & V2B~\cite{hui20213d} &{70.5} &{48.3} &{50.1} &40.8 &{58.4}\\
        & BAT~\cite{zheng2021box} &{65.4} &{45.7} &{52.4} &33.7 &{55.0} \\
        & PTTR~\cite{zhou2022pttr} &{65.2} &{50.9} &{52.5} &65.1&{58.4} \\
        & CAT~\cite{gao2023spatio} & 66.6 & {51.6} & {53.1} & {67.0} & {58.9}\\
        & TAT~\cite{lan2022temporal} &72.2 &57.4 &58.9 &74.2 & 64.7 \\

        \cmidrule{2-7}
        & M-Vanilla$^*$ (Ours) &{65.5} &{56.0} &{52.7} &{67.1} &{60.3}\\
        & $M^2$-Track (Ours) &{65.5} &{61.5} &{53.8} &{73.2} &{62.9}\\
        & $M^2$-Track$^*$ (Ours) &{71.1} &\textbf{61.8} &{62.8} &{75.9} &{66.5}\\
        & $M^2$-Track$^*$ $3 \times$ (Ours) &\textbf{74.7} &{59.9} &\textbf{63.0} &\textbf{76.1} &\textbf{67.3}\\
        \midrule
        \midrule
        \multirow{15}*{\rotatebox{90}{Precision}}
        & SC3D~\cite{Giancola_2019_CVPR} &57.9 & 37.8 &47.0 &70.4 &48.5 \\
        & SC3D-RPN~\cite{zarzar2019efficient} &51.0 & 47.8 &- &{81.2} &-  \\
        & P2B~\cite{qi2020p2b} &72.8 &49.6 &48.4 &44.7 &60.0 \\
        & 3DSiamRPN~\cite{fang20203d} &{76.2} &{56.2} &{52.8} &49.0 &{64.9} \\
        & LTTR~\cite{cui20213d} &{77.1} &{56.8} &{45.6} &{89.9} &{65.8} \\
        & PTT~\cite{shan2021ptt} &{81.8} &{72.0} &{52.5} &47.3 &{74.2}\\
        & V2B~\cite{hui20213d} &{81.3} &{73.5} &{58.0} &49.7 &{75.2}\\
        & BAT~\cite{zheng2021box} &{78.9} &{74.5} &{67.0} &45.4 &{75.2} \\
        & PTTR~\cite{zhou2022pttr} &{77.4} &{81.6} &{61.8} &90.5&{77.8} \\
        & CAT~\cite{gao2023spatio} & {81.8} & {77.7} & {69.8} & {90.1} & {79.1}\\
        & TAT~\cite{lan2022temporal} &83.3& 84.4 &69.2 &93.9 &82.8 \\
        \cmidrule{2-7}
        & M-Vanilla$^*$ (Ours) &{79.6} &{84.6} &{69.1} &{92.6} &{81.1}\\
        & $M^2$-Track (Ours) &{80.8} &{88.2} &{70.7} &{93.5} &{83.4}\\
        & $M^2$-Track$^*$ (Ours) &{82.7} &\textbf{88.7} &\textbf{78.5} &{94.0} & 85.2\\
        & $M^2$-Track$^*$ $3 \times$ (Ours) &\textbf{84.9} &\textbf{88.7} &{78.1} &\textbf{94.2} &\textbf{86.2}\\
        \toprule[.05cm]
  
      \end{tabular}
    \end{center}
    
    \label{tab:kitti}
    \end{table}

\begin{table*}
  \renewcommand\tabcolsep{8pt} 
  \footnotesize
  \caption{Comparison of $M^2$-Track against state-of-the-arts on the NuScenes and Waymo Open Dataset.
  \textit{Mean} shows the average result weighed by frame numbers.
  \textbf{Bold} and \underline{underline} denote the best performance and the second-best performance respectively. Improvements over previous state-of-the-art are shown in \textit{Italic}.}
    \begin{center}
    \begin{tabular}{lc|cccccc|ccc}
        \toprule[.05cm]
        \multirow{3}*{}
        &  Dataset  & \multicolumn{6}{c|}{{NuScenes}} & \multicolumn{3}{c}{{Waymo Open Dataset}}\\ 
        & Category  & Car & Pedestrian & Truck & Trailer & Bus  & Mean & Vehicle   & Pedestrian  &Mean\\
        & Frame Number &\textit{64,159} & \textit{33,227} &\textit{13,587} &\textit{3,352} &\textit{2,953} &\textit{117,278} &\textit{1,057,651} &\textit{510,533} &\textit{1,568,184} \\
        \midrule
        \midrule
        \multirow{6}*{\rotatebox{90}{Success}}
        & SC3D~\cite{Giancola_2019_CVPR} & 22.31 & 11.29 & 30.67 & 35.28 & 29.35 & 20.70 &- & - &- \\
        & P2B~\cite{qi2020p2b} &38.81 & 28.39 &42.95 &48.96 &32.95 &36.48 &28.32 & 15.60 & 24.18\\
        & BAT~\cite{zheng2021box} &{40.73} & {28.83} &{45.34} &{52.59} &{35.44} & {38.10} &\underline{35.62} &\underline{22.05} &\underline{31.20} \\
        & CAT~\cite{gao2023spatio} & \underline{43.34} & \underline{30.68} &  \underline{47.64} & \underline{57.90} & \underline{43.30} & \underline{40.67} & - & - & - \\
        \cmidrule{2-11}
        & $M^2$-Track (Ours) &\textbf{55.85} & \textbf{32.10} &\textbf{57.36} &\textbf{57.61} &\textbf{51.39} & \textbf{49.23} &\textbf{43.62} &\textbf{42.10} &\textbf{43.13} \\
        & \textit{Improvement} &\color{RoyalBlue}{\textit{$\uparrow$12.51}} & \color{RoyalBlue}{\textit{$\uparrow$1.42}} &\color{RoyalBlue}{\textit{$\uparrow$9.72}} &\textcolor[rgb]{0.7,0.0,0.0}{\textit{$\downarrow$-0.29}} &\color{RoyalBlue}{\textit{$\uparrow$8.09}} & \color{RoyalBlue}{\textit{$\uparrow$8.56}} &\color{RoyalBlue}{\textit{$\uparrow$8.00}} &\color{RoyalBlue}{\textit{$\uparrow$20.05}} &\color{RoyalBlue}{\textit{$\uparrow$11.92}} \\
        \midrule
        \midrule
        \multirow{6}*{\rotatebox{90}{Precision}}
        & SC3D~\cite{Giancola_2019_CVPR} & 21.93 & 12.65 & 27.73 & 28.12 & 24.08 & 20.20 &- & - &-\\
        & P2B~\cite{qi2020p2b} &43.18 & 52.24 &41.59 &40.05 &27.41 &45.08 &35.41 & 29.56 & 33.51 \\
        & BAT~\cite{zheng2021box} &{43.29} & {53.32} &{42.58} &{44.89} &{28.01} &{45.71} &\underline{44.15} &\underline{36.79} &\underline{41.75}\\
        & CAT~\cite{gao2023spatio} & \underline{49.41} & \underline{56.67} &  \underline{48.10} & \underline{55.31} & \underline{41.42} & \underline{51.28} & - & - & -\\
        \cmidrule{2-11}
        & $M^2$-Track (Ours) &\textbf{65.09} & \textbf{60.92} &\textbf{59.54} &\textbf{58.26} &\textbf{51.44} &\textbf{62.73} &\textbf{61.64} &\textbf{67.31} &\textbf{63.48}\\
        & \textit{Improvement} &\color{RoyalBlue}{\textit{$\uparrow$15.68}} & \color{RoyalBlue}{\textit{$\uparrow$4.25}} &\color{RoyalBlue}{\textit{$\uparrow$11.44}} &\color{RoyalBlue}{\textit{$\uparrow$2.95}} &\color{RoyalBlue}{\textit{$\uparrow$10.02}} & \color{RoyalBlue}{\textit{$\uparrow$11.45}} &\color{RoyalBlue}{\textit{$\uparrow$17.49}} &\color{RoyalBlue}{\textit{$\uparrow$30.52}} &\color{RoyalBlue}{\textit{$\uparrow$21.73}} \\
        \toprule[.05cm]

      \end{tabular}
    \end{center}
    
    \label{tab:nusc_waymo}

\end{table*}

\noindent\textbf{Evaluation Metrics.} We evaluate the models using the One Pass Evaluation (OPE)~\cite{wu2013online}. It defines \textit{overlap} as the Intersection Over Union (IOU) between the predicted and ground-truth BBox, and defines \textit{error} as the distance between two BBox centers. We report the \textit{Success} and \textit{Precision} of each model in the following experiments. \textit{Success} is the Area Under the Curve (AUC) with the \textit{overlap} threshold varying from 0 to 1. \textit{Precision} is the AUC with the \textit{error} threshold from 0 to 2 meters.
For fair comparisons, we adhere to the established practice~\cite{Giancola_2019_CVPR,qi2020p2b,zheng2021box} of training and testing models separately for each category.
To provide an overall metric encompassing all classes, we report the average results, weighted by the number of frames in each category. 
\subsection{Fully-supervised Learning}\label{sec:fully_exp}
\subsubsection{Comparison with State-of-the-arts}\label{sec:main_comparion}
\noindent\textbf{Results on KITTI.}
We compare $M^2$-Track with 11 top-performance approaches~\cite{Giancola_2019_CVPR,zarzar2019efficient,qi2020p2b,fang20203d,cui20213d,shan2021ptt,zheng2021box,hui20213d,zhou2022pttr,gao2023spatio,lan2022temporal}, which have published results on KITTI.
To further demonstrate the potential of the proposed motion-centric paradigm, we additionally include two variants of $M^2$-Track in the comparison:
\begin{itemize}
\setlength{\itemsep}{0pt}
\setlength{\parsep}{0pt}
\setlength{\parskip}{0pt}
    \item M-Vanilla is the proof-of-concept motion-centric tracker as described in Sec.~\ref{sec:vanilla}. It only uses motion modeling to do the tracking without any bells and whistles.
    \item $M^2$-Track $3 \times$ uses \textbf{three} consecutive frames to do the tracking using our multi-frame ensembling strategy in Sec.~\ref{sec:multi}. It selects the final target bbox from two proposals predicted by RTMs in different time intervals.
\end{itemize}
The comparison is given in Tab.~\ref{tab:kitti}. 
For our methods, we use \textbf{$^*$} to indicate that the corresponding model is trained with the improved motion augmentation including the coin-flip test and temporal flipping
(Sec~\ref{sec:aug}). Otherwise, the model is trained using the basic motion augmentation as in our conference paper~\cite{zheng2022beyond}.
We keep this notation in the following experiments.

As shown in Tab.~\ref{tab:kitti}, our methods benefits both rigid and non-rigid object tracking, outperforming current approaches under all categories mostly by large margins.
When comparing $M^2$-Track and $M^2$-Track$^*$, we can see significant improvements for vehicle objects (cars/vans) after using the improved motion augmentation.
We attribute this to coin-flip test, which helps preserve real static motion and thus avoids the model being biased toward moving vehicles.
By contrast, the coin-flip test's improvements on pedestrians and cyclists are minor since these objects are mostly moving.
More analysis about motion augmentation is given in Sec.~\ref{sec:analysis}.
Since distractors are usually more pervasive for smaller objects, our improvements for pedestrians and cyclists are more significant than those for vehicle objects when compared to previous matching-based approaches.
Please refer to the appendix for more details about distractors.
It is indeed exciting to observe that M-Vanilla$^*$ surpasses all the matching-based methods on average, even with its simple architecture. 
The only exception is TAT~\cite{lan2022temporal}, which employs 8 template frames in its approach.
This outcome serves as strong evidence that our advancements primarily stem from the effectiveness of the motion-centric paradigm.
Thanks to the multi-frame ensembling, $M^2$-Track$^*$ $3\times$ further boosts the performance especially for cars, achieving the best overall results.

\begin{table}[t]

  \footnotesize
  \centering
  \caption{Influence of Motion Augmentation. ``Aug" stands for the basic motion augmentation. Coin and Temp. stand for the coin-flip test and the temporal-flipping augmentation, respectively.}
  \resizebox{\columnwidth}{!}{
    \begin{tabular}{c|ccc|cc}
        \toprule
     Method & Aug.& Coin & Temp. & Success & Precision \\

    \midrule\midrule
    \multirow{2}*{BAT~\cite{zheng2021box}}
    & \xmark & \xmark & \xmark & 65.37  & 78.88  \\
    & \cmark & \xmark & \xmark & 63.59 \textcolor[rgb]{0.7,0.0,0.0}{\small $\downarrow$ 1.78} &  76.99 \textcolor[rgb]{0.7,0.0,0.0}{\small $\downarrow$ 1.89} \\
    \midrule
    \multirow{2}*{P2B~\cite{qi2020p2b}}
    & \xmark & \xmark & \xmark & 56.20 & 72.80 \\
    & \cmark & \xmark & \xmark &  55.21 \textcolor[rgb]{0.7,0.0,0.0}{\small $\downarrow$ 0.99} &  71.51 \textcolor[rgb]{0.7,0.0,0.0}{\small $\downarrow$ 1.29} \\
    \midrule
    \multirow{5}*{$M^2$-Track}
    & \xmark & \xmark & \xmark & 65.29    & 77.12  \\
    & \cmark & \xmark & \xmark & 65.49 \color{RoyalBlue}{\small $\uparrow$ 0.20} & 80.81 \color{RoyalBlue}{\small $\uparrow$ 3.69}\\
    & \cmark & \cmark & \xmark & 69.93 \color{RoyalBlue}{\small $\uparrow$ 4.64} & 82.07 \color{RoyalBlue}{\small $\uparrow$ 4.95}\\
    & \cmark & \xmark & \cmark & 65.11 \textcolor[rgb]{0.7,0.0,0.0}{\small $\downarrow$ 0.18} & 78.62 \color{RoyalBlue}{\small $\uparrow$ 1.50}\\

    & \cmark & \cmark & \cmark & 71.14 \color{RoyalBlue}{\small $\uparrow$ 5.85} & 82.67 \color{RoyalBlue}{\small $\uparrow$ 5.55}\\
    \bottomrule
  \end{tabular}%
  }
  \label{tab:augmentation}%

\end{table}%

\noindent\textbf{Results on NuScenes \& WOD.}
We select three representative open-source works: SC3D~\cite{Giancola_2019_CVPR}, P2B~\cite{qi2020p2b} and BAT~\cite{zheng2021box} as our competitors on NuScenes and WOD. The results on NuScenes except for the Pedestrian class are provided by~\cite{zheng2021box}. We use the published codes of these competitors to obtain other results absent in~\cite{zheng2021box}.
SC3D~\cite{Giancola_2019_CVPR} is omitted for WOD comparison due to its costly training time.
Besides, we also compare our method with the spatial-temporal tracker CAT~\cite{gao2023spatio} using their published results on NuScenes.
As shown in Tab.~\ref{tab:nusc_waymo}, $M^2$-Track exceeds all the competitors on average, mostly by a large margin for each category.
On such two challenging datasets with pervasive distractors and drastic appearance changes, the performance gap between previous approaches and $M^2$-Track becomes even 
larger (\eg, more than \textbf{30\% }precision gain on Waymo Pedestrian).
Note that for large objects (\ie, Truck, Trailer, and Bus), even if the predicted centers are far from the target (reflected from lower precision), the output BBoxes of the previous model may still overlap with the ground truth (results in higher success).
In contrast, motion modeling helps to improve not only the success but also the precision by a large margin (\eg, {\textbf{+10.02\%}} gain on Bus) for large objects.
Visualization results are provided in Fig.~\ref{fig:visualization} and the supplementary.

\begin{table}
  \renewcommand\tabcolsep{4pt} 
  \footnotesize

  \caption{The performance of $M^2$-Track* (with improved motion augmentation) for novel object tracking. We compare its performance with the models separately trained (Sep.) for each class. \textcolor{RoyalBlue}{Blue} denotes seen objects while \textcolor[rgb]{0.7,0.0,0.0}{red} denotes unseen objects during the training.}
  
    \begin{center}
    \begin{tabular}{cc|cc|cc|c}
        \toprule[.05cm]
        \multirow{2}*{$M^2$-Track$^*$}
        & Cat. & Car   & Pedestrian & Van & Cyclist &Mean \\
        & Frame &\textit{6424} &\textit{6088} &\textit{1248} &\textit{308} &\textit{14068} \\
        \midrule
        \midrule
        \multirow{2}*{{Success}}
        & Sep. &\textbf{71.14} &{61.80} &\textbf{62.84} &\textbf{75.89} &\textbf{66.47}\\
        \cmidrule{2-7}
        & Joint &\color{RoyalBlue}{69.80} &\color{RoyalBlue}\textbf{64.47} &\textcolor[rgb]{0.7,0.0,0.0}{52.16} &\textcolor[rgb]{0.7,0.0,0.0}{69.02} &{65.91}\\
        \midrule
        \midrule
        \multirow{2}*{{Precision}}
        
        & Sep.  &\textbf{82.67} &{88.70} &\textbf{78.45} &\textbf{94.04} & \textbf{85.15}\\
        \cmidrule{2-7}
        & Joint &\color{RoyalBlue}{80.86} &\color{RoyalBlue}\textbf{90.12} &\textcolor[rgb]{0.7,0.0,0.0}{64.83} &\textcolor[rgb]{0.7,0.0,0.0}{92.61} & 83.70\\
       
        \toprule[.05cm]
  
      \end{tabular}
    \end{center}
    
    \label{tab:general_tracking}
    \end{table}
    
\subsubsection{Tracking Novel Objects}\label{sec:general_tracking}
To ensure a fair comparison, both Tab.~\ref{tab:kitti} and Tab.~\ref{tab:nusc_waymo} adhere to the practice of training a separate model for each category.
Nevertheless, one of the significant advantages of SOT over MOT lies in its ability to track arbitrary novel objects.
Separately training a model for each category is not applicable to novel object tracking.
In this section, we evaluate the performance of our model when it is jointly trained on multiple classes. Additionally, we assess its generalization ability by applying it to novel objects without retraining.
Specifically, we train a $M^2$-Track$^*$ model using car and pedestrian sequences from the KITTI train split. Subsequently, we evaluate its performance on the KITTI test split, considering both the seen classes (cars and pedestrians) and the unseen ones (cyclists and vans).
The results are presented in Tab.~\ref{tab:general_tracking}, which also includes the per-class performance of the model separately trained on the corresponding category.
As demonstrated in Tab.~\ref{tab:general_tracking}, the jointly trained model achieves comparable performance on seen classes, and notably, its performance on pedestrians even exceeds that of the separately tuned model by noticeable margins (2.67\% higher in success and 1.42\% higher in precision).
The performance of the jointly trained model on vans and cyclists is also satisfying, given that the model has never seen these classes during the training. For unseen vans and cyclists, while it performs worse than our separately trained model, the jointly trained model is still comparable to some top-ranking matching-based methods (\eg, PTTR~\cite{zhou2022pttr}, CAT~\cite{gao2023spatio}), which are also separately tuned.
In summary, our method demonstrates great generalization ability and can effectively handle novel objects without the need for retraining.

    \begin{table}
      \renewcommand\tabcolsep{4pt} 
      \footnotesize
    
      \caption{Comparison with MOT methods.}
      
        \begin{center}
        \begin{tabular}{cc|cc}
            \toprule[.05cm]
            
            & Category & Car   & Pedestrian \\
            \midrule
            \midrule
            \multirow{3}*{{Success}}
            & AB3D~\cite{Weng2020_AB3DMOT} &{37.5 } &{17.6 } \\
            & PC3T~\cite{wu20213d}  &{51.9 } &{23.6} \\
            \cmidrule{2-4}
            & $M^2$-Track$^*$ (Ours) &\textbf{71.1} &\textbf{61.8}\\
            \midrule
            \midrule
            \multirow{3}*{{Precision}}
            
            & AB3D~\cite{Weng2020_AB3DMOT} &{42.3 } &{27.3} \\
            & PC3T~\cite{wu20213d}  &{59.2 } &{34.1} \\
            \cmidrule{2-4}
            & $M^2$-Track$^*$ (Ours) &\textbf{82.7} &\textbf{88.7}\\
           
            \toprule[.05cm]
      
          \end{tabular}
        \end{center}
        \label{tab:mot}
     \end{table}

\subsubsection{Comparison with MOT Approaches}\label{sec:mot}
To further demonstrate the essence of 3D SOT, we compare our method with 3D MOT approaches under the metrics of success and precision.
We select AB3D~\cite{Weng2020_AB3DMOT} and PC3T~\cite{wu20213d} as the competitors, which are two representative 3D MOT methods KITTI MOT leaderboard.
For a fair comparison, we use the ground truth BBoxes to initialize the detection results for AB3D~\cite{Weng2020_AB3DMOT} and PC3T~\cite{wu20213d} in the first frame instead of using the outputs from their detectors.
After that, we run AB3D~\cite{Weng2020_AB3DMOT}/PC3T~\cite{wu20213d} on all the test sequences and then collect the output tracklets for each instance of interest.
Finally, we compute the success and precision by comparing the output tracklets with the ground truths.
%
As evidenced in Tab.~\ref{tab:mot}, a substantial performance gap exists between 3D SOT and 3D MOT, with SOT significantly outperforming MOT.
Since MOT cares about all the objects in the scenes and relies on a detector to localize the objects, it cannot provide precise localization for each object as SOT.
Moreover, MOT cannot handle novel objects, which is a significant advantage of SOT as demonstrated in Sec.~\ref{sec:general_tracking}.

\begin{table}[t]
  \centering
  \footnotesize
  \caption{Apply $M^2$-Track on Appearance Matching-Based Methods. $d$ denotes the margin used for search area generation. $^*$ after our methods denotes that models are trained with improved motion augmentation.}
    \begin{tabular}{c|cc}
    \toprule
    Method & Success & Precision \\
    \midrule\midrule
    PTT~\cite{shan2021ptt} & 67.80 & 81.80 \\
    V2B~\cite{hui20213d} & {70.50} & 81.30 \\
    TAT~\cite{lan2022temporal} &72.20 & 83.30 \\
    \midrule
    $M^2$-Track & 65.49 & 80.81 \\
    $M^2$-Track + BAT~\cite{zheng2021box}, $d=2$& 69.22 \color{RoyalBlue}{\small $\uparrow$ 3.73} & 81.09 \color{RoyalBlue}{\small $\uparrow$ 0.28} \\
    $M^2$-Track + P2B~\cite{qi2020p2b}, $d=2$& 70.21 \color{RoyalBlue}{\small {$\uparrow$ 4.72}} & {81.80 }\color{RoyalBlue}{\small {$\uparrow$ 0.99}} \\
    \midrule
    $M^2$-Track$^*$  & 71.14 & 82.67 \\
    $M^2$-Track$^*$  + BAT~\cite{zheng2021box}, $d=2$& 71.80 \color{RoyalBlue}{\small $\uparrow$ 0.66} & 83.52 \color{RoyalBlue}{\small $\uparrow$ 0.85} \\
    $M^2$-Track$^*$  + BAT~\cite{zheng2021box}, $d=1$& \textbf{73.04} \color{RoyalBlue}{\small $\uparrow$ 1.90} & \textbf{85.44} \color{RoyalBlue}{\small $\uparrow$ 2.77} \\
    \bottomrule
    \end{tabular}%
  \label{tab:appearanc_matching}%
\end{table}%
\subsubsection{Combine with Appearance Matching}\label{sec:motion_matching_exp}
In this section, we investigate the tracking performance achieved by combining $M^2$-Track with existing matching-based approaches using the way described in Sec.~\ref{sec:motion_matching}.
Following Sec.~\ref{sec:motion_matching}, we first use a well-trained $M^2$-Track to output an initial target BBox $\mathcal{B}_{motion}$ and then generate the search area $\mathcal{P}_{search}$ and target template $\mathcal{P}_{temp}$.
Specifically, we expand the initial BBox $\mathcal{B}_m$ by a margin of $d$ meter(s) and sample 1024 points inside to generate a search area ${P}_{search}$.
And the target template is merged with the target point cloud in the first frame and then downsampled to 512 points.
Next, we apply a fully-trained matching-based tracker to the search area and target template, resulting in a refined target BBox.
To evaluate the tracking performance of this combination, we conducted experiments on KITTI Cars, as presented in Tab.\ref{tab:appearanc_matching}.
We selected P2B\cite{qi2020p2b} and BAT~\cite{zheng2021box} as the representative matching-based methods. 
Both P2B and BAT are configured with their official settings, except for the margins used in the search area generation.
Specifically, we considered two different margins: $d=2$ (officially adopted in P2B and BAT) and $d=1$.

Tab.~\ref{tab:appearanc_matching} confirms that such a simple combination effectively boosts the performance of $M^2$-Track.
We also notice that the improvement becomes less significant as $M^2$-Track becomes stronger with the improved motion augmentation.
This aligns with our intuition that the matching-based refinement becomes less impactful when the initial bounding box is already accurate.
Furthermore, narrowing down the search area for the matching-based method leads to further performance improvement (as seen in the last two rows of Tab.~\ref{tab:appearanc_matching}), as it reduces the chances of being distracted by negative matches.
Last but not least, our best combination model ($M^2$-Track$^*$ + BAT, d=1) outperforms the top-ranking TAT~\cite{lan2022temporal} by considerable margins.
We believe that one can further boost 3D SOT by combining motion-based and matching-based paradigms with a more delicate design.

\subsubsection{Analysis Experiments}\label{sec:analysis}

\begin{figure}[t]
  \centering
  \includegraphics[width=0.8\linewidth]{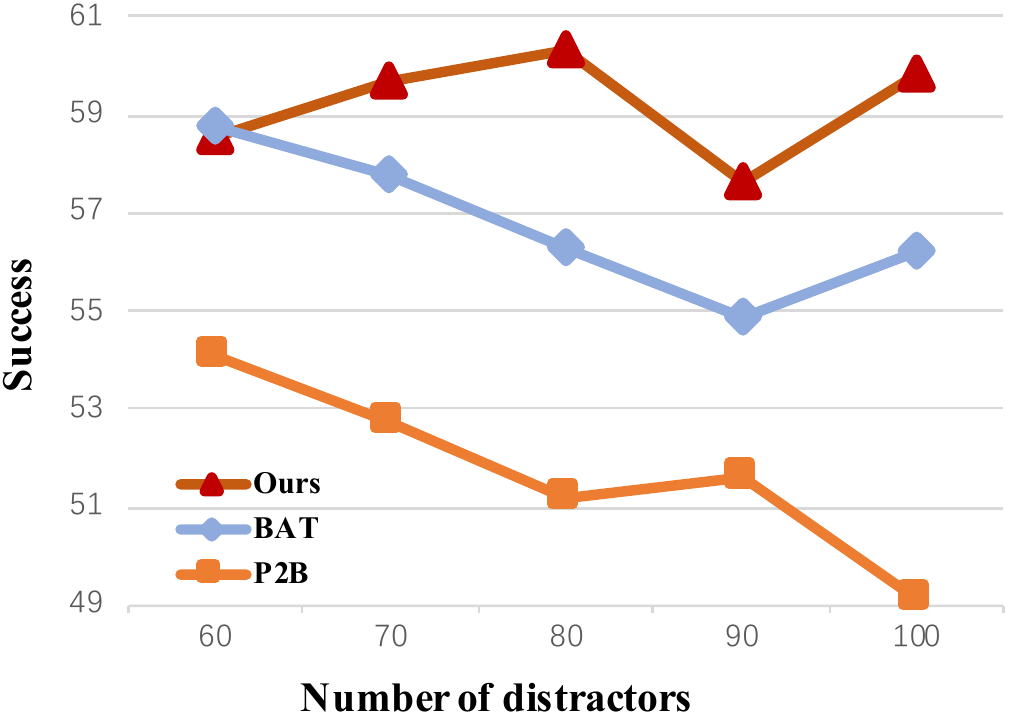}
  \caption{Robustness analysis with variant numbers of distractors.}
   \label{fig:distractors}
\end{figure}

In this section, we extensively analyze $M^2$-Track's performance under various settings regarding the distractor distribution, data augmentation and model configuration.
%
%
%
All the experiments are conducted on the \textit{Car category of KITTI} unless otherwise stated.

\begin{table}[t]
  \centering
  \footnotesize
  \caption{Analysis of the Segmentation Network in $M^2$-Track. $^*$ denotes the model trained with the improved motion augmentation. Seg. Net. stands for the segmentation network.}
  \begin{tabular}{c|ccc}
  
  \toprule
  Model & Seg. Net. & Success & Precision  \\
  \midrule\midrule
  M-Vanilla$^*$    &   -  & 65.45    & 79.63        \\
  \midrule
  $M^2$-Track$^*$ & PointNet~\cite{qi2017pointnet}           &\textbf{71.14}   & {82.67}      \\
  $M^2$-Track$^*$ &  PointNet++~\cite{qi2017pointnet++}         & 68.33   & 82.72      \\
  $M^2$-Track$^*$ & PointTransformer~\cite{zhao2021point}            & 70.97   & \textbf{82.75}      \\
  \bottomrule
  \end{tabular}
  \label{tab:seg_net}%
\end{table}

\begin{table}[t]
  \centering
  \footnotesize
  \caption{Number of Frames in Multi-frame Ensembling for $M^2$-Track. All the models are trained with improved motion augmentation.}
  \begin{tabular}{c|ccc}
  
  \toprule
  Num Frames & Success & Precision &  \\
  \midrule\midrule
  2         & 71.14    & 82.67      &  \\
  \midrule
  3          & \textbf{74.65}   & \textbf{84.93}     &  \\
  4          & 73.20   & 83.58     &  \\
  5          & 70.72   & 81.23     & \\
  \bottomrule
  \end{tabular}
  \label{tab:num_frames}%
\end{table}

\begin{table*}[htbp]
  \centering
  \caption{Results of $M^2$-Track when different modules are ablated. The last row denotes the full model. \textbf{Bold} denotes the largest change.}
    \begin{tabular}{cccc|cc|cc}
    \toprule
    {\multirow{2}{*}{\shortstack[c]{\\Box Aware\\ Enhancement}}} & 
    {\multirow{2}{*}{\shortstack[c]{\\Prev Box \\ Refinement}}} & 
    {\multirow{2}{*}{\shortstack[c]{\\Motion\\Classification}}} & 
    {\multirow{2}{*}{\shortstack[c]{\\Stage-II}}} & \multicolumn{2}{c|}{Kitti} & \multicolumn{2}{c}{NuScenes} \\
\cmidrule{5-8}          &       &       &       &  
{Success} & {Precision} & {Success} & {Precision} \\
    \midrule\midrule
               & \cmark     & \cmark     & \cmark     & 62.00 \textcolor[rgb]{0.7,0.0,0.0}{\small $\downarrow$ 3.49}   & 76.15 \textcolor[rgb]{0.7,0.0,0.0}{\small \textbf{$\downarrow$ 4.66}}&  53.68 \textcolor[rgb]{0.7,0.0,0.0}{\small \textbf{$\downarrow$ 2.17}}    &62.47 \textcolor[rgb]{0.7,0.0,0.0}{\small $\downarrow$ 2.62}  \\
    \cmark     &            & \cmark     & \cmark     & 64.23 \textcolor[rgb]{0.7,0.0,0.0}{\small $\downarrow$ 1.26}   & 78.12 \textcolor[rgb]{0.7,0.0,0.0}{\small $\downarrow$ 2.69}&    54.70 \textcolor[rgb]{0.7,0.0,0.0}{\small $\downarrow$ 1.15}   &  61.94 \textcolor[rgb]{0.7,0.0,0.0}{\small \textbf{$\downarrow$ 3.15}} \\
    \cmark     & \cmark     &            & \cmark     & 65.74 \color{RoyalBlue}{\small $\uparrow$ 0.25}     & 80.29 \textcolor[rgb]{0.7,0.0,0.0}{\small $\downarrow$ 0.52}&  54.88 \textcolor[rgb]{0.7,0.0,0.0}{\small $\downarrow$ 0.97}     & 64.40 \textcolor[rgb]{0.7,0.0,0.0}{\small $\downarrow$ 0.69} \\
    \cmark     & \cmark     & \cmark     &            & 61.29 \textcolor[rgb]{0.7,0.0,0.0}{\small \textbf{$\downarrow$ 4.20}}   & 77.31 \textcolor[rgb]{0.7,0.0,0.0}{\small $\downarrow$ 3.50}&    54.66 \textcolor[rgb]{0.7,0.0,0.0}{\small $\downarrow$ 1.99}   & 64.15 \textcolor[rgb]{0.7,0.0,0.0}{\small $\downarrow$ 0.94} \\
    \cmark     & \cmark     & \cmark     & \cmark     & 65.49                & 80.81 & 55.85 & 65.09 \\
    \bottomrule
    \end{tabular}%
  \label{tab:ablation}%
\end{table*}%

\noindent\textbf{Robustness to distractors.}
Though achieving promising improvement on NuScenes and WOD, $M^2$-Track brings little improvement on the Car of KITTI.
To explain this, we look at the scenes of three datasets and find that the surroundings of most cars in KITTI are free of distractors, which are pervasive in NuScenes and WOD (see the supplementary).
Although appearance-matching-based methods are sensitive to distractors, they provide more precise results than our motion-based approach in distractor-free scenarios. But as the number of distractors increases, these methods suffer from noticeable performance degradation due to ambiguities from the distractors.
To verify this hypothesis, we randomly add $K$ car instances to each scene of KITTI, and then re-train and evaluate different models using this synthesis dataset. 
As shown in Fig.~\ref{fig:distractors}, $M^2$-Track consistently outperforms the other two matching-based methods in scenes with more distractors, and the performance gap grows as $K$ increases.
Thanks to the box-awareness, BAT~\cite{zheng2021box} can aid such ambiguities to some extent.
But our performance is more stable than BAT's when more distractors are added.
Besides, the first row in Fig.~\ref{fig:visualization} shows that, when the number of points decreases due to occlusion, BAT is misled by a distractor and then tracks off course, while $M^2$-Track keeps holding tight to the ground truth.
All these observations demonstrate the robustness of our approach. 

\noindent\textbf{Influence of motion augmentation.}
We improve the performance of $M^2$-Track using the motion augmentation in training, which is not adopted in previous approaches.
For a fair comparison, we re-train BAT~\cite{zheng2021box} and P2B~\cite{qi2020p2b} using the same configurations in their open-source projects except additionally adding motion augmentation. 
Tab.~\ref{tab:augmentation} shows that motion augmentation instead has an adverse effect on both BAT and P2B.
Our model benefits from motion augmentation since it explicitly models target motion and is robust to distractors.
In contrast, motion augmentation may move a target closer to its potential distractors and thus harm those appearance-matching-based approaches.
We also analyze the effectiveness of the coin-flip test and the temporal-flipping, which are two enhancements over the basic motion augmentation.
Compared to the basic motion augmentation, adding the coin-flip test brings noticeable improvements, especially in terms of success.
The temporal-flipping generates flipped motions, which are usually unrealistic in natural scenes.
Besides, some of the target motions synthesized by the basic motion augmentation are also unnatural.
Therefore, adding the temporal-flipping alone to the basic motion augmentation instead harms its effectiveness.
But the temporal-flipping can also benefit the training when applied together with the coin-flip test.
This reflects that preserving original target motions in the data is critical for motion augmentation. 

\noindent\textbf{Role of the segmentation network.}
One major component in $M^2$-Track is the segmentation network, which plays a pivotal role in both Stage I and Stage II.
For simplicity, we only choose a PointNet-based segmentation network~\cite{qi2017pointnet} for $M^2$-Track.
In Tab.~\ref{tab:seg_net}, we conducted experiments with more powerful segmentation networks, such as PointNet++\cite{qi2017pointnet++} and PointTransformer\cite{zhao2021point}.
Surprisingly, the results demonstrate that the tracking performance remains relatively insensitive to the choice of segmentation network, and employing a stronger network does not necessarily lead to improved performance.
Furthermore, it is noteworthy that even in the absence of the segmentation network, M-Vanilla achieves comparable performance with $M^2$-Track while surpassing the majority of previous methods in Table~\ref{tab:kitti}.
These findings further emphasize that the success of $M^2$-Track primarily hinges on effective motion modeling,  validating the significance of our motion-centric paradigm.

\noindent\textbf{Ablations on the frame sampling.}
Tab.~\ref{tab:num_frames} presents the performance of $M^2$-Track when using different numbers of frames for the multi-frame ensembling.
The first entry in Tab.~\ref{tab:num_frames} (num frames = 2) means that a standard $M^2$-Track is used without the multi-frame ensembling.
The results indicate that increasing the number of frames does not necessarily lead to improved performance.
With a larger frame number, the multi-frame ensembling has to predict relative motion for a longer time interval, making it more challenging to achieve accurate results.
Overall, the optimal choice appears to be using three frames, which allows us to achieve favorable performance while maintaining computational efficiency.

\noindent\textbf{Ablations on model components.}
In Tab.~\ref{tab:ablation}, we conduct an exhaustive ablation study on both KITTI and NuScenes to understand the components of $M^2$-Track.
Specifically, we respectively ablate the \textit{box-aware feature enhancement}, \textit{previous BBox refinement}, \textit{binary motion classification} and $2^{nd}$ stage from $M^2$-Track.
In general, the effectiveness of the components varies across the datasets, but removing any one of them causes performance degradation. 
The only exception is the \textit{binary motion classification} used in the $1^{st}$ stage, which causes a slight drop on KITTI in terms of success.
We suppose this is due to the lack of static objects for KITTI's cars, which results in a biased classifier.
Besides, Tab.~\ref{tab:ablation} shows that $M^2$-Track keeps performing competitively even with module ablated, especially on NuScenes.
This reflects that the main improvement of $M^2$-Track is from the motion-centric paradigm instead of the specific pipeline design.


\begin{table*}[t]
	\centering
	\caption{Tracking Results on KITTI Cars with different label ratios. Frame number shows the number of labeled frames used for training. \textit{Improvement} denotes SEMIM's improvement over the baseline $M^2$-Track$^*$.}
	\begin{tabular}{cc|cccccc}
		\toprule
		& Breakpoint (k) &  0&2&4&8&16\\
		 Metric &   Method (Frame)    & {243 (1\%)}  & {3956 (20\%)} & {5137 (26\%)} & {10266 (53\%)}  & {19522 (100\%)} \\ \midrule\midrule
		\multirow{6}[0]{*}{\rotatebox{90}{Success}}& P2B~\cite{qi2020p2b}   & -      & 42.3     & -&-& 56.2  \\
		& BAT~\cite{zheng2021box}    & -    &   43.8       & -&-& 65.4  \\
    \cmidrule{2-7}
		& $M^2$-Track      & 7.6                    & 51.2   & 56.8   & 63.9 &        {65.5}    \\
		& $M^2$-Track$^*$ & 12.7               & 60.6   & 63.5  & 67.3 &        \textbf{71.1}   \\
		& $M^2$-Track$^*$ w/ SEMIM & \textbf{29.4}               & \textbf{65.2}   & \textbf{67.4 }  & \textbf{70.3 }&     -   \\
		& \textit{Improvement} &  \color{RoyalBlue}{\textit{$\uparrow$16.7}}                         &\color{RoyalBlue}{\textit{$\uparrow$4.6}}   & \color{RoyalBlue}{\textit{$\uparrow$3.9}}   & \color{RoyalBlue}{\textit{$\uparrow$3.4}}  & -\\
		\midrule
    \midrule
		\multirow{6}[0]{*}{\rotatebox{90}{Precision}}& P2B~\cite{qi2020p2b}      & -      & 57.2     & - &  - &   72.8 \\
		& BAT~\cite{zheng2021box}      & -    & 56.2     & - & - &     78.9  \\
    \cmidrule{2-7}
		& $M^2$-Track &     5.8                     & 65.2   & 71.1  & 77.0        &       {80.8} \\
		& $M^2$-Track$^*$ &    13.8                         & 76.2   & 78.9 &   80.0 &       \textbf{82.7} \\
		& $M^2$-Track$^*$ w/ SEMIM &   \textbf{34.7}              & \textbf{79.9}   & \textbf{81.4 }  & \textbf{83.0} &       -  \\
		& \textit{Improvement} &        \color{RoyalBlue}{\textit{$\uparrow$20.9}}                  & \color{RoyalBlue}{\textit{$\uparrow$3.7}}  & \color{RoyalBlue}{\textit{$\uparrow$2.5}}   & \color{RoyalBlue}{\textit{$\uparrow$3.0}}  &         -  \\
		\bottomrule
	\end{tabular}%
	\label{tab:kitti_car}%
\end{table*}%
\begin{table*}[t]
	\centering
	\caption{Tracking Results on KITTI Pedestrians with different label ratios.}
	\begin{tabular}{cc|ccccc}
		\toprule
		& Breakpoint (k) &2&8&12&14&16\\
		 Metric &   Method (Frame)     & {314 (7\%)}  & {446 (10\%)} & {770 (17\%)}& {1821 (40\%)}  & {4600 (100\%)} \\\midrule\midrule
		\multirow{6}[0]{*}{\rotatebox{90}{Success}}& P2B~\cite{qi2020p2b}  &- & -&- & - &28.7\\
		& BAT~\cite{zheng2021box}      & -&- &- &  - &45.7\\
    \cmidrule{2-7}
		& $M^2$-Track  &4.8& 6.2& 11.6&  51.1 &{61.5}\\
		& $M^2$-Track$^*$ &18.8& 25.6& 35.9&  55.0 &\textbf{61.8}\\
		& $M^2$-Track$^*$ w/ SEMIM &\textbf{50.5}  & \textbf{49.9}& \textbf{56.1}&  \textbf{63.7 }& -\\
		& \textit{Improvement} &\color{RoyalBlue}{\textit{$\uparrow$31.7}} & \color{RoyalBlue}{\textit{$\uparrow$24.3}}& \color{RoyalBlue}{\textit{$\uparrow$20.2}}&  \color{RoyalBlue}{\textit{$\uparrow$8.7}} & -\\
		\midrule\midrule
		\multirow{6}[0]{*}{\rotatebox{90}{Precision}}& P2B~\cite{qi2020p2b} & -&- &- &-&  49.6     \\
		& BAT~\cite{zheng2021box}   & -&- &- &- &74.5    \\
    \cmidrule{2-7}
		& $M^2$-Track &  9.7 & 13.8& 30.3& 75.8 &{88.2} \\
		& $M^2$-Track$^*$ &  40.3 & 48.5& 59.0& 82.8 & \textbf{88.7}\\
		& $M^2$-Track$^*$ w/ SEMIM &  \textbf{79.8} & \textbf{79.0}& \textbf{84.1}& \textbf{90.8 }& -  \\
		& \textit{Improvement} & \color{RoyalBlue}{\textit{$\uparrow$39.5}} & \color{RoyalBlue}{\textit{$\uparrow$30.5}} & \color{RoyalBlue}{\textit{$\uparrow$25.1}}& \color{RoyalBlue}{\textit{$\uparrow$8.0}} & -  \\
		\bottomrule
	\end{tabular}%
	\label{tab:kitti_pedestrian}%
\end{table*}%

\begin{table}[t]

  \centering
  \caption{Results on WOD validation set. \cmark/\xmark ~indicates the datasets used in the training.  \textit{Improvement} denotes improvement over the KITTI-pretrained $M^2$-Track$^*$.}
    \resizebox{\linewidth}{!}{
    \begin{tabular}{c|cccc|cc}
    	\toprule
     Method & $\mathcal{D}_{kitti}^l$ & $\mathcal{D}_{val}^u$ & $\mathcal{D}_{train}^l$ & $\mathcal{D}_{train}^u$ & Success & Precision \\

    \midrule
    {P2B~\cite{qi2020p2b}} &\xmark&\xmark&\cmark &    \xmark& 28.32 & 35.41 \\
    {BAT~\cite{zheng2021box}} &\xmark&\xmark&\cmark&    \xmark& 35.62 & 44.15 \\
    {$M^2$-Track}&\xmark&\xmark&\cmark&    \xmark& 43.62 & \textbf{61.64} \\
    \midrule
    $M^2$-Track$^*$ &\cmark     &    \xmark &    \xmark   &    \xmark& 35.75 & 44.54 \\
    \midrule
    $M^2$-Track$^*$ UDA &\cmark     & \xmark  &    \xmark&    \cmark   & {45.88} & 56.50 \\
    \textit{Improvement} &-&-&- & - & \color{RoyalBlue}{\textit{$\uparrow$10.13}} & \color{RoyalBlue}{\textit{$\uparrow$11.96}} \\
    \midrule
    $M^2$-Track$^*$ Offline &\cmark     & \cmark  &    \xmark   &    \xmark& \textbf{46.68} & 58.46 \\
    \textit{Improvement} &-&-&- & - & \color{RoyalBlue}{\textit{$\uparrow$10.93}} & \color{RoyalBlue}{\textit{$\uparrow$13.92}} \\
    \bottomrule
    \end{tabular}%
    }
  \label{tab:waymo}%
\end{table}%

\begin{table}[]
  \centering
  \caption{Ablation Study with 20\% labels (3956 labeled frames) on KITTI Cars. PL stands for pseudo labels. CC./Pseudo Aug. stands for the cycle-consistent loss and the pseudo-label-based motion augmentation, respectively.}
  \begin{tabular}{c|cc|cc}
    \toprule
    Model &  {CC.} & {Pseudo Aug.} & Success & Precision \\\midrule
    $M^2$-Track w/o PL & - & - & 51.21 & 65.21\\ 
    \midrule
    \multirow{4}*{$M^2$-Track w/ PL}
    & \xmark & \xmark & 49.52 & 65.56\\
    & \cmark & \xmark & 56.02 & 71.81\\
    & \xmark & \cmark & 58.85 & 75.63\\
    & \cmark & \cmark & 59.09 & 75.79\\

        \bottomrule
  \end{tabular}%
  \label{tab:semi_ablation_cars}%
\end{table}%

\begin{table}[t]
  \centering
  \caption{Applying SEMIM's components to fully supervised training. 20\% and 100\% training data on KITTI Cars are used. CC./GT Aug. stands for the cycle-consistent loss and the GT-based motion augmentation, respectively. All the models are trained with improved motion augmentation.}
  \begin{tabular}{c|cc|cc}
    \toprule
    Training Data &  {CC.} & {GT Aug.} & Success & Precision \\\midrule
    \multirow{4}*{20\%}& \xmark & \xmark & 60.60 & 76.20\\ 
    \cmidrule{2-5}
    & \cmark & \xmark & 60.18 & 76.08\\ 
    & \xmark & \cmark & \textbf{65.68} & \textbf{79.87}\\ 
    & \cmark & \cmark & 64.13 & 78.90\\ 
    \midrule
    \multirow{4}*{100\%}
    & \xmark & \xmark & 71.14 & 82.67\\
    \cmidrule{2-5}
    & \cmark & \xmark & \textbf{72.39} & \textbf{85.32}\\
    & \xmark & \cmark & 71.46 & 83.53\\
    & \cmark & \cmark & 71.14 & 83.43\\
        \bottomrule
  \end{tabular}%
  \label{tab:semi_ablation_cars_fully}%
\end{table}%

\subsection{Semi-supervised Learning}\label{sec:semi_exp}
\noindent\textbf{Implementation Details.}
For the pre-training, we train an $M^2$-Track for 60 epochs. After that, we use the \textit{last} model to generate the pseudo labels on unlabeled data.
During the mixed-training stage, we clip the gradients whose norms are greater than 1 to avoid exploding gradients in the cycle training.
The possibility $p$ for applying the pseudo-label-based 
motion augmentation is set to 0.5.
The scale factor $\gamma$ in \textit{delete-cut-paste} operation is set to 1.25.
$\lambda$ and $\alpha$ in Eqn.~\eqref{eq:loss_forward} and Eqn.~\eqref{eq:loss_total} are empirically set to 0.1.
Other training configurations are kept the same as those in the fully-supervised experiments.

\subsubsection{Results on KITTI}

We first use the KITTI dataset to evaluate our online tracking performance under the SSL setting.
Each scene in the KITTI training set contains multiple track sequences. We obtain our labeled and unlabeled training sets by dividing the KITTI training scenes 0-16 into two non-overlapping sub-splits with a breakpoint $k$,
where we retain the labels in scenes 0-$k$ and drop the annotations in the scenes $(k\texttt{+}1)$-16.

\noindent\textbf{Results on Cars.}
We first evaluate SEMIM for cars in Tab.~\ref{tab:kitti_car}, which are the most common objects in autonomous driving.
For comparison, we set up two baselines $M^2$-Track and $M^2$-Track$^*$, which are fully trained using partial data under different labeled ratios.
Being consistent with Tab.~\ref{tab:kitti}, $M^2$-Track/$M^2$-Track$^*$ is trained with the basic/improved motion augmentation.
Since the improved motion augmentation more than doubles the data, We find that its effectiveness becomes more significant when the training data is limited, greatly benefiting the pre-training stage of SEMIM.
Under the SSL setting, SEMIM consistently improves the performance of the baselines under all labeled ratios.
With only 1\% labels, SEMIM amazingly boosts its $M^2$-Track$^*$'s poor performance by 16.7\%/20.9\% in terms of success/precision, demonstrating its ability to compensate for the lack of labels.
Surprisingly, SEMIM only needs 26\% labels to perform on par with its 100\%-supervised baseline,
and achieves state-of-the-art results when more labels are available.
We also notice that SEMIM's improvement over the baseline also consistently decreases as the number of labels increases. 
This is reasonable because more labeled data produces less noisy pseudo labels and provides more supervision to relieve overfitting.
Last but not least, even without any SSL techniques, our method shows better generalizability with limited training data when compared to previous matching-based approaches.
%



\noindent\textbf{Results on Pedestrians.}
Besides rigid objects, SEMIM also works on deformable objects like pedestrians. 
Different from cars, pedestrians usually cluster together, which is likely to distract the tracker.
Besides, the total amount of labeled pedestrians is much smaller than that of cars in KITTI, making the task more challenging.
Overall, Tab.~\ref{tab:kitti_pedestrian} shows a similar trend as Tab.~\ref{tab:kitti_car}.
But we notice that SEMIM's improvement for pedestrians is even more significant.
Trained with only 7\% labeled data (only 314 frames), the baseline $M^2$-Track failed to learn any meaningful information from the data, achieving only 4.8\%/9.7\% in terms of success/precision.
Even with the improved motion augmentation, $M^2$-Track$^*$ cannot achieve satisfying results (18.8\%/40.3\%) due to extremely limited training samples.
After applying SEMIM, the model's performance is significantly enhanced to 50.5\%/79.8\%, even surpassing 100\%-supervised P2B~\cite{qi2020p2b} and BAT~\cite{zheng2021box}.
We attribute this improvement to the pseudo-label-based motion augmentation in SEMIM, which covers sufficient motion priors of slow-moving objects.
As we expected, SEMIM also surpasses the powerful 100\% supervised $M^2$-Track$^*$ with only 40\% labels.
Please refer to the appendix for visualization results of SEMIM under different labeled ratios and object types.

\subsubsection{Results on WOD}
In this section,  we focus on adapting $M^2$-Track from KITTI to WOD, without using any labels from the latter.
At the beginning, we define the notations of several data splits as follows:
\begin{itemize}
  \setlength{\itemsep}{0pt}
  \setlength{\parsep}{0pt}
  \setlength{\parskip}{0pt}
      \item $\mathcal{D}_{kitti}^l$: the full KITTI training set (scenes 0-16) \textit{with} labels;
      \item $\mathcal{D}_{train}^l$: the training set of WOD \textit{with} labels;
      \item $\mathcal{D}_{train}^u$: the training set of WOD \textit{without} labels;
      \item $\mathcal{D}_{val}^u$: the validation set of WOD \textit{without} labels.
  \end{itemize}
Note that all the above splits are about car/vehicle objects.
\noindent\textbf{Unsupervised domain adaptation.}
As a semi-supervised technique, SEMIM can be directly adopted to the task of unsupervised domain adaptation (UDA). 
We first use WOD's validation set to evaluate the performance of  our best model, which is trained with $\mathcal{D}_{kitti}^l$ using the improved motion augmentation.
As shown in Tab.~\ref{tab:waymo}, the model only achieves 35.75\%/44.54\% in terms of success/precision, which is far lower than the result that we achieve on the KITTI test set (71.14\%/82.67\%).
The performance drop is mainly due to the domain gap between KITTI and WOD data.
After applying SEMIM using the labeled source data $\mathcal{D}_{kitti}^l$ and the unlabeled target data $\mathcal{D}_{train}^u$, we s
ee a significant improvement ($>$ 10\%) in terms of both success and precision.
We also compare our UDA model with the fully-supervised models which are trained using labels from the massive training set $\mathcal{D}_{train}^l$.
Even without any annotations from WOD, our UDA model performs comparably with the fully-supervised $M^2$-Track, while surpassing P2B~\cite{qi2020p2b} and BAT~\cite{zheng2022beyond} by large margins.
This shows that SEMIM also works well even when the labeled and unlabeled datasets are in different domains, which is valuable for practical applications.
%
%

\noindent\textbf{Offline auto-labeling.}
In this experiment, we train a model with SEMIM using the labeled data $\mathcal{D}_{kitti}^l$ and the unlabeled data $\mathcal{D}_{val}^u$.
After that, we use the trained model to make predictions on $\mathcal{D}_{val}^u$, and compare them with the ground truth labels.
Such an operation can be regarded as an offline labeling processing because the trained model has seen all the testing data (without labels) before it makes predictions on them.
Since the offline processing can fully exploit the statistics information in $\mathcal{D}_{val}^u$, it is supposed to be better than the online predictions, where the testing data is only available during the testing.
As shown in the last two rows in Tab.~\ref{tab:waymo}, the model using the offline processing performs better than the UDA model, even though $\mathcal{D}_{val}^u$ is much smaller than $\mathcal{D}_{train}^u$.
This suggests that SEMIM has great potential in offline auto-labeling, which helps to relieve the need for laborious human labeling.

\subsubsection{Ablation Study on SEMIM}
We extensively analyze the effectiveness of each component of SEMIM in Tab.~\ref{tab:semi_ablation_cars}.
All the experiments in Tab.~\ref{tab:semi_ablation_cars} are conducted on KITTI cars with 20\% annotations (\ie, breakpoint $k=2$), and the models are trained using the basic motion augmentation.
Considering the first two models in Tab.~\ref{tab:semi_ablation_cars}, we can see that the pseudo labels may even harm the training when applied alone.
This is because pseudo labels contain too much noise when the labels are insufficient during the pre-training.
By contrast, we can see a noticeable improvement when singly adding the cycle-consistent loss or the pseudo-label-based motion augmentation.
Especially, the pseudo-label-based motion augmentation brings a more significant improvement in terms of both success and precision ($>$10\%).
We also noticed that the effectiveness of the cycle-consistent loss is submerged by the pseudo-label-based motion augmentation when combining them together.
This is reasonable because the samples after the pseudo-label-based motion augmentation are treated as labeled data, which is not used in the computation of the cycle-consistent loss.
Despite this, the full model still achieves the best result in Tab.~\ref{tab:semi_ablation_cars}.

\subsubsection{SEMIM helps fully-supervised training.}\label{sec:fully_semim}
For previous semi-supervised experiments, SEMIM's components are only applied to unlabeled data.
In this section, we apply SEMIM's components to fully-labeled data to see whether they also help fully-supervised training.
Specifically, we additionally introduce a cycle-consistent loss term and apply the GT-based motion augmentation to labeled data.
The GT-based motion augmentation is modified from the pseudo-label-based motion augmentation.
Since labeled data are all associated with ground truths, it is unnecessary to generate pseudo labels for them.
Therefore, for pseudo-label-based motion augmentation, we conduct the \textit{delete-cut-paste} operation by replacing the pseudo labels with the ground truths.
Compared to pseudo-label-based motion augmentation which focuses on maintaining unseen motion introduced by pseudo labels while reducing the noise in them,  GT-based motion augmentation mainly augments the relative motion of a target object using other objects' ground truth movements.

We conducted experiments by fully training $M^2$-Track$^*$ separately using 20\% and 100\% labeled data from KITTI cars, and the results are shown in Tab.~\ref{tab:semi_ablation_cars_fully}.
Indeed, the cycle-consistent loss and the GT-based motion augmentation both offer benefits to fully-supervised training. However, when combined, they may exhibit redundancy or conflicting effects, leading to less improvement in performance.
Moreover, the effectiveness of each component when applied individually is contingent on the amount of training data available. The GT-based motion augmentation tends to be more effective when training data is limited (\ie, 20\%), as it helps alleviate overfitting in such scenarios. Conversely, the cycle-consistent loss becomes more powerful when ample training data is provided, as it can capture temporal coherence and improve tracking performance.
\section{Conclusions}\label[]{sec:conclusion}
In this work, we revisit 3D SOT in LiDAR point clouds and propose to handle it with a new motion-centric paradigm, which is proven to be an excellent complement to the matching-based Siamese paradigm.
In addition to the new paradigm, we propose a specific motion-centric tracking pipeline $M^2$-Track, which significantly outperforms the state-of-the-arts in various aspects.
Extensive analysis confirms that the motion-centric model is robust to distractors and appearance changes and can directly benefit from previous matching-based trackers.
Thanks to the motion-centric paradigm, we further extend our $M^2$-Track to handle semi-supervised learning in LiDAR SOT.
While using much fewer labels, the semi-supervised model performs on par with its 100\% supervised counterpart.
It also shows its impressive ability in domain adaptation and offline auto-labeling.
We believe that the motion-centric paradigm can serve as a primary principle to guide future architecture designs.
%

\ifCLASSOPTIONcompsoc
  \section*{Acknowledgments}
\else
  \section*{Acknowledgment}
\fi

{\noindent This work was supported in part by the Basic Research Project No. HZQB-KCZYZ-2021067 of Hetao Shenzhen HK S\&T Cooperation Zone, by the National Key R\&D Program of China with grant No.2018YFB1800800, by Shenzhen Outstanding Talents Training Fund, by Guangdong Research Project No. 2017ZT07X152 and No. 2019CX01X104, by the Guangdong Provincial Key Laboratory of Future Networks of Intelligence (Grant No. 2022B1212010001), by the NSFC 61931024\&8192 2046, by NSFC-Youth 62106154\&62302399, by zelixir biotechnology company Fund, by Tencent Open Fund, and by ITSO at CUHKSZ.}

\ifCLASSOPTIONcaptionsoff
  \newpage
\fi



\bibliographystyle{IEEEtran}
\bibliography{ref}

%
\begin{IEEEbiography}[{\includegraphics[width=1in,height=1.25in,clip,keepaspectratio]{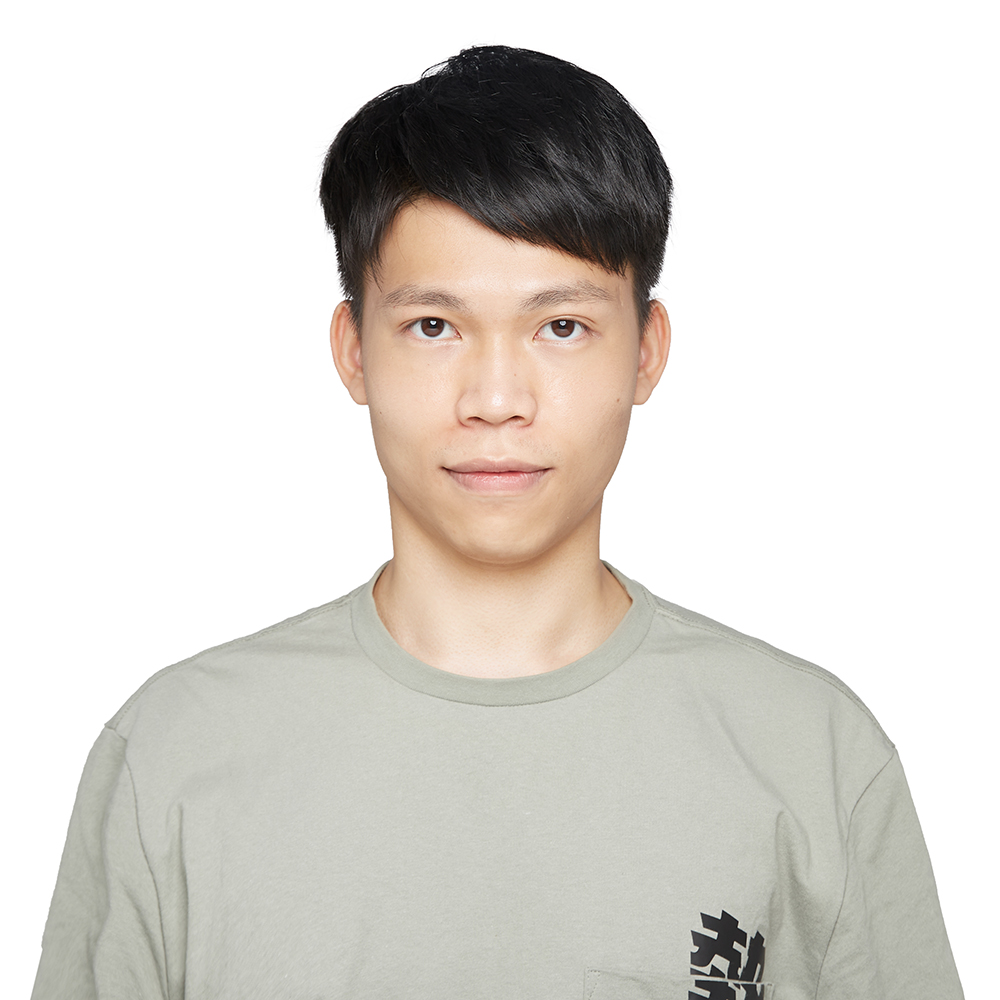}}]{Chaoda Zheng} received the B.Eng. and M.Eng. degrees from the South China University of Technology. He is currently pursuing the Ph.D. degree at the Chinese University of Hong Kong (Shenzhen). His research interests focus on 3D computer vision, especially 3D point cloud analysis, in the field of which he has published multiple top conferences or journal papers, such as CVPR, ICCV, ECCV, NeurIPS, TIP, etc.
\end{IEEEbiography}
\begin{IEEEbiography}[{\includegraphics[width=1in,height=1.25in,clip,keepaspectratio]{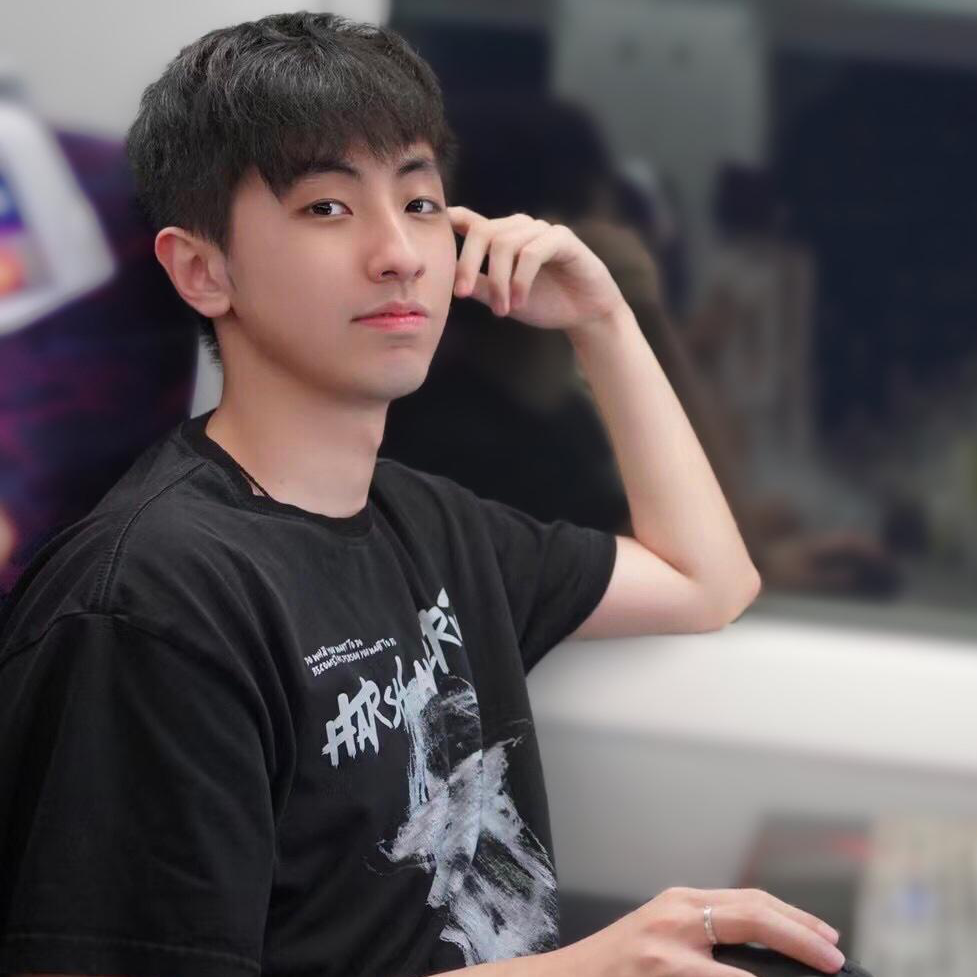}}]{Xu Yan} received the B.S. degree from the San Yat-sen University and Master degree from the Chinese University of Hong Kong (Shenzhen). He is currently pursuing the Ph.D. degree at the Chinese University of Hong Kong (Shenzhen). His research interests focus on 3D computer vision, especially 3D point cloud analysis. He has published more than 10 papers in top conferences, such as CVPR, ICCV, ECCV, NeurIPS, AAAI, IJCAI, ISBI, etc.
\end{IEEEbiography}
\begin{IEEEbiography}[{\includegraphics[width=1in,height=1.25in,clip,keepaspectratio]{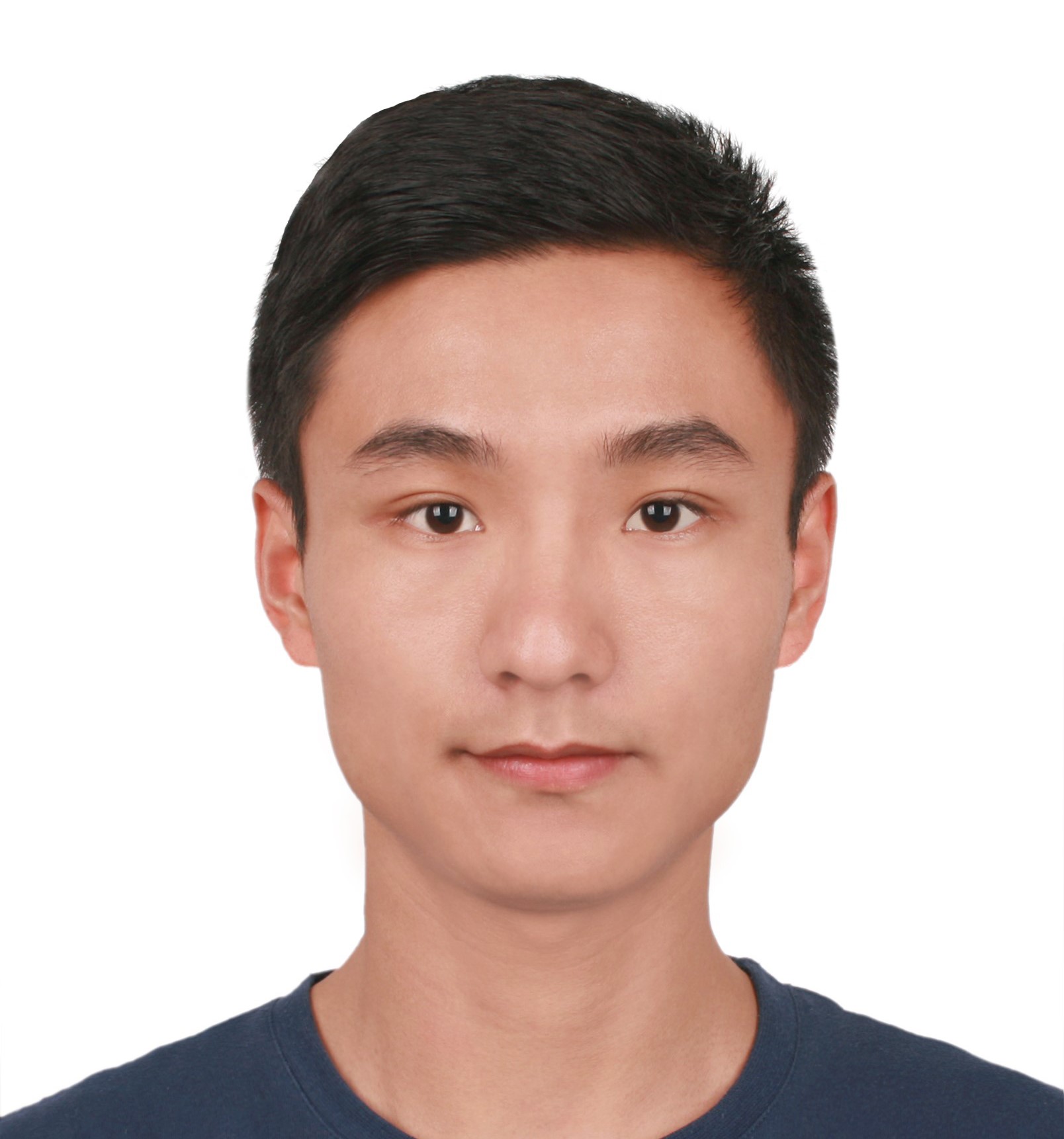}}]{Haiming Zhang} received the B.S. and M.S. degrees from the Beijing Institute of Technology. He is currently pursuing the Ph.D. degree in the Chinese University of Hong Kong (Shenzhen). His research interests include 3D point cloud analysis and generative adversarial networks. He has published a paper in CVPR.
\end{IEEEbiography}
\begin{IEEEbiography}[{\includegraphics[width=1in,height=1.25in,clip,keepaspectratio]{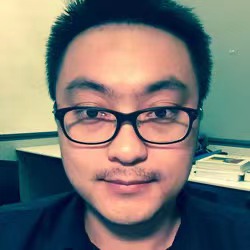}}]{Baoyuan Wang}, Ph.D. is the VP of engineering at Xiaobing.ai, a startup spun off from Microsoft in 2020. Prior to this, he was a Sr.Principal Researcher and Manager at Microsoft AI platform and HoloLen Mix-reality team from 2015 to 2021, and a lead researcher at MSRA from 2012 to 2015. His research interests include computational photography, digital human synthesis, and conversational AI. He got both his Bechler and Ph.D. degrees in computer science at Zhejiang University in 2007 and 2012 respectively. He was an engineering intern at Infosys, Mysore, India from 2006 to 2007 and a research intern at Microsoft from 2009 to 2012.
\end{IEEEbiography}
\begin{IEEEbiography}[{\includegraphics[width=1in,height=1.25in,clip,keepaspectratio]{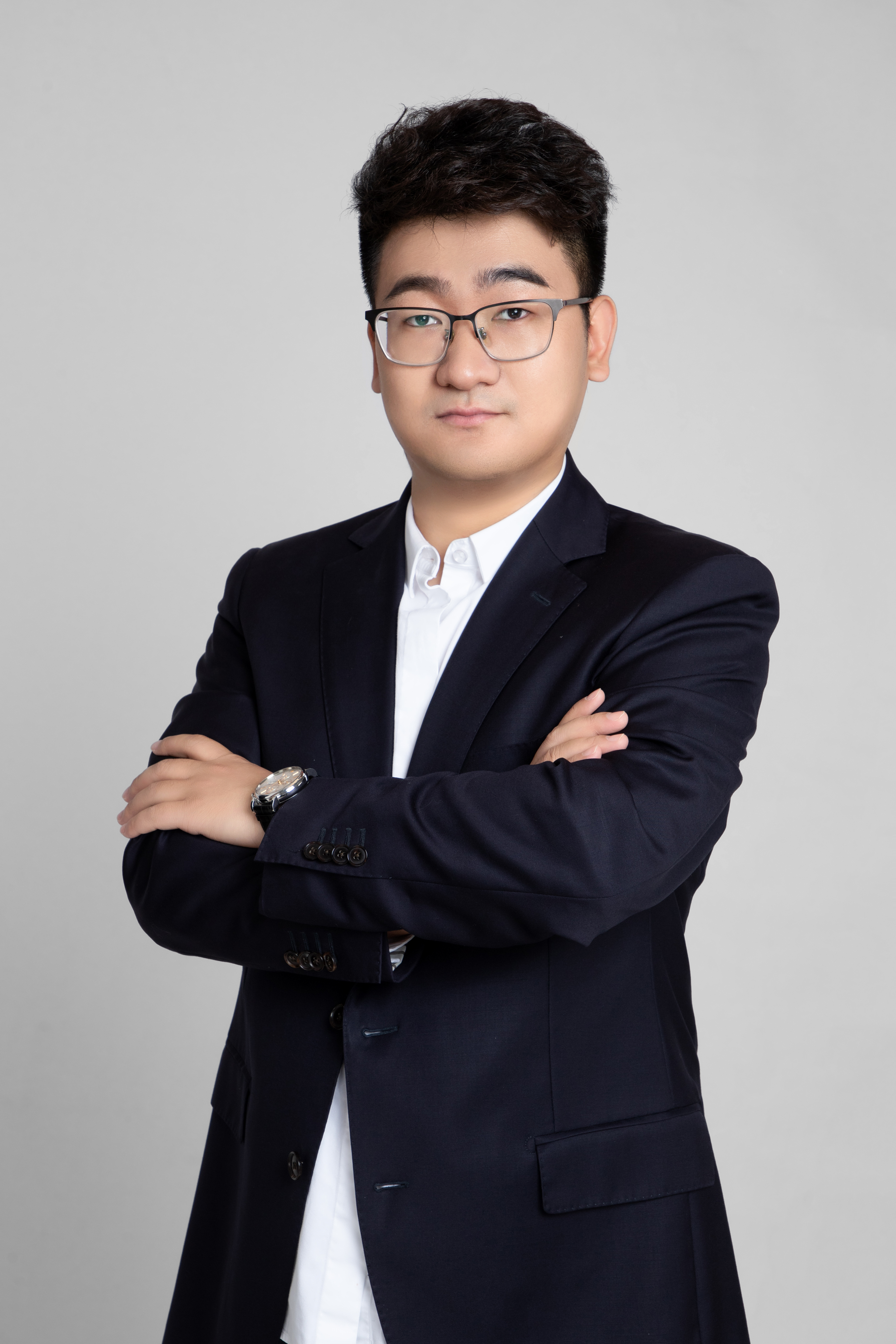}}]{Shenghui Cheng} is a westlake fellow at Westlake University, China. He received a Ph.D. in computer science from Stony Brook University and held research positions in the Institute of Medical Research of the University of Leipzig, Germany, Brookhaven National Laboratory and Harvard Medical School. He also served as a consultant for World Bank and Cedar Sinai Medical Center. He founded the Dagoo data visualization platform, served as a mentor of the Stanford Artificial Intelligence Global Executive Leadership Program, was the executive chairman of the CSIG-VIS Big Data Summit Forum, and served on the program committee of IEEE VIS, IEEE Pacific Vis, Chinavis and others.
\end{IEEEbiography}
\begin{IEEEbiography}[{\includegraphics[width=1in,height=1.25in,clip,keepaspectratio]{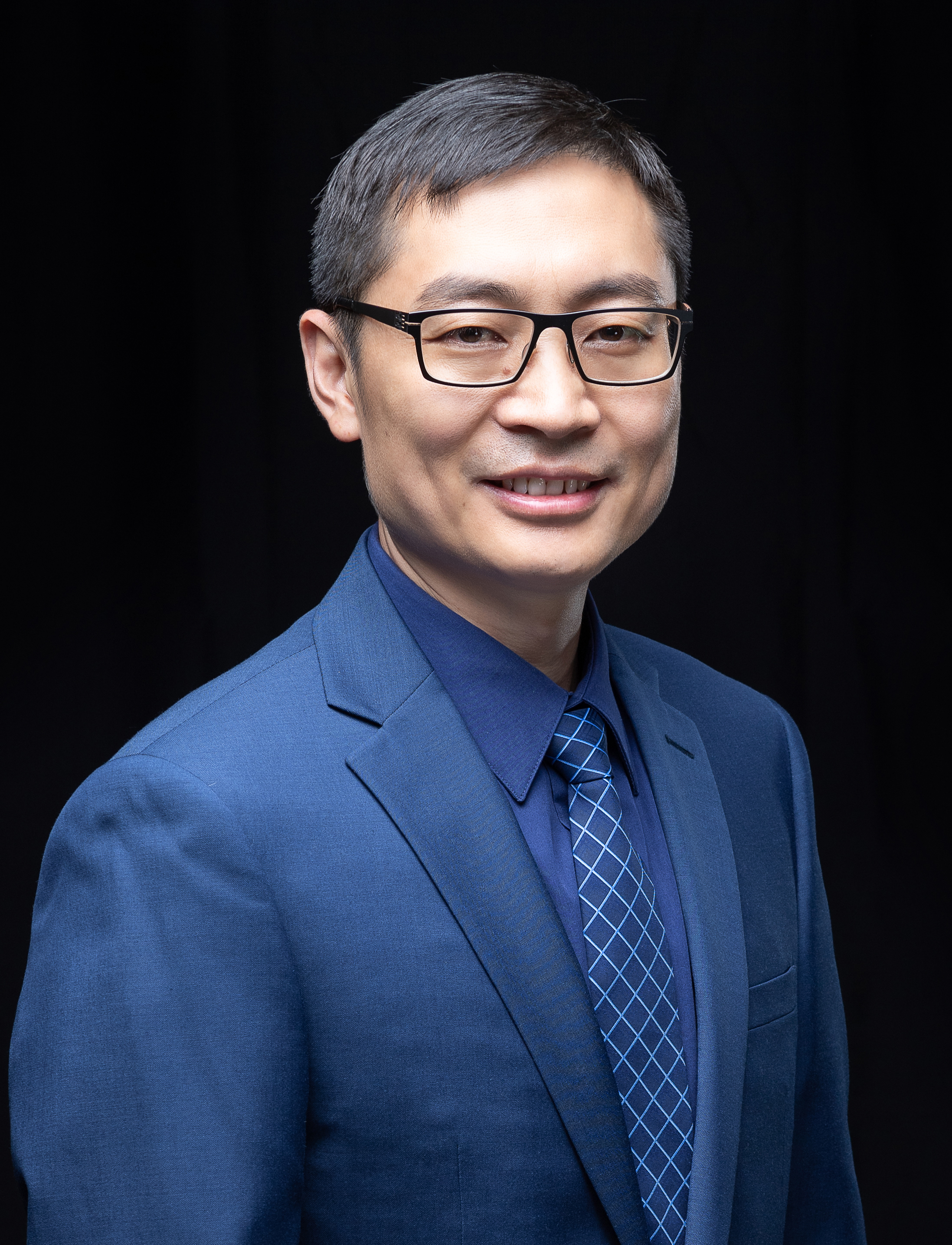}}]{Shuguang Cui} (S’99-M’05-SM’12-F’14) received his Ph.D in Electrical Engineering from Stanford University, California, USA, in 2005. Afterwards, he has been working as assistant, associate, full, Chair Professor in Electrical and Computer Engineering at the Univ. of Arizona, Texas A\&M University, UC Davis, and CUHK at Shenzhen respectively. He has also served as the Executive Dean for the School of Science and Engineering and is currently the Director for Future Network of Intelligence Institute (FNii) at CUHK, Shenzhen, and the Executive Vice Director at Shenzhen Research Institute of Big Data. His current research interests focus on data driven large-scale system control and resource management, large data set analysis, IoT system design, energy harvesting based communication system design, and cognitive network optimization. He was selected as the Thomson Reuters Highly Cited Researcher and listed in the Worlds’ Most Influential Scientific Minds by ScienceWatch in 2014. He was the recipient of the IEEE Signal Processing Society 2012 Best Paper Award. He has served as the general co-chair and TPC co-chairs for many IEEE conferences. He has also been serving as the area editor for IEEE Signal Processing Magazine, and associate editors for IEEE Transactions on Big Data, IEEE Transactions on Signal Processing, IEEE JSAC Series on Green Communications and Networking, and IEEE Transactions on Wireless Communications. He has been the elected member for IEEE Signal Processing Society SPCOM Technical Committee (2009 2014) and the elected Chair for IEEE ComSoc Wireless Technical Committee (2017 2018). He is a member of the Steering Committee for IEEE Transactions on Big Data and the Chair of the Steering Committee for IEEE Transactions on Cognitive Communications and Networking. He was also a member of the IEEE ComSoc Emerging Technology Committee. He was elected as an IEEE Fellow in 2013, an IEEE ComSoc Distinguished Lecturer in 2014, and IEEE VT Society Distinguished Lecturer in 2019. In 2020, he won the IEEE ICC best paper award, ICIP best paper finalist, the IEEE Globecom best paper award. In 2021, he won the IEEE WCNC best paper award.
\end{IEEEbiography}

\begin{IEEEbiography}[{\includegraphics[width=1in,height=1.25in,clip,keepaspectratio]{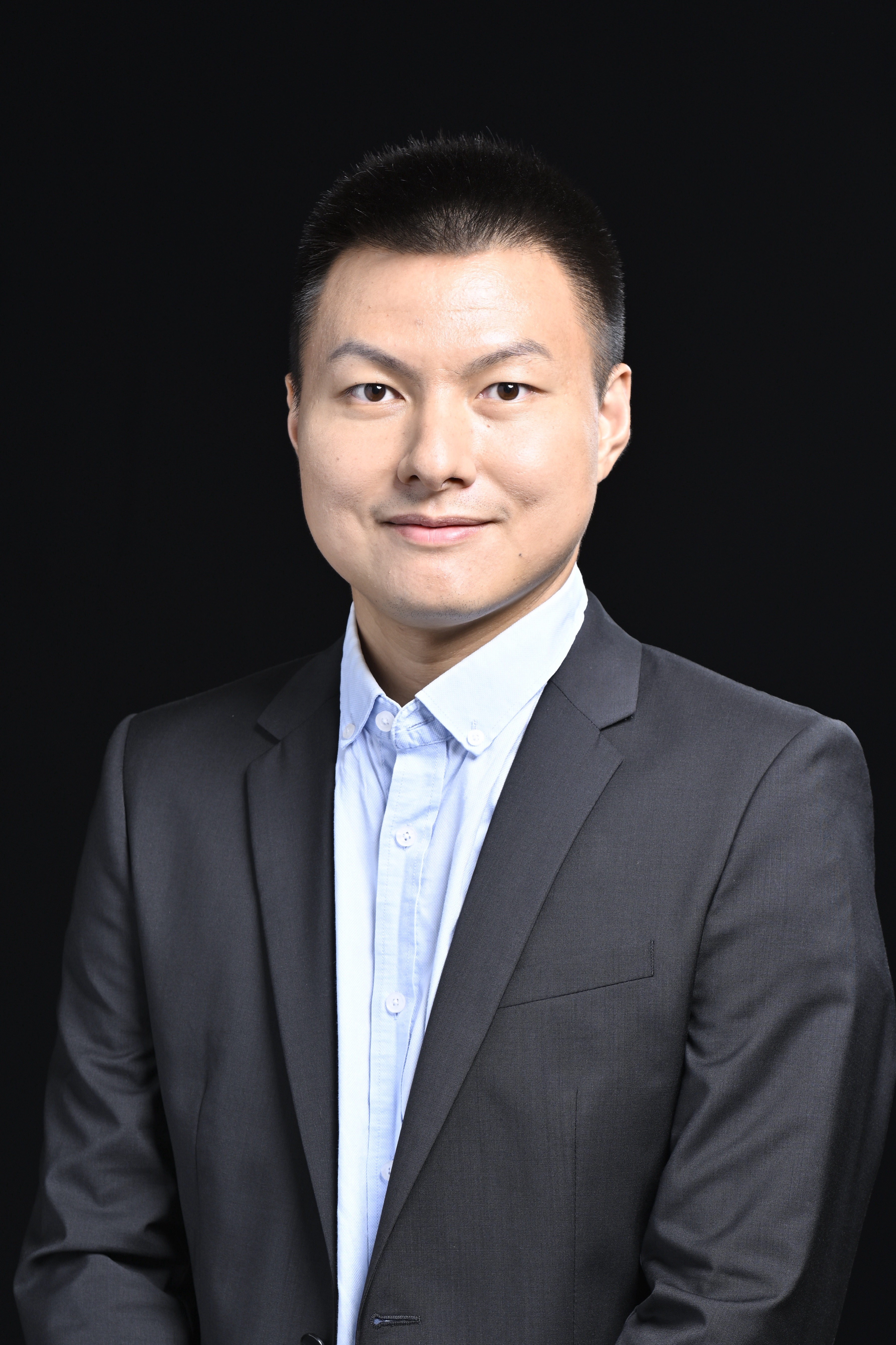}}]{Li Zhen} is currently an assistant professor at the School of Science and Engineering (SSE) of The Chinese University of Hong Kong (Shenzhen)/Future Intelligent Network Research Institute (FNii) of The Chinese University of Hong Kong (Shenzhen). He is also a research scientist at the Shenzhen Institute of Big Data (SRIBD) and a special researcher at the South China Hospital Affiliated to Shenzhen University. Dr. Li Zhen was selected for the 2021-2023 Seventh China Association for Science and Technology Young Talent Support Project. Dr. Zhen Li received his PhD in Computer Science from the University of Hong Kong (2014-2018), a MS in Communication and Information Systems from Sun Yat-Sen University (2011-2014), and a BS in Automation from Sun Yat-Sen University (2007-2011). He was also a visiting scholar at the University of Chicago in 2018 and a visiting student at the Toyota Technical Institute (TTIC) in Chicago in 2016. His research interests include interdisciplinary research in artificial intelligence, 3D vision, computer vision, and deep learning-assisted medical big data analysis. He has published more than 30 papers in top conferences and journals, such as top journals Cell Systems and Nature Communications, IEEE TNNLS, IEEE TMI, PLOS CB, Bioinformatics, JCIM etc., and top conferences CVPR, ICCV, ECCV, NeurIPS, AAAI, IJCAI, ACL, ECAI, MICCAI , RECOMB, ISBI, etc.

\end{IEEEbiography}
%
%
%




\appendices

\section{Implementation Details}\label{supp:details}
\noindent\textbf{Network Architecture}

\noindent The target segmentation network is a PointNet segmentation network~\cite{qi2017pointnet}, where each input point is firstly processed by a 5-layer MLP with output channel sizes of 64, 64, 64, 128, 1024. A max-pooling is then applied along the point dimension to obtain a 1024-dim global embedding, which is then copied and concatenated with the output of the second layer (64-dim). The concatenated features are then processed by another 5-layer per-point MLP with output channel sizes 512, 256, 128, 128, 2. Every layer of the MLP except the last one has batch normalization and ReLU. The output logits are used to extract the target points from the input.

\noindent At the $1^{st}$-stage, we use a vanilla PointNet~\cite{qi2017pointnet} to encode the spatial-temporal target points (with the temporal channel) into a 256-dim embedding. The PointNet includes a 4-layer per-point MLP with output sizes 64, 128, 256, 512, a point-wise max-pooling layer, and another MLP with output sizes 512, 256. We have batch normalization and ReLU for every layer of the MLP. On top of the encoded embedding, we independently apply two MLPs with 3 hidden layers (128,128,128) to obtain the motion state (6-dim) and the {\textbf{R}}elative {\textbf{T}}arget {\textbf{M}}otion (RTM) for previous box refinement (4-dim). The motion state includes a 4-dim RTM and a 2-dim motion classification logit.

\noindent At the $2^{nd}$-stage, we use a similar PointNet~\cite{qi2017pointnet} as in the $1^{st}$-stage to regress a 4-dim RTM on the denser target point cloud (without the temporal channel). The PointNet includes a 4-layer per-point MLP with output sizes 64, 128, 256, 512, a point-wise max-pooling layer, and another MLP with output sizes 512, 256, 4. We have batch normalization and ReLU for every layer of the MLP except the last one.

\begin{algorithm}
  \caption{Workflow of the $1^{st}$-stage}\label{alg:stage1}
  \textbf{Input}:  Segmented target points $\mathcal{\widetilde{P}}_{t-1,t}$ and possible target BBox $\mathcal{{B}}_{t-1}$ at the previous frame.

  \textbf{Output}:  A relative target motion state ( including a RTM $\mathcal{M}_{t-1,t}$ and 2D binary motion state logits), a refined target BBox $\mathcal{\widetilde{B}}_{t-1}$ at the previous frame, and a coarse target BBox $\mathcal{B}_{t}$ at the current frame.
  
  \begin{algorithmic}[1]
    \State Use a PointNet to encode $\mathcal{\widetilde{P}}_{t-1,t}$ to an embedding $\mathcal{E}$.
    \State Obtain an RTM with respect to $\mathcal{{B}}_{t-1}$ by applying an MLP to the embedding $\mathcal{E}$.
    \State Obtain the refined BBox $\mathcal{\widetilde{B}}_{t-1}$ at the previous frame by transforming $\mathcal{{B}}_{t-1}$ using the RTM predicted in \textit{step 2}.
    \State Obtain the motion state by applying another MLP to the embedding $\mathcal{E}$. The the motion state includes an RTM $\mathcal{M}_{t-1,t} \in \mathbb{R}^4$ and a 2D logit indicating whether the target is dynamic or not.
    \State If the target is dynamic, obtain the coarse $\mathcal{B}_{t}$ by transforming $\mathcal{\widetilde{B}}_{t-1}$ using the RTM $\mathcal{M}_{t-1,t}$. Otherwise, set $\mathcal{B}_{t} = \mathcal{\widetilde{B}}_{t-1}$.
  \end{algorithmic}
\end{algorithm}

\begin{algorithm}
  \caption{Workflow of the $2^{nd}$-stage}\label{alg:stage2}
  \textbf{Input}: Segmented target points $\mathcal{\widetilde{P}}_{t-1,t}$, the coarse target BBox $\mathcal{B}_{t}$ at the current frame and the motion state predicted in the $1^{st}$-stage.

  \textbf{Output}:  A refined target BBox $\mathcal{\widetilde{B}}_{t}$ at the current frame.
  
  \begin{algorithmic}[1]
    \State Extract $\mathcal{\widetilde{P}}_{t-1} \in \mathbb{R}^{M_{t-1} \times 3}$ and $\mathcal{\widetilde{P}}_{t} \in \mathbb{R}^{M_t \times 3}$ from $\mathcal{\widetilde{P}}_{t-1,t} \in \mathbb{R}^{({M_{t-1} + M_{t})} \times 4}$ according to the timestamp. 
    \State If the target is dynamic, transform $\mathcal{\widetilde{P}}_{t-1}$ to $\mathcal{\hat{P}}_{t-1} $ using the RTM $\mathcal{M}_{t-1,t}$. Otherwise, simply set $\mathcal{\hat{P}}_{t-1} = \mathcal{\widetilde{P}}_{t-1}$.
    \State Form a denser target point cloud $\mathcal{\hat{P}}_{t} \in \mathbb{R}^{(M_{t-1} + M_t) \times 3}$ by merging $\mathcal{\hat{P}}_{t-1}$ and $\mathcal{\widetilde{P}}_{t}$.
    \State Transform $\mathcal{\hat{P}}_{t}$ to the canonical coordinate system defined by $\mathcal{B}_{t}$.
    \State Apply a PointNet on the canonical $\mathcal{\hat{P}}_{t}$ to regress a RTM.
    \State Obtain the refined BBox $\mathcal{\widetilde{B}}_{t}$ by transforming $\mathcal{B}_{t}$ using the RTM predicted in \textit{step 5}.
  \end{algorithmic}
\end{algorithm}
\begin{figure*}[t!]
  \centering
   \includegraphics[width=0.9\linewidth]{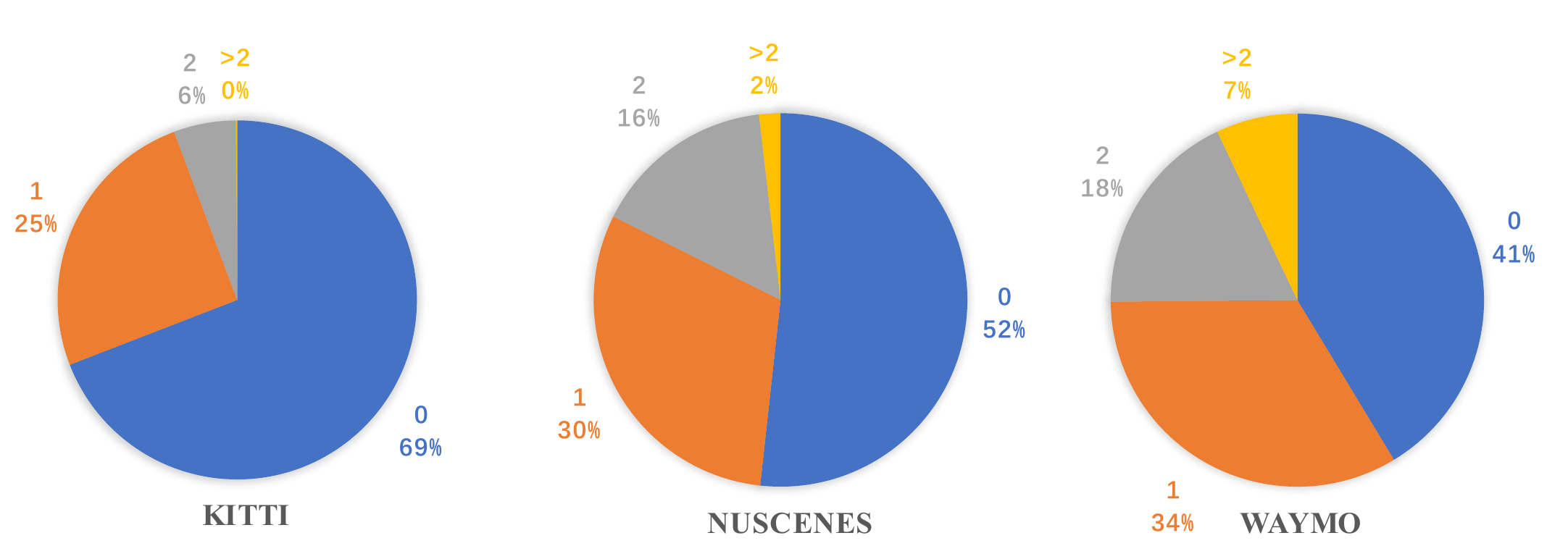}

   \caption{
    Distributions of distractors for \textbf{Cars (Vehicles)} in KITTI, NuScenes, and Waymo Open Dataset. We enlarge each target BBox by \textbf{2 meters} and count the distractors inside. Objects with the same category as the target are regarded as distractors.
     }
   \label{fig:statistics}
\end{figure*}
\begin{figure*}[h]
  \centering
   \includegraphics[width=1\linewidth]{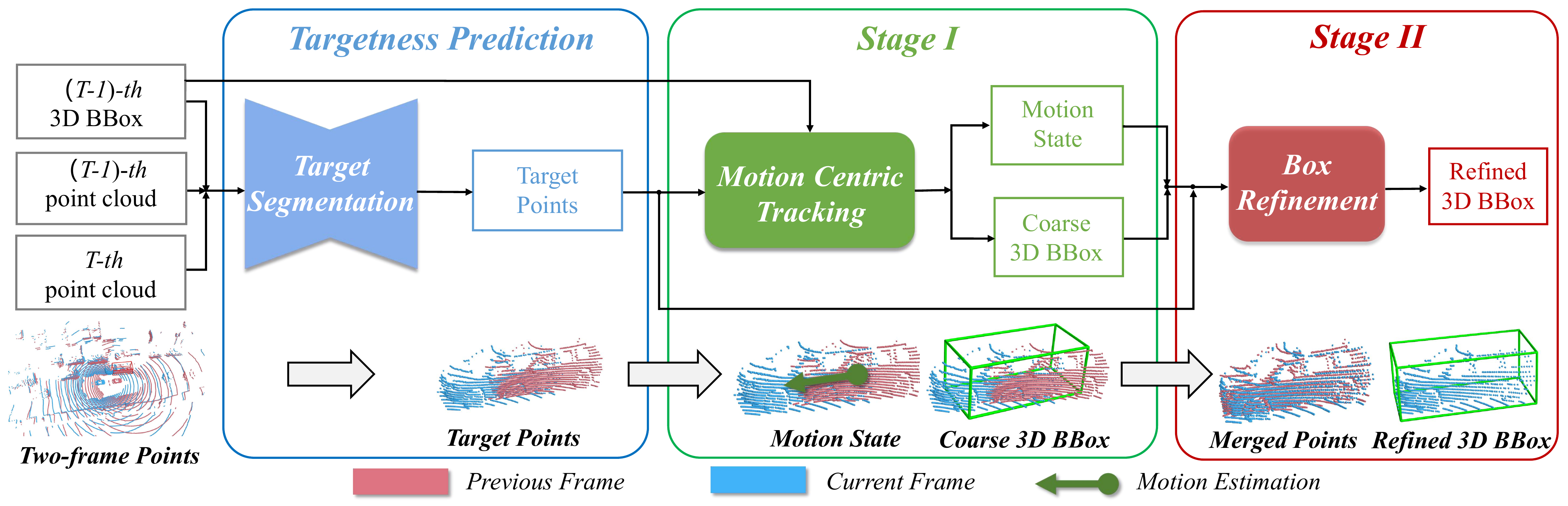}

   \caption{
    The pipeline of $M^2$-Track. The data flow is illustrated with \textbf{black} lines.
     }
   \label{fig:pipeline}
\end{figure*}
\begin{figure*}[h]
  \centering
   \includegraphics[width=1\linewidth]{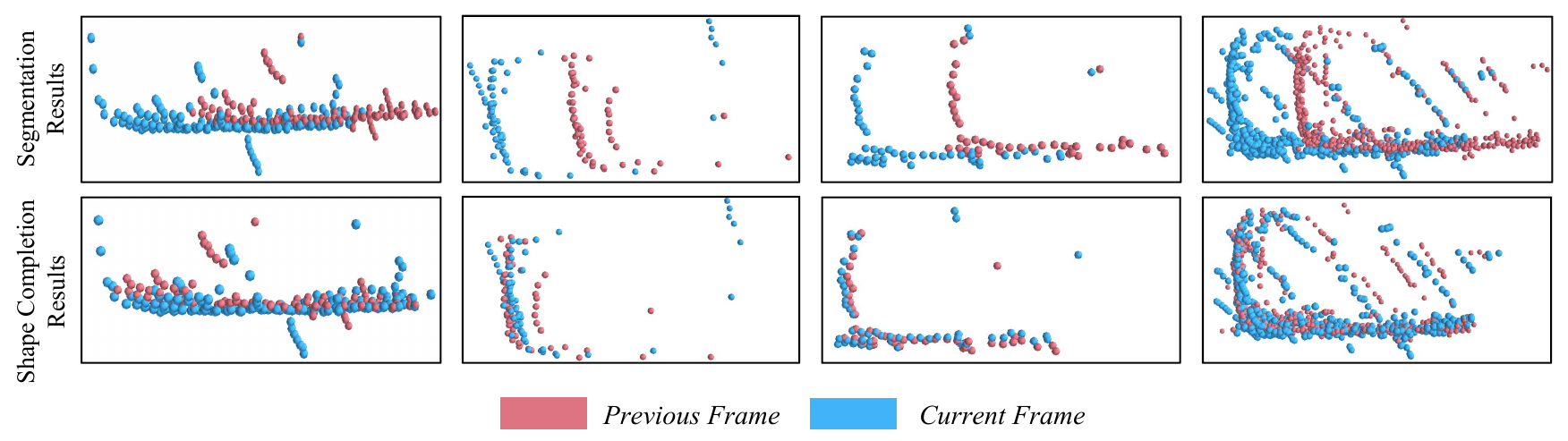}

   \caption{
     Visualization of the target segmentation and motion-assisted shape completion. Pictures in the same column are from the same case. The completion results demonstrate that our model learns good enough relative target motions.
     }
   \label{fig:seg_motion_vis}
\end{figure*}

\begin{figure*}[h]
  \centering
   \includegraphics[width=1\linewidth]{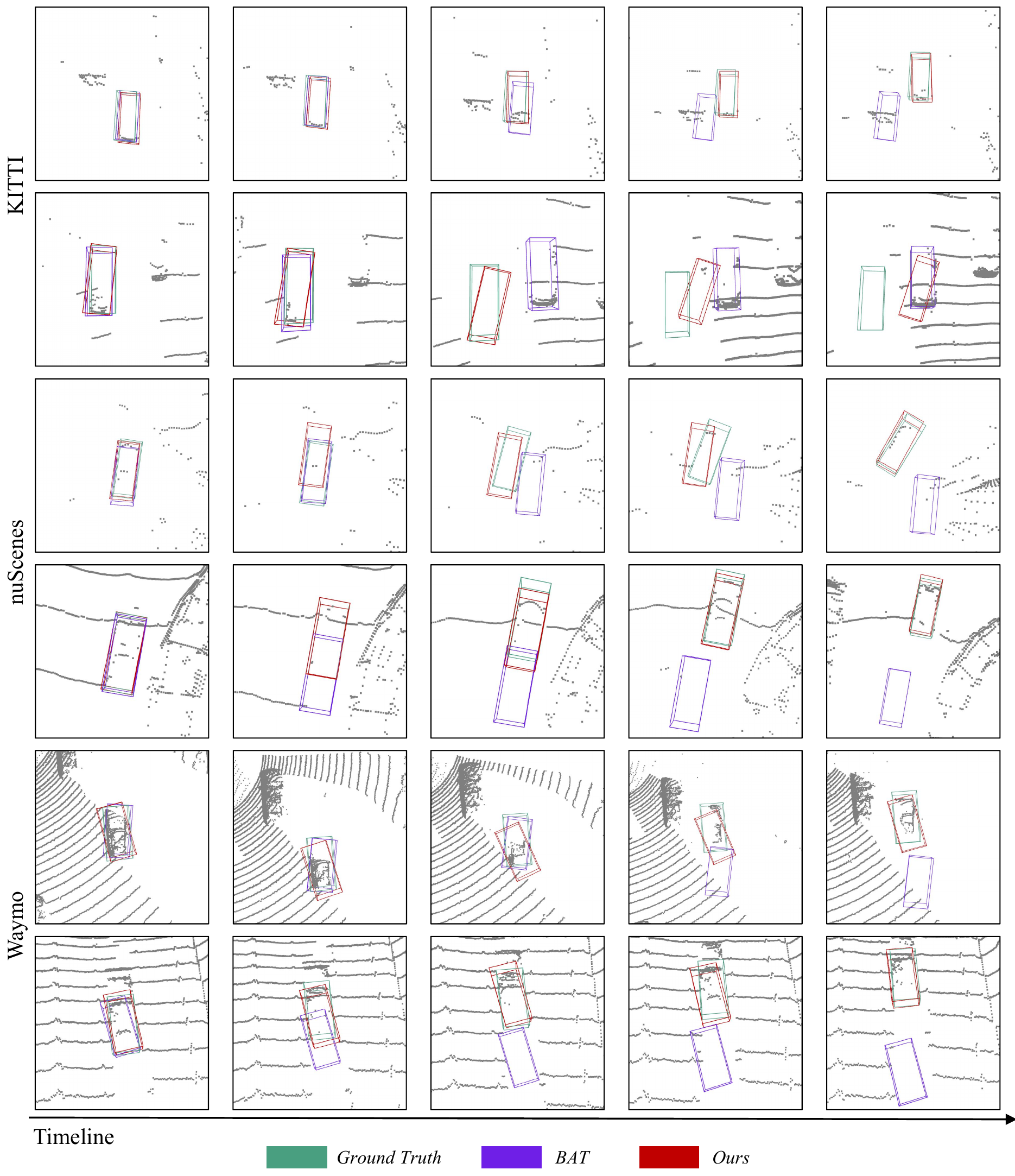}

   \caption{
     Visualization results for Cars.
     }
   \label{fig:car_vis}
\end{figure*}
\begin{figure*}[t!]
  \centering
   \includegraphics[width=1\linewidth]{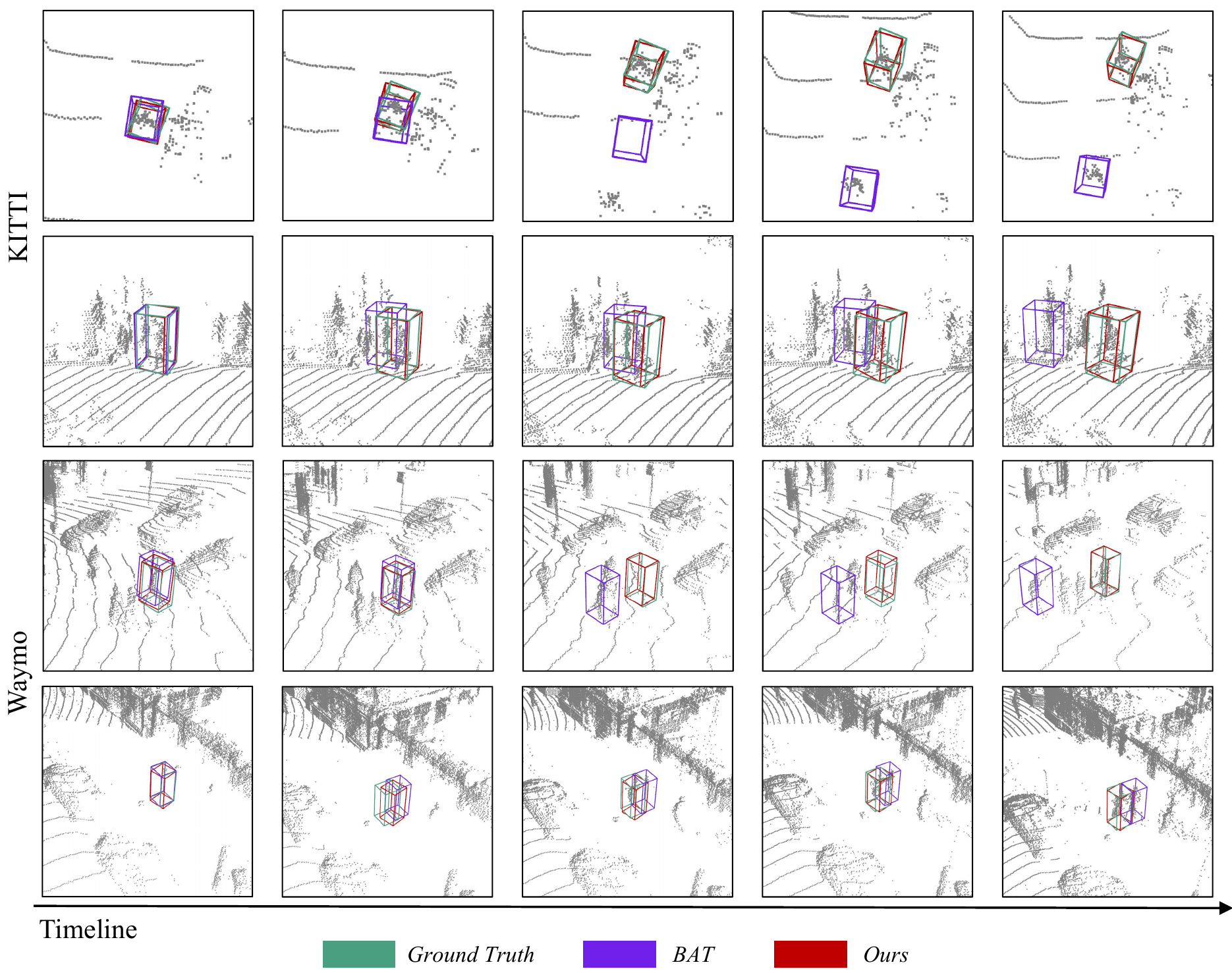}

   \caption{
     Visualization results for Pedestrians.
     }
   \label{fig:pedestrian_vis}
\end{figure*}
\noindent\textbf{Detailed Workflow}\\
The overall pipeline with the data flow of $M^2$-Track is presented in Fig.~\ref{fig:pipeline}.
The detailed description of the $1^{st}$-stage and $2^{nd}$-stage are provided in Alg.~\ref{alg:stage1} and  Alg.~\ref{alg:stage2}, respectively.

\noindent\textbf{4DOF RTM Transformation}\\
We focus on the 4DOF RTM within two successive frames. Given a 4D RTM $(\Delta x,\Delta y,\Delta z,\Delta \theta)$, we can construct a transformation matrix $\mathcal{T} \in \mathbb{R}^{4 \times 4}$ as follows:
\begin{equation}
  \begin{aligned}
    \begin{bmatrix} \cos \left( \Delta \theta \right)  & -\sin \left( \Delta \theta \right)  & 0 & \Delta x \\ \sin \left( \Delta \theta \right)  & \cos \left( \Delta \theta \right)  & 0 & \Delta y \\ 0 & 0 & 1 & \Delta z \\ 0 & 0 & 0 & 1 
    \end{bmatrix}
\end{aligned}
\end{equation}
We define the above process as a function $\mathcal{T}: \mathbb{R}^4 \mapsto \mathbb{R}^{4 \times 4}$.

\noindent\textbf{Point Transformation.} Given any object point $p = (x,y,z)$ in a BBox $\mathcal{B} = (x_b,y_b,z_b,\theta_b,w_b,l_b,h_b)$, we can transform it using an $\text{RTM} = (\Delta x,\Delta y,\Delta z,\Delta \theta)$ under the homogeneous coordinates:
\begin{equation}
\label{eq:transform}
  \begin{aligned}
    \begin{bmatrix}
      \hat x \\ \hat y \\ \hat z \\ 1
      \end{bmatrix}
    =
    \mathcal{T}(\mathcal{B}[:4])
    \times
    \mathcal{T}(\text{RTM})
    \times
    \mathcal{T}(\mathcal{B}[:4])^{-1}
    \times
    \begin{bmatrix}
    x \\ y \\ z \\ 1
    \end{bmatrix}
\end{aligned}
\end{equation}
Here $\hat p = (\hat x,\hat y,\hat z)$ denotes the transformed point. And $\mathcal{B}[:4] = (x_b,y_b,z_b, \theta_b)$ is the 4DOF pose of the BBox $\mathcal{B}$. 
Note that we transform all scenes from different datasets to the same \textit{right-handed} coordinate system with the $z$-axis pointing upward.

\noindent\textbf{Box Transformation.} For a BBox $\mathcal{B} = (x,y,z,\theta,w,l,h)$, we transform its center using Eqn.~\eqref{eq:transform}. The transformed BBox is $\mathcal{\hat B} = (\hat x,\hat y,\hat z,\theta + \Delta \theta,w,l,h)$, where $(\hat x,\hat y,\hat z)$ is the transformed center, $(\theta + \Delta \theta)$ is the transformed heading angle, and $w,l,h$ are the width, length and height of the BBox which remain unchanged.\\

\section{Hyper-Parameters and Model Complexity}\label{sec:miscellaneous}
\begin{table}[]
  \centering
  \caption{Study of the hyper-parameters for fully-supervised $M^2$-Track. All the models are trained with improved motion augmentation. Our default setting is given in the first row}
  \begin{tabular}{ccccc|cc}
  \toprule
  $\lambda_1$ & $\lambda_2$ & $\lambda_3$ & $\lambda_4$ & $p$   & Success & Precision \\
  \midrule
  0.1        & 0.1        & 1.0          & 1.0          & 0.5 & 71.14   & 82.67     \\
  \midrule
  0.2        & 0.1        & 1.0          & 1.0          & 0.5 & \textbf{72.13}   & \textbf{84.51}    \\
  0.4        & 0.1        & 1.0          & 1.0          & 0.5 & 70.88   & 83.20         \\
  \midrule
  0.1        & 0.2        & 1.0          & 1.0          & 0.5 & 70.08   & 82.14     \\
  0.1        & 0.4        & 1.0          & 1.0          & 0.5 & 70.68   & 82.61     \\
  \midrule
  0.1        & 0.1        & 0.5          & 1.0          & 0.5 & 72.06   & 83.48     \\
  0.1        & 0.1        & 1.5          & 1.0          & 0.5 & 70.57   & 82.79     \\
  \midrule
  0.1        & 0.1        & 1.0          & 0.5          & 0.5 & 69.22   & 81.76     \\
  0.1        & 0.1        & 1.0          & 1.5          & 0.5 & 70.16   & 82.14     \\
  \midrule
  0.1        & 0.1        & 1.0          & 1.0          & 0.3 & 68.12   & 81.42     \\
  0.1        & 0.1        & 1.0          & 1.0          & 0.7 & 69.07   & 81.70    \\
  \bottomrule
  \end{tabular}
  \label{tab:hyper_fully}
\end{table}
\begin{table}[]
  \centering
  \caption{Study of the hyper-parameters for semi-supervised $M^2$-Track with SEMIM. All the models are trained with improved motion augmentation. Our default setting is given in the first row.}
  \begin{tabular}{cccccc}
  \toprule
  $\lambda$ & $\alpha$ & $\gamma$ & $p$ & Success               & Precision             \\
  \midrule
  0.1       & 0.1      & 1.25     & 0.5 & 65.20                 & 79.90                 \\
  \midrule
  0.05      & 0.1      & 1.25     & 0.5 & \textbf{65.81}                 & \textbf{80.07}                 \\
  0.2       & 0.1      & 1.25     & 0.5 & 65.26                 & 79.43                 \\
  0.3       & 0.1      & 1.25     & 0.5 & 66.80                 & 80.59                 \\
  \midrule
  0.1       & 0.05     & 1.25     & 0.5 & 65.46                 & 78.96                 \\
  0.1       & 0.2      & 1.25     & 0.5 & 66.04                 & 79.85                 \\
  0.1       & 0.3      & 1.25     & 0.5 & 63.85                 & 78.59                 \\
  \midrule
  0.1       & 0.1      & 1.00        & 0.5 & 65.67                 & 79.00                 \\
  0.1       & 0.1      & 1.75     & 0.5 & 65.11                 & 79.39                 \\
  \midrule
  0.1       & 0.1      & 1.25     & 0.3 & 65.56 & 79.37 \\
  0.1       & 0.1      & 1.25     & 0.7 & 64.08 & 79.93 \\
  \bottomrule
  \end{tabular}
  \label{tab:hyper_semi}%
  \end{table}

\subsubsection{Hyper-parameters}\label{sec:hyper}
To showcase our method's ability to achieve excellent results with minimal tuning, we investigate the influence of hyper-parameters on both fully-supervised and semi-supervised training. 
All the experiments are conducted on KITTI cars. And we only use 20\% labels for the semi-supervised training (\ie, breakpoint $k=2$).

\noindent\textbf{Fully-supervised training.}
Tab.~\ref{tab:hyper_fully} shows the results of fully-supervised $M^2$-Track under different hyper-parameter configurations.
The $\lambda_1$, $\lambda_2$, $\lambda_3$, and $\lambda_4$ are the weights in Eqn.~\eqref{eq:loss_full} in the main paper. And $p$ denotes the probability used in the coin-flip test for the improved motion augmentation.

\noindent\textbf{Semi-supervised training.}
Tab.~\ref{tab:hyper_semi} displays the results of semi-supervised $M^2$-Track with SEMIM under various hyper-parameter configurations.
$\lambda$ and $\alpha$ are the weights in Eqn.~\eqref{eq:loss_forward} and Eqn.~\eqref{eq:loss_total} in the main paper, respectively. And $\gamma$ and $p$ are the scale factor and the probability used in the pseudo-label-based motion augmentation.

As demonstrated in Tab.~\ref{tab:hyper_fully} and Tab.~\ref{tab:hyper_semi}, both $M^2$-Track and SEMIM exhibit insensitivity to hyper-parameters, showcasing robust performance across different configurations.
Interestingly, our adopted settings do not necessarily yield the best results in both fully-supervised and semi-supervised settings. This observation suggests that there is room for further improvement through extensive tuning of the hyper-parameters.
In summary, our methods show promising capabilities even without extensive tuning, and there is potential for further enhancements by exploring more refined hyper-parameter configurations.

\subsubsection{Model Complexity}\label{sec:complexity}
In addition to achieving high performance, our method also demonstrates remarkable efficiency. To compare our approach's complexity with other high-efficient trackers P2B~\cite{qi2020p2b} and BAT~\cite{zheng2021box}, we conduct complexity evaluations on a single Nvidia 3090 GPU.

For FLOPs, we leverage the \textit{torchprofile} library to compute the MACs of the models and then convert them to FLOPs. To assess memory consumption, we train each model with batch size = 64 and report its corresponding GPU memory usage.
In terms of speed, we conduct independent evaluations on the inference speed of each model with batch size = 1. Unlike BAT~\cite{zheng2021box}, which solely reports the time of model forwarding, we additionally consider the pre- and post-processing time during inference including the time for calculating metrics.
Specifically, we test each model on full Pedestrian sequences (6088 frames) in KITTI test split and calculate the FPS (frames per second) by dividing the total time by the number of frames. We perform three runs for each model and report the average FPS. However, it's worth noting that due to varying hardware specifications (\eg, CPU, GPU, memory, storage, etc.), the reported speed may differ across different platforms, particularly when accounting for data processing time. Hence, the absolute reported speed is provided for reference purposes only, and one should interpret it with consideration of the relative relations between different models. 

As depicted in Tab.~\ref{tab:efficiency_compare}, our top-performing model $M^2$-Track surpasses previous trackers in terms of FLOPs, speed, and memory consumption. Notably, our proof-of-concept model M-Vanilla achieves unparalleled efficiency in all aspects.
Remarkably, M-Vanilla achieves comparable tracking performance to $M^2$-Track while consuming over $\sim$80\% fewer FLOPs and $\sim$70\% less memory. This substantial reduction in computational complexity highlights the efficiency gains achieved by M-Vanilla.
In Tab.~\ref{tab:flops_breakdown}, we present a detailed breakdown of the FLOPs for $M^2$-Track. Notably, the segmentation network accounts for approximately 70\% of the FLOPs in the full model, demonstrating that M-Vanilla's exceptional efficiency primarily stems from the removal of the segmentation network.

\begin{table}[]
  \renewcommand\tabcolsep{4pt} 
  \footnotesize
  \centering
  \caption{Efficiency Comparison.  Train. Mem. stand for GPU memory consumption during the training (batch size = 64)}
  \begin{tabular}{ccccc}
  \toprule
  Method  & FLOPs      & Params Size  & Speed  & Train. Mem. \\
  \midrule\midrule
  $M^2$-Track (Ours) & 5.07G & 8.54MB  &  57.4 FPS                       & 10037MB   \\
  M-Vanilla (Ours)& 0.71G  & 2.56MB &  75.9 FPS                     & 3085MB    \\
  \midrule
  P2B~\cite{qi2020p2b}    & 8.51G & 5.11MB & 24.9 FPS               & 17331MB   \\
  BAT~\cite{zheng2021box}     & 5.50G & 5.64MB &  43.6 FPS            & 14195MB  \\
  \bottomrule
  \end{tabular}
  \label{tab:efficiency_compare}
\end{table}

\begin{table}[]
  \centering
  \caption{FLOPs breakdown of $M^2$-Track.}
  \begin{tabular}{cccc}
  \toprule
  Segmentation & Stage I    & Stage II   & Total      \\
  \midrule
  3.64G   & 0.71G & 0.71G& 5.07G \\
  \bottomrule
  \end{tabular}
  \label{tab:flops_breakdown}
  \end{table}

\section{More Analysis}
\noindent\textbf{Distractor Statistics}\label{supp:analysis}

%
\noindent We count the number of distractors in each object's neighborhood in the training set of KITTI~\cite{Geiger2012CVPR}, NuScenes~\cite{caesar2020nuscenes}, and Waymo Open Dataset (WOD)~\cite{sun2020scalability} respectively.
Specifically, we enlarge each target BBox by 2 meters and count the number of annotated BBoxs which not only intersect with the enlarged area but also have the same category as the target. 
Fig.~\ref{fig:statistics} illustrates the distributions of distractors for cars/vehicles.
It shows that more than two-thirds (\textbf{69\%}) of the regions in KITTI are free of distractors.
While in NuScenes and WOD, distractors are very common, especially for WOD.
Besides, though we only consider a pretty small neighborhood around the target,  some regions in NuScenes and WOD have even more than two distractors. 
We do the same analysis on the pedestrians in KITTI and find that 68.3\% of the regions has at least 1 distractor(s).
All of these observations, together with our main experiment results, prove that $M^2$-Track is much more robust to distractors than previous matching-based approaches.
%

\noindent\textbf{Larger Search Area for NuScenes Cars}\\
By default, we enlarge the (predicted) target BBox at previous frame by \textbf{2 meters} and collect points inside to generate the inputs. This strategy is also adopted in P2B~\cite{qi2020p2b} and BAT~\cite{zheng2021box} to generate their search areas.
The 2 meters larger area is sufficient for KITTI and WOD, where keyframes are sampled at 10Hz.
However, NuScenes only provides keyframes at 2Hz. Thus, the target may move more than 2 meters even within two consecutive keyframes.
For a fair comparison, we only report our results with 2 meters in the main manuscript.
In Tab.~\ref{tab:offset}, we re-evaluate our performance on NuScenes Cars with \textbf{5 meters} larger search area. Note that using a larger area does not incur more computational costs because we keep the number of sampled points unchanged.
As shown in Tab.~\ref{tab:offset}, we can further improve the performance of $M^2$-Track by using larger search areas.
However, due to the increase of distractors and sparsity, larger search areas instead harm the performance of P2B and BAT.

\noindent\textbf{Limitations}\\
Unlike appearance matching, our motion-centric model requires a good variety of motion in the training data to ensure its generalization on data sampled with different frequencies. For instance, our model suffers from considerable performance degradation if trained with 2Hz data but tested with 10Hz data because the motion distribution of the 2Hz and 10Hz data differs significantly. But fortunately, we can aid this using a well-design motion augmentation strategy.

%
\begin{table}[t]
  \footnotesize
  \centering
  \caption{Larger search area for NuScenes Cars.}
    \begin{tabular}{c|cc}
    \toprule
    Method & Success & Precision \\
    \midrule\midrule
    BAT~\cite{zheng2021box} (2m) & 40.73  & 43.29  \\
    BAT~\cite{zheng2021box} (5m)& 37.14 \textcolor[rgb]{0.7,0.0,0.0}{\small $\downarrow$ 3.59} &  39.92 \textcolor[rgb]{0.7,0.0,0.0}{\small $\downarrow$ 3.37} \\
    \midrule
    P2B~\cite{qi2020p2b} (2m)& 38.81 & 43.18 \\
    P2B~\cite{qi2020p2b}  (5m)&  38.48 \textcolor[rgb]{0.7,0.0,0.0}{\small $\downarrow$ 0.33} &  42.15 \textcolor[rgb]{0.7,0.0,0.0}{\small $\downarrow$ 1.03} \\
    \midrule
    $M^2$-Track (2m)& 55.85    & 65.09  \\
    $M^2$-Track  (5m)& 58.35 \color{RoyalBlue}{\small $\uparrow$ 2.50} & 67.04 \color{RoyalBlue}{\small $\uparrow$ 1.95}\\
    \bottomrule
    \end{tabular}%
  \label{tab:offset}%
\end{table}%

  

\begin{figure*}[]
	\begin{centering}
		\includegraphics[width=\textwidth]{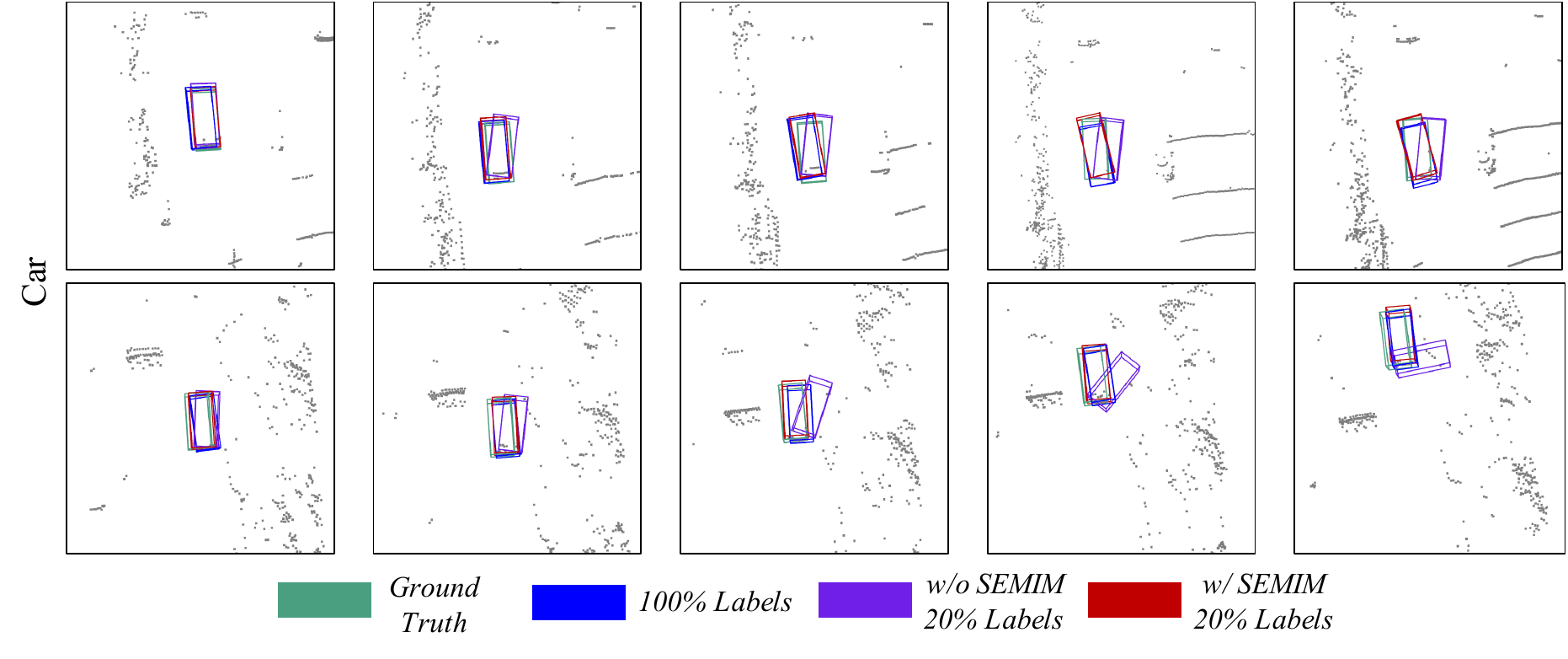}
		\caption{Visualization for KITTI Cars under the semi-supervised setting.}
		\label{fig:vis_car}
	\end{centering}	
\end{figure*}

\begin{figure*}[]
	\begin{centering}
		\includegraphics[width=1\textwidth]{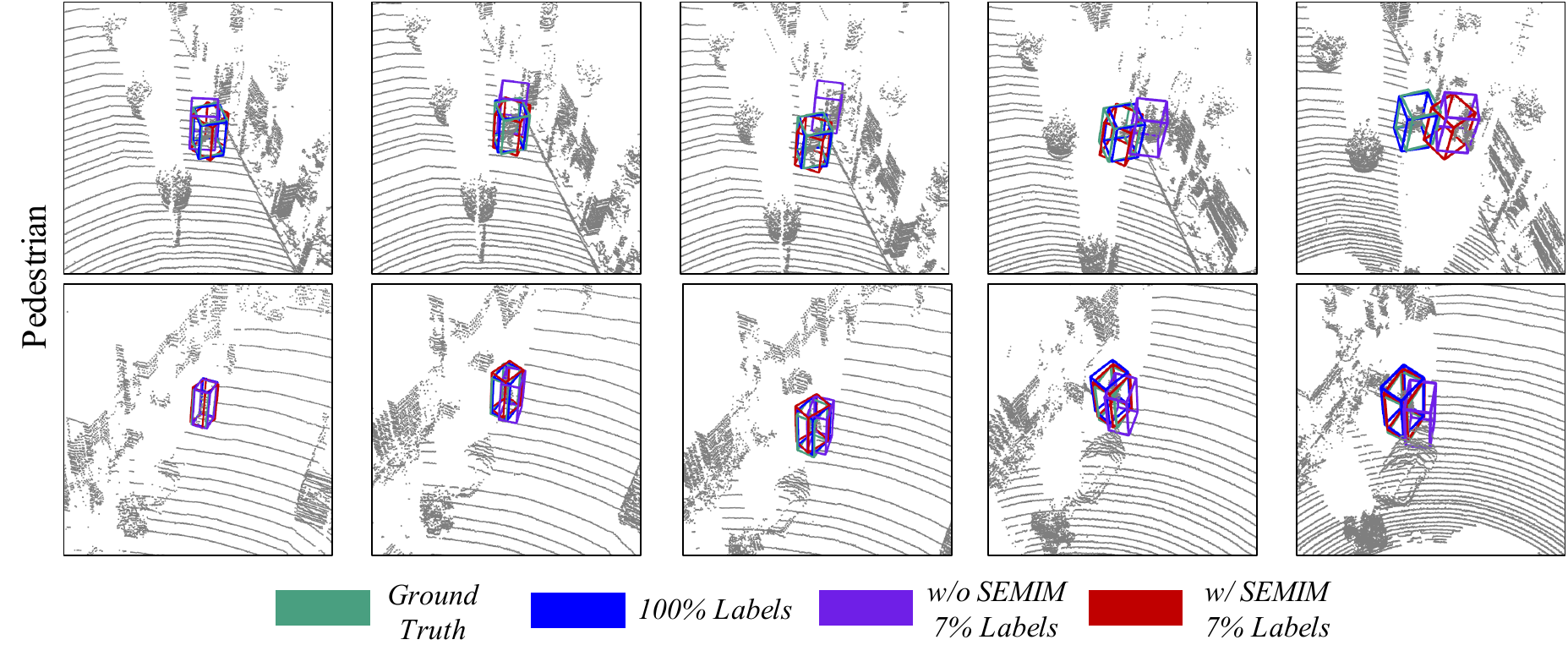}
		\caption{Visualization for KITTI Pedestrians under the semi-supervised setting.}
		\label{fig:vis_pedestrian}
	\end{centering}	
\end{figure*}

\section{Visualization}
\noindent\textbf{Target Segmentation}\\
Our model depends on the target segmentation to learn the relative target motion (RTM).
The first row in Fig.~\ref{fig:seg_motion_vis} shows our target segmentation results.
We can see that most segmented points are from the target objects, demonstrating the effectiveness of spatial-temporal learning.

\noindent\textbf{Motion-assisted Shape Completion}\\
In the $2^{nd}$-stage, we leverage the RTM $\mathcal{M}_{t-1,t}$ to complete the target point cloud at the current frame.
As shown in the second row in Fig.~\ref{fig:seg_motion_vis}, our method correctly merges the point clouds from two consecutive frames, using the predicted RTM.
These results demonstrate that the RTMs are correctly modeled by our method.

\noindent\textbf{Advantageous Cases in Fully-supervised Setting}\\
More qualitative comparison results are in Fig.~\ref{fig:car_vis} and Fig.~\ref{fig:pedestrian_vis}.
We also provide animated results in the attached video.
We can observe that our $M^2$-Track consistently shows its advantage when the scene is sparse, the relative target motion is large, or distractors exist in the target's neighborhood.
However, since our $M^2$-Track only takes LiDAR point clouds as input, it fails on extremely sparse scenarios where the number of target points is almost zero (\eg the second row in Fig.~\ref{fig:car_vis}).
Actually, this is a common issue for LiDAR-based SOT and could be probably solved by using multi-modal data (\eg. RGB images).

\noindent\textbf{Advantageous Cases in Semi-supervised Setting}\\
Fig.~\ref{fig:vis_car} and Fig.~\ref{fig:vis_pedestrian} show the visualization results for cars and pedestrians in KITTI under the semi-supervised setting.
The visualization results confirm that SEMIM significantly improves the performance of $M^2$-Track when labels are limited.
\end{document}